\documentclass{article}

\usepackage{ifthen}
\newcommand{\paperversion}{arxiv}

\PassOptionsToPackage{numbers, compress}{natbib}

\ifthenelse{\equal{\paperversion}{arxiv}}
{\usepackage{arxiv}}
{\usepackage{iclr2023_conference}}
\usepackage{times}
\usepackage[utf8]{inputenc} % allow utf-8 input
\usepackage[T1]{fontenc}    % use 8-bit T1 fonts
\usepackage{url}            % simple URL typesetting
\usepackage{amsfonts}       % blackboard math symbols
\usepackage{nicefrac}       % compact symbols for 1/2, etc.
\usepackage{microtype}
\usepackage{graphicx}
\usepackage{booktabs} % for professional tables
\usepackage[backref=page]{hyperref}
\usepackage{ifthen}
\usepackage[inline]{enumitem}
\usepackage{colortbl}
\usepackage{graphicx}
\usepackage{float}
\usepackage{caption} 
\usepackage{subcaption}
\usepackage{overpic}
\usepackage{mathtools}
\usepackage{tabularx}
\usepackage{amsmath,amssymb}
\usepackage{xspace}
\usepackage{adjustbox}
\usepackage{multirow}
\usepackage{wrapfig}
\usepackage{tikz}
\usepackage[nameinlink]{cleveref}
\usepackage[acronym]{glossaries}
\usepackage{placeins}
\usepackage{amssymb}% http://ctan.org/pkg/amssymb
\usepackage[normalem]{ulem}
\usepackage{pifont}% http://ctan.org/pkg/pifont
\usepackage{colortbl}

\usepackage{scalerel}
\usepackage{mleftright}
\usepackage{dsfont}
\usepackage{bm}

\glsdisablehyper{}
\newacronym{SL}{SL}{supervised learning}
\newacronym{SSL}{SSL}{self-supervised learning}
\newacronym{GM}{GM}{generative models}
\newacronym{AE}{AE}{auto-encoder based algorithms}
\newacronym{VAE}{VAE}{Variational Autoencoder}
\newacronym{IWAE}{IWAE}{Importance Weighted Autoencoder}
\newacronym{ID}{ID}{in-distribution}
\newacronym{OOD}{OOD}{out-of-distribution}
\newacronym{MI}{MI}{mutual information}
\newacronym{ECE}{ECE}{Expected Calibration Error}
\newacronym{AdaECE}{AdaECE}{Adaptive ECE}
\newacronym{SGD}{SGD}{Stochastic Gradient Descent}

\DeclareMathOperator{\1}{{}\mathds{1}}

\newcommand{\aoco}{\ensuremath{\text{acc}_o(f,c_o)}}
\newcommand{\aoci}{\ensuremath{\text{acc}_o(f,c_i)}}
\newcommand{\aici}{\ensuremath{\text{acc}_i(f,c_i)}}

\makeatletter
\newcommand{\subalign}[1]{%
  \vcenter{%
    \Let@ \restore@math@cr \default@tag
    \baselineskip\fontdimen10 \scriptfont\tw@
    \advance\baselineskip\fontdimen12 \scriptfont\tw@
    \lineskip\thr@@\fontdimen8 \scriptfont\thr@@
    \lineskiplimit\lineskip
    \ialign{\hfil$\m@th\scriptstyle##$&$\m@th\scriptstyle{}##$\hfil\crcr
      #1\crcr
    }%
  }%
}
\makeatother

\newcommand{\std}[1]{(\textit{\scriptsize$\pm$\scriptsize #1})}

\definecolor{pink}{HTML}{ff00ff}
\definecolor{orange}{HTML}{ff9900}
\definecolor{blue}{HTML}{0000ff}
\definecolor{lightblue}{HTML}{D1EDFF}
\definecolor{lb}{HTML}{E6F4FF}
\definecolor{db}{HTML}{ADD9FF}
\definecolor{lg}{HTML}{E6FCF1}
\definecolor{dg}{HTML}{B6EDCC}
\definecolor{g}{HTML}{F5F5F5}

\usepackage{amsmath,amsfonts,bm}

\def\eqref#1{equation~\ref{#1}}
\def\1{\bm{1}}

\DeclareMathAlphabet{\mathsfit}{\encodingdefault}{\sfdefault}{m}{sl}
\SetMathAlphabet{\mathsfit}{bold}{\encodingdefault}{\sfdefault}{bx}{n}

\usepackage{algorithm}
\usepackage{algorithmic}
\renewcommand{\algorithmiccomment}[1]{{\tt /\!\!/#1}}

\title{How Robust is Unsupervised Representation Learning to Distribution Shift?}

\author{Yuge Shi\thanks{Corresponding author, \texttt{yshi@robots.ox.ac.uk}}\\
Department of Engineering Science\\
University of Oxford\\
\And
Imant Daunhawer \& Julia E.~Vogt\\
Department of Computer Science \phantom{PLACEHOLDER}\\
ETH Zurich\\ 
\And
Philip H.S.~Torr\\
Department of Engineering Science\\
University of Oxford\\
\And
Amartya Sanyal\\
Department of Computer Science \& ETH AI Center\\
ETH Zurich
}

\ifthenelse{\equal{\paperversion}{arxiv}}{\arxivfinalcopy}{\iclrfinalcopy}

\begin{document}

\maketitle

% \begin{abstract}
% The vulnerability of machine learning models to spurious correlations has mostly been discussed using \gls{SL} models, and there lack insight on how spurious correlation affects other popular pre-training regimes, such as various \gls{SSL} algorithms and \gls{AE}.
% %
% In this work, we examine how vulnerable these pre-trained models are compared to supervised pre-training to spurious correlation.
% %
% Through experiments on both toy and realistic distribution shift tasks, we show \gls{SSL} to be consistently the best performing regime, followed by \gls{AE} and \gls{SL}.
% %
% Following observations that linear classifier itself can be susceptible to spurious correlation, we also experiment with a new evaluation scheme with the linear head trained on \gls{OOD} data.
% %
% We show that this simple change improves test performance significantly on all datasets that we investigate.
% %
% \end{abstract}

\begin{abstract}
\noindent
The robustness of machine learning algorithms to distributions shift
is primarily discussed in the context of \gls{SL}. 
As such, there is a lack of insight on the robustness of the representations learned from unsupervised methods, such as \gls{SSL} and \gls{AE}, to distribution shift.
%, such as robustness to spurious correlation and domain generalisation.
%
We posit that the \emph{input-driven} objectives of unsupervised
algorithms lead to representations that are more robust to
distribution shift than the \emph{target-driven} objective of SL. We
verify this by extensively evaluating the performance of \gls{SSL} and
\gls{AE} on both synthetic and realistic distribution shift datasets.
Following observations that the linear layer used for classification
itself can be susceptible to spurious correlations, we evaluate the
representations using a linear head trained on a small amount of
\gls{OOD} data, to isolate the robustness of the learned
representations from that of the linear head.
We also develop ``controllable'' versions of existing realistic domain
generalisation datasets with adjustable degrees of distribution
shifts. This allows us to study the robustness of different learning
algorithms under versatile yet realistic distribution shift
conditions. Our experiments show that representations learned from
unsupervised learning algorithms generalise better than SL under a
wide variety of extreme as well as realistic distribution shifts.

% We find that representations learned from unsupervised learning algorithms 1) exhibits stronger generalisation performance than SL under extreme distribution shift, and 2) experiences smaller performance drop when evaluating on \gls{OOD} as compared to \gls{ID} data.
%
% in addition, we observe that the generalisation performance of all
% models on all datasets can be significantly improved by retraining the
% final linear layer on small amounts of OOD data, highlighting the
% importance of isolating the linear head bias when evaluating
% generalisation performance.
%
\end{abstract}

\section{Introduction} \label{sec:introduction}

Machine Learning~(ML) algorithms are classically designed under the
statistical assumption that the training and test data are drawn from
the same distribution. However, this assumption does not hold in most
cases of real world deployment of ML systems.
%
%However, this is potentially detrimental for the deployment of ML systems in the real world.
% The i.i.d. assumption between the train and test set of most machine learning datasets is harmful for developing machine learning systems for the real world.
%
% The downstream tasks, in which ML algorithms are deployed, typically
% operate in different environments than the one where the training data
% was drawn from. 
For example, medical researchers might obtain their training data from hospitals in Europe, but deploy their trained models in Asia; the changes in conditions such as imaging equipment and demography result in a shift in the data distribution between train and test set~\citep{dockes2021preventing, glocker2019machine,henrich2010most}. 
% can arise due to multiple factors~\citep{dockes2021preventing}~e.g. changes in imaging equipment~\citep{glocker2019machine} and changes in demography~\citep{henrich2010most}.
% consist of data drawn from different distributions than the train
% set for various reasons including 
%
To perform well on such tasks requires the models to generalise to
unseen distributions --- an important property that is not evaluated
on standard machine learning datasets like ImageNet, where the train
and test set are sampled i.i.d. from the same distribution.

With increasing attention on this issue, researchers have been probing
the generalisation performance of ML models by creating datasets that
feature distribution shift tasks~\citep{koh2021wilds, domainbed,
shah2020pitfalls} and proposing algorithms that aim to improve
generalisation performance under distribution
shift~\citep{ganin2016dann, arjovsky2019irm, sun2016coral,
Sagawa2020dro, fish}. 
In this work, we identify three specific problems with current
approaches in distribution shift problems, in computer vision, and
develop a suite of experiments to address them.

\begin{figure*}[t]
  \centering
%   \vspace{-1em}
  \includegraphics[trim={0.5cm 0 0 0},clip, width=0.8\linewidth]{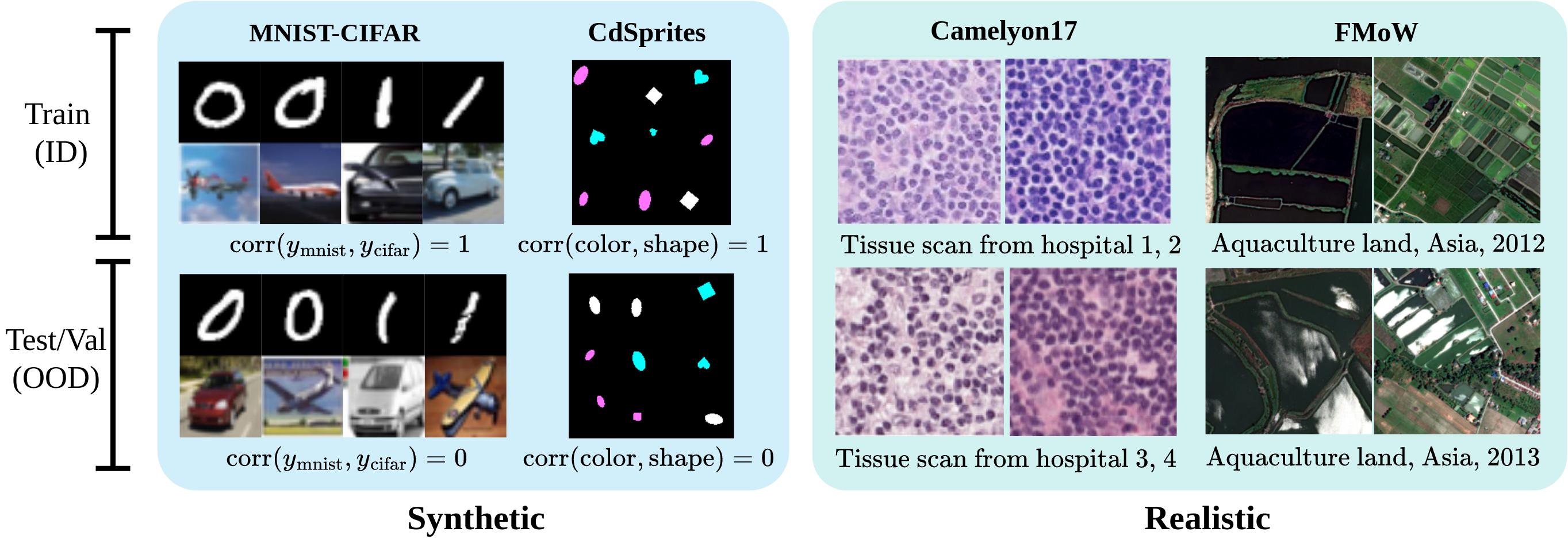}
  \vspace{-1em}
  \caption{\textbf{Synthetic vs. realistic distribution shift:} The distribution shift in synthetic datasets (\textit{left}, MNIST-CIFAR and CdSprites) are usually extreme and controllable (adjusted via changing the correlation); for realistic datasets (\textit{right}, WILDS-Camelyon17 and FMoW) distribution shift can be subtle, hard to identify and impossible to control.}
   \vspace{-20pt}
  \label{fig:synthetic_v_real}
\end{figure*}

  \subsection{Existing problems and Contributions} \label{sec:background}
  % \begin{itemize}[itemsep=1em,leftmargin=1em,labelwidth=*,align=left]
  % \vspace{-8pt}
  \paragraph{Problem 1: The outdated focus on supervised regime for distribution shift}
  In ML research, distribution shift has been studied in various
  contexts under different terminologies such as simplicity
  bias~\citep{shah2020pitfalls}, dataset
  bias~\citep{torralba2011datasetbias1}, shortcut
  learning~\citep{geirhos2020shortcut}, and domain adaptation and
  generalisation \citep{koh2021wilds, domainbed}.
  Most of these work are carried out under the scope of supervised
  learning (SL), including various works that either investigate
  spurious correlations~\citep{shah2020pitfalls, hermann2020shapes,
  kalimeris2019sgd} or those that propose specialised methods to
  improve generalisation and/or avoid shortcut
  solutions~\citep{arjovsky2019irm, ganin2016dann, Sagawa2020dro,
  teney2022evading}.
  However, recent research~\citep{shah2020pitfalls, geirhos2020shortcut} highlighted the extreme vulnerability of SL methods to spurious correlations: they are susceptible to learning only features that are irrelevant to the true labelling functions yet highly predictive of the labels.
  This behaviour is not surprising given SL's
  \emph{\textbf{target-driven}} objective: when presented with two
  features that are equally predictive of the target label, SL models
  have no incentive to learn both as learning only one of them
  suffices to predict the target label. 
  This leads to poor generalisation when the learned feature is
  missing in the \gls{OOD} test set.

  On the other hand, in recent times, research in computer vision has
  seen a surge of unsupervised representation learning algorithms.
  These include \acrfull{SSL} algorithms~(e.g.,
  \citet{chen2020simclr,grill2020byol,chen2021simsiam}), which learn
  representations by enforcing invariance between the representations
  of two distinctly augmented views of the same image,  and
  \acrfull{AE}~\citep{rumelhart1985ae,kingma2013vae,higgins2016betavae,burda2015iwae},
  which learn representations by reconstructing the input image.
  % such as \acrfull{SSL} and \acrfull{AE} \citep{chen2020simclr, grill2020byol, kingma2013vae, bao2021beit, he2021mae}.
  %
  The immense popularity of these methods are mostly owed to their
  impressive performance on balanced \gls{ID} test datasets
  %  While their impressive performance on various tasks led to
  % their popularity, these methods are mostly known for their strong
  % performance on balanced \gls{ID} test datasets 
  --- how they perform
  on distribution shift tasks remains largely unknown. 
  However, in distribution shift tasks, it is particularly meaningful
  to study unsupervised algorithms. This is because, in comparison to
  SL, their learning objectives are more \emph{\textbf{input-driven}}
  i.e.  they are incentivised to learn representations that most
  accurately represent the input data  \citep{chen2020simclr,vib}.
  When presented with two features equally predictive of the labels, unsupervised learning algorithms encourage the model to go beyond learning what's enough to predict the label, and instead focus on maximising the mutual information between the learned representations and the input.
  We hypothesise that this property of unsupervised representation
  learning algorithms helps them avoid the exploitation of spurious
  correlations, and thus fare better under distribution shift, compared
  to SL.
  
  \emph{\textbf{Contribution}: Systematically evaluate SSL and AE on
  distribution shift tasks.} We evaluate and compare the
  generalisation performance of unsupervised representation learning
  algorithms, including \gls{SSL} and \gls{AE}, with standard
  supervised learning.  See \cref{sec:setup} for more details on our
  experiments.
  
  \vspace{-8pt}
  \paragraph{Problem 2: Disconnect between synthetic and realistic datasets}
  Broadly speaking, there exists two types of datasets for studying
  distribution shift: synthetic datasets where the shift between
  train/test distribution is explicit and controlled (e.g. MNIST-CIFAR
  \citep{shah2020pitfalls}, CdSprites \citep{fish}) and realistic
  datasets featuring implicit distribution shift in the real world
  (e.g. WILDS \citep{koh2021wilds}). 
  We provide visual examples in \cref{fig:synthetic_v_real}.
  %
  % However, even though both types of dataset sets out to study OOD generalisation, their usage are commonly seen in two different sub-fields: typically, works that discuss simplicity bias, dataset bias and shortcut learning focus on synthetic datasets \citep{teney2022evading, shah2020pitfalls, ahn2022mitigating}, while those that propose new domain generalisation/adaptation algorithms emphasise evaluating their model on realistic datasets \citep{sun2016coral, ganin2016dann}.
  
  Synthetic datasets allow for explicit control of the distribution
  shift and are, thus, an effective diagnostic tool for generalisation
  performance. However, the simplistic nature of these datasets poses
  concerns about the generality of the findings drawn from these
  experiments;
  a model's robustness to spurious correlation on certain toy datasets
  is not very useful if it fails when tested on similar real-world
  problems.
  On the other hand, realistic datasets often feature distribution
  shifts that are subtle and hard to define~(see
  \cref{fig:synthetic_v_real}, right).
  As a result, generalisation performances of different algorithms
  tend to fluctuate across datasets~\citep{koh2021wilds, domainbed}
  with the cause of said fluctuations remaining unknown.
  
  % however we cannot explain what caused the change in results.
  
%   \begin{figure*}[t]
%       \centering
%       \includegraphics[trim={0.5cm 0 0 0},clip, width=0.8\linewidth]{images/hook.png}
%       \vspace{-1em}
%       \caption{\textbf{Synthetic vs. realistic distribution shift:} The distribution shift in synthetic datasets (\textit{left}, MNIST-CIFAR and CdSprites) are usually extreme and controllable (adjusted via changing the correlation); for realistic datasets (\textit{right}, WILDS-Camelyon17 and FMoW) distribution shift can be subtle, hard to identify and impossible to control.}
%       \vspace{-20pt}
%       \label{fig:synthetic_v_real}
%   \end{figure*}
  \begin{figure*}[t]
    \centering
    \captionsetup[subfigure]{belowskip=1ex}
    \begin{minipage}{\linewidth}
        \centering
        OOD Accuracy (higher is better)
    \end{minipage}
    \begin{subfigure}[t]{0.16\linewidth}
      \centering
      \includegraphics[width=\linewidth]{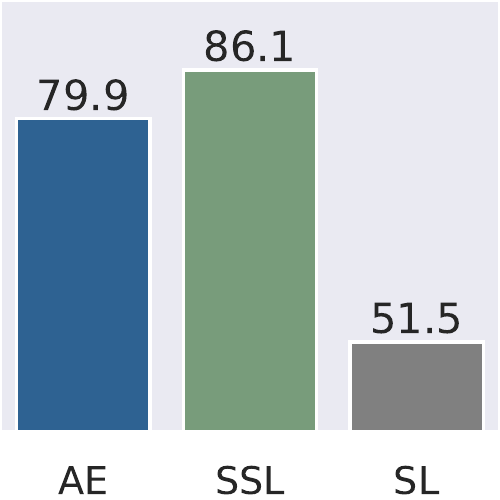}
    \end{subfigure}
    \begin{subfigure}[t]{0.16\linewidth}
      \centering
      \includegraphics[width=\linewidth]{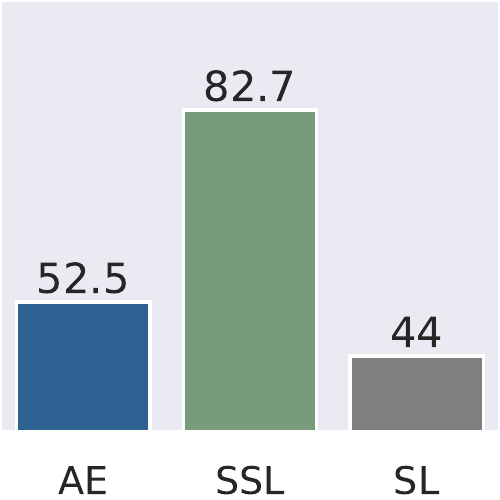}
    \end{subfigure}
    \begin{subfigure}[t]{0.16\linewidth}
      \centering
      \includegraphics[width=\linewidth]{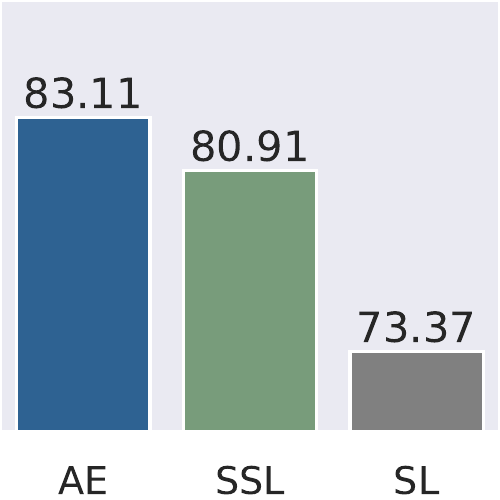}
    \end{subfigure}   
    \begin{subfigure}[t]{0.16\linewidth}
      \centering
      \includegraphics[width=\linewidth]{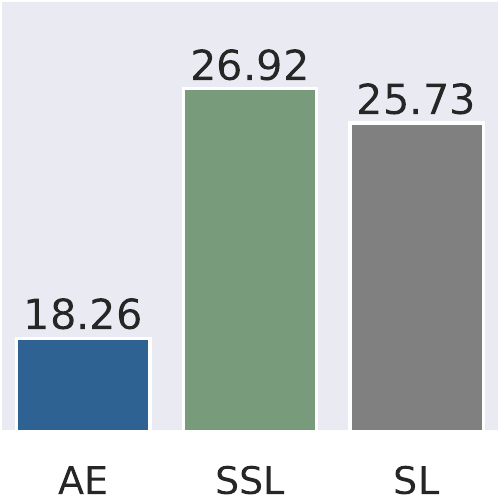}
    \end{subfigure}  
    \begin{subfigure}[t]{0.16\linewidth}
      \centering
      \includegraphics[width=\linewidth]{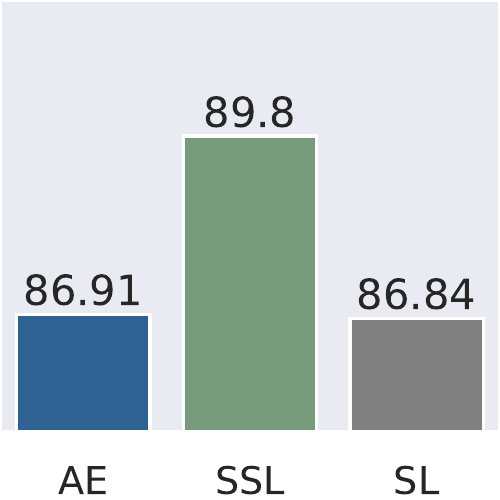}
    \end{subfigure}
    \begin{subfigure}[t]{0.16\linewidth}
      \centering
      \includegraphics[width=\linewidth]{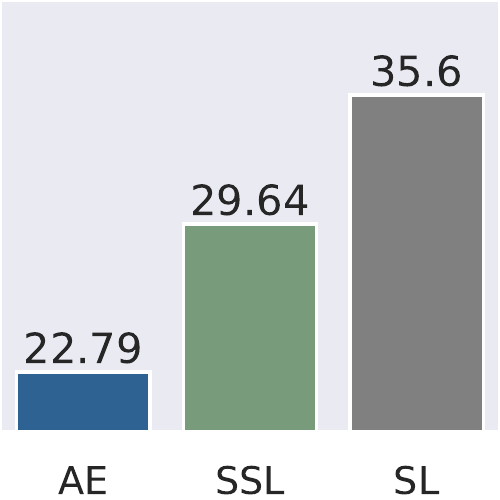}
    \end{subfigure} 
    \begin{minipage}{\linewidth}
        \centering
        Shift Sensitivity (lower is better)
        % Shift Sensitivity:~OOD Error - ID Error (lower is better)
    \end{minipage}  
    \begin{subfigure}[t]{0.16\linewidth}
      \centering
      \includegraphics[width=\linewidth]{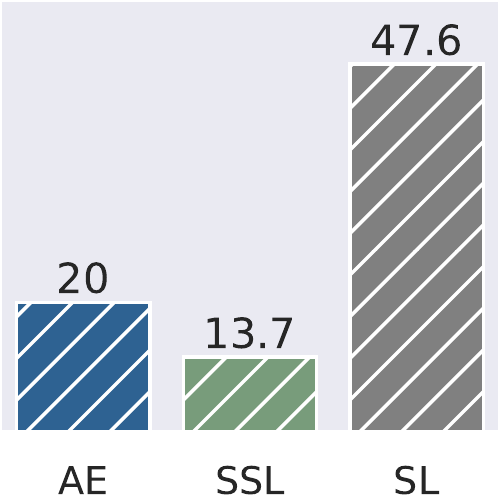}
      \caption{\scriptsize MNIST-CIFAR} \label{fig:ms_summary}
    \end{subfigure}
    \begin{subfigure}[t]{0.16\linewidth}
      \centering
      \includegraphics[width=\linewidth]{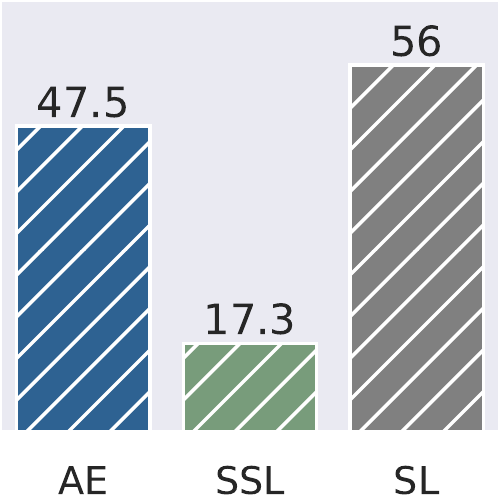}
      \caption{\scriptsize \centering CdSprites }\label{fig:cdsprites_summary}
    \end{subfigure}
    \begin{subfigure}[t]{0.16\linewidth}
      \centering
      \includegraphics[width=\linewidth]{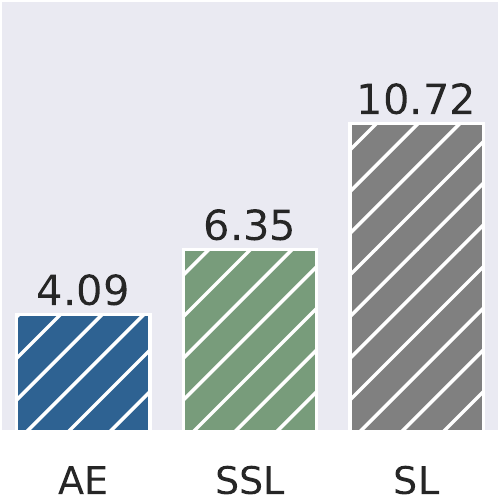}
      \caption{\scriptsize \centering Camelyon17-CS}
      \label{fig:camelyon_c_summary}
    \end{subfigure}   
    \begin{subfigure}[t]{0.16\linewidth}
      \centering
      \includegraphics[width=\linewidth]{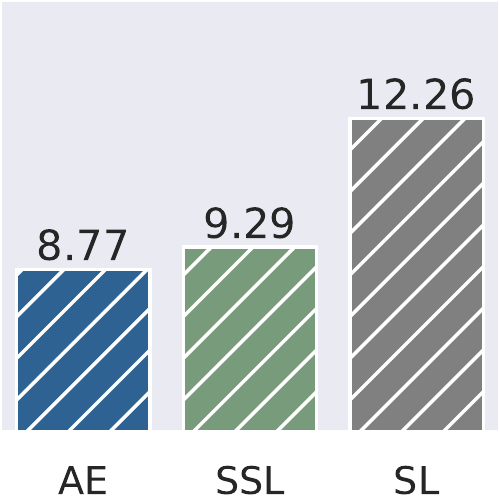}
      \caption{\scriptsize \centering FMoW-CS}\label{fig:fmow_c_summary}
    \end{subfigure}  
    \begin{subfigure}[t]{0.16\linewidth}
      \centering
      \includegraphics[width=\linewidth]{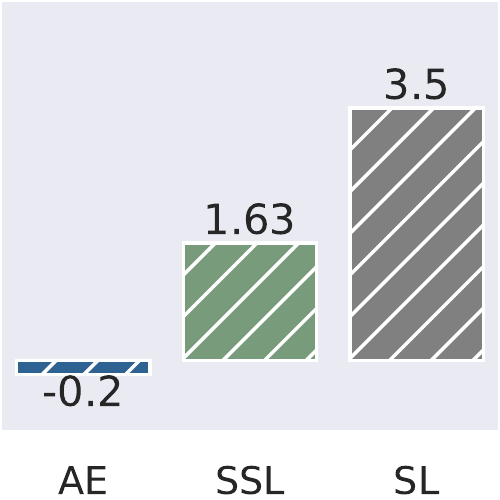}
      \caption{\scriptsize Camelyon17}\label{fig:camelyon_summary}
    \end{subfigure}
    \begin{subfigure}[t]{0.16\linewidth}
      \centering
      \includegraphics[width=\linewidth]{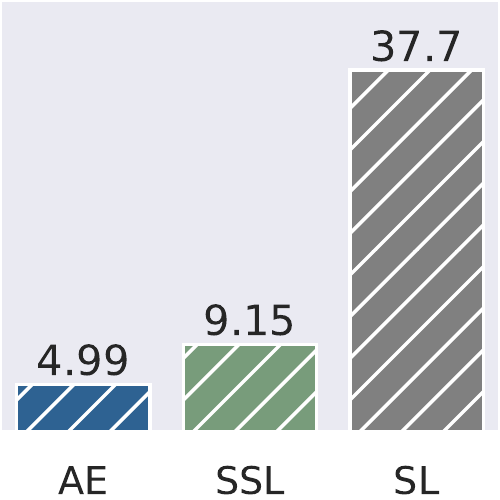}
      \caption{\scriptsize FMoW}\label{fig:fmow_summary}
    \end{subfigure} 
    
    \vspace{-1em}
    \caption{Performance of auto-encoder (AE), \acrfull{SSL},
    \acrfull{SL} models. \textbf{\textit{Top row}}: the OOD test set
    accuracy (\%) using linear heads trained on OOD data;
    \textbf{\textit{Bottom row}}: Shift Sensitivity (see \cref{sec:setup} for definition), measures models' sensitivity to distribution shift. Note here $r_{\text{id}}=1$ for
    Camelyon17-CS, FMoW-CS, and CdSprites~(see~\cref{sec:wilds_controllable,subsec:cdsprites_experiment}).\looseness=-1}
    \label{fig:summary}
    \vspace{-12pt}
  \end{figure*}
  
  \emph{\textbf{Contribution}: Controllable but realistic distribution
  shift tasks.} In addition to evaluating models on both synthetic and
  realistic datasets, we subsample realistic domain generalisation
  datasets in WILDS to artificially inject explicit spurious
  correlations between domains and labels. 
  This allows us to directly control the level of shift in these
  realistic distribution shift datasets.
  We refer to them as \emph{Controllable-Shift~(CS) datasets}.
  %
%   The resulting datasets utilise both real-world data and has
%   distribution shift that is controllable via the correlation
%   coefficient between domain and label information.
  %
  \footnote{While datasets like ImageNet-9 and SVSF also provides
  distribution shift controls for real images, our sub-sampling of
  realistic domain generalisation datasets allows us to gain such
  control {\em cheaply} on a {\em large} set of datasets, and in
  addition provide further analysis on models' performance on {\em
  existing benchmarks} (WILDS).}
  
  \vspace{-8pt}
  \paragraph{Problem 3: The linear classifier head strongly biases the evaluation protocol}
  The most popular evaluation protocol for representation learning
  algorithms (both supervised and unsupervised) is \emph{linear
  probing}. This involves freezing the weights of the representation
  learning model and training a linear classifier head on top of that
  to predict the labels.
  For distribution shift problems, this linear classifier is typically trained on the same training set as the representation learning backbone.
  \citet{kang2019decoupling, menon2020overparameterisation} observed
  the interesting phenomenon that this final layer linear classifier
  can be extremely susceptible to spurious correlations, causing poor
  OOD performance. Under simple, synthetic set up, they showed that SL
  models' performance on OOD test set can be dramatically improved by
  simply retraining the linear classifier on data where the spurious
  correlation is absent.
  This indicates that the linear classifier can be a strong source of
  model bias in distribution shift tasks, and to disentangle the
  linear classifier bias from the generalisation performance of the
  learned representations, it is advisable to re-train the linear head
  on OOD data during evaluation.
  Note that although retraining linear classifier head is already standard practice in transfer learning, its application is necessary as the pre-training task and target task are typically different; on the other hand, retraining linear head is neither necessary nor standard practice in distribution shift problems, despite the recognition of linear head bias in recent work \citep{kang2019decoupling, menon2020overparameterisation}.
  
  \emph{\textbf{Contribution}: OOD linear head.} When reporting OOD
  accuracy, we use a linear head trained on small amount of left-out
  \gls{OOD} data as opposed to \gls{ID} data, as is standard practice.
  This allows us to isolate the bias of the linear head from the
  generalisability of learned representations.
  We also quantify the linear head bias to highlight the importance of this treatment.
  With these results, we wish to establish OOD linear head evaluation
  as a standard protocol for evaluating robustness of representation
  learning algorithms to distribution shift.

  In summary, we develop a suite of experiments and datasets to
  evaluate the performance of various representation learning
  paradigms under distribution shift.~\Cref{fig:summary} provides a
  summary of our results, comparing a range of methods from the
  following classes of algorithms:
  \begin{enumerate*}[label=(\roman*)]
  \item \gls{SSL},
  \item \gls{AE}, and 
  \item \gls{SL}.
  \end{enumerate*}
  Note that though the intuition that unsupervised objectives should be better at distribution shift tasks is well-established in theory \citep{chen2020simclr,vib}, state-of-the-art methods are predominantly developed under SL. To the best of our knowledge, we are the first to systematically evaluate and compare unsupervised representation learning methods to SL under distribution shift.
  The models are evaluated on both synthetic and realistic
  distribution shift datasets. Further, the models are also evaluated
  on CS datasets that contains controllable, explicit spurious
  correlations in realistic datasets.
  The main takeaways from this paper are:\looseness=-1

  \begin{itemize}[nosep,leftmargin=1em,labelwidth=*,align=left]
     \item \textbf{SSL and AE are more robust than SL to extreme
     distribution shift:}~\Cref{fig:ms_summary,fig:cdsprites_summary,fig:camelyon_c_summary,fig:fmow_c_summary}
     shows results on distribution shift scenarios where the training
     set encodes extreme spurious correlations. In this setting, for
     both synthetic (\cref{fig:ms_summary,fig:cdsprites_summary}) and
     real world (\cref{fig:camelyon_c_summary,fig:fmow_c_summary})
     datasets, SSL and AE consistently outperforms SL in terms of OOD
     accuracy (top row); 
     \item \textbf{Compared to SL, SSL and AE's performance drop less under distribution shift}: The bottom row of~\cref{fig:summary}
     compares the shift sensitivity ($s$) of different models 
     (see \cref{sec:setup} for definition). Smaller
     $s$ is desirable as it indicates lower sensitivity to
     distribution shift. Results show that SSL and AE algorithms are
     significantly more stable under distribution shift than SL;
    %  \item \textbf{SSL and AE's performance is less sensitive to
    %  distribution shift than SL}: The bottom row of~\Cref{fig:summary}
    %  shows the difference between ID and OOD accuracy (i.e. shift
    %  sensitivity $s$; see \cref{sec:setup} for definition). Smaller
    %  $s$ is desirable as it indicates lower sensitivity to
    %  distribution shift. Results show that SSL and AE algorithms are
    %  significantly more stable under distribution shift  than SL;
     \item \textbf{Generalisation performance on distribution shift
     tasks can be significantly improved by retraining the linear
     head:} We show a large performance boost for all models, when
     evaluated using linear head trained on a small amount of OOD
     data, in contrast to the baseline linear head trained on ID data.
     The surprising gain of this cheap procedure, even on realistic problems, highlights the importance of
     isolating the linear head bias when evaluating generalisation
     performance.\looseness=-1
  \end{itemize}
\section{Seting Up} \label{sec:setup}

In \cref{sec:introduction} we identified three problems in the
existing literature that we wish to address in this work.
In this section, we will introduce the necessary experimental set-up in further details. In brief, we compare eight ML
algorithms on six datasets using three relevant metrics.

\textbf{Algorithms:} \emph{$\times$3 SSL, $\times$4 AE, $\times$1 SL.}
We compare seven unsupervised representation learning algorithms against SL, including three \gls{SSL} algorithms 1) SimCLR \citep{chen2020simclr}, 2) SimSiam \citep{chen2021simsiam}, and 3)
BYOL \citep{grill2020byol}; and four \gls{AE} algorithms 1)
Autoencoder \citep{rumelhart1985ae}, 2) \gls{VAE} \citep{kingma2013vae}, 3) $\beta$-\gls{VAE} \citep{higgins2016betavae} and 4) \gls{IWAE} \citep{burda2015iwae}.
These popular methods in \gls{SSL} and latent generative models have
not yet been systematically evaluated under distribution shift tasks
prior to our work.
We compare the performance of these models against a standard
supervised learning (SL) algorithm used as a representation learning
model.

\textbf{Datasets:} \emph{$\times$2 synthetic, $\times$2 realistic,
$\times$2 controllable shift.}
We evaluate our models on two synthetic datasets, namely MNIST-CIFAR~(\citet{shah2020pitfalls}; see \cref{subsec:mnist_cifar}) and CdSprites~(\citet{fish}; see~\cref{subsec:cdsprites_experiment}), as well as two
realistic datasets from WILDS~\citep{koh2021wilds}:
Camelyon17 and FMoW (see~\cref{sec:wilds_original}).
However, as mentioned in~\cref{sec:background}, both the synthetic and
the realistic datasets have their own drawbacks. To further
understand the models' performance and draw conclusions that are
generalisable, we also provide a framework
for creating {\em controllable shift} datasets from realistic datasets
like those in WILDS, by subsampling the data to introduce spurious correlations between the domain and label information in the training set.
Changing this correlation varies the degree of distribution shift
between the~(ID) train and~(OOD) test split, which allows us to
analyse the models' performance more effectively under realistic, yet
controllable, distribution shift.
We refer to this controllable shift versions of the two datasets
Camelyon17-CS and FMoW-CS, and provide further details on the datasets
in~\cref{sec:wilds_controllable}.

\textbf{Evaluation:} \emph{3 metrics.}
%
% We use three metrics to understand various facets of this problem.
Before discussing our proposed metrics, we first define some necessary
notations. 
We separate a model trained to perform a classification task into two
parts, namely, 1) backbone $f$, denoting the part of the model that
generates representations from the data, and 2) final linear head $c$,
which takes the representations from $f$ and outputs the final
prediction.
Further, we refer to the final linear head trained on representations
from the ID train set as $c_i$, and that trained on representations
from the OOD test set as $c_o$.
Since the backbone $f$ is always trained on the ID train set,
we do not make any notation distinction on its training distribution. 
We also denote the accuracy of $f$ and $c$ on the ID test data as
acc$_i(f,c)$, and on the OOD test data as acc$_o(f,c)$.\looseness=-1

As noted in \cref{sec:background}, we report the OOD accuracy of the
algorithms using linear heads trained on OOD data (instead of those
trained on ID data as per standard practice), i.e. \aoco. This is
necessary to disentangle the bias of the linear head from that of the
representations.
To highlight the importance of this treatment in isolating the
generalisability of the representation learning algorithm from the
that of the linear head, we also define the \emph{linear head bias}.
It is the difference between the OOD test accuracy evaluated by OOD
linear head and that evaluated by the ID linear head, i.e.
\begin{align}
    b = \aoco - \aoci. \label{eq:b}
\end{align}

In a related work,~\citet{taori2020measuring} proposed to evaluate the \emph{effective robustness} of OOD generalisation defined as $\rho = \aici - \aoci$, which quantifies the drop in performance (e.g. accuracy) when evaluating the model on OOD test set vs. ID test set.
A small $\rho$ is desirable, as it indicates that the performance of
the model is relatively insensitive to a distribution shift\footnote{We negate the original definition of effective
robustness from~\citet{taori2020measuring} for ease of
understanding.}.
However, we note that a simple decomposition of \emph{effective
robustness}~(\(\rho\)) shows a hidden {\em linear head bias}~(\(b\))
term
\vspace{3pt}
\begin{align}
    \underbrace{\aici - \aoci}_{\text{effective robustness}~\rho} &= \aici - \aoci  - \aoco + \aoco \notag \\
    &= \underbrace{\aoco - \aoci}_{\text{linear head bias}~b} + \underbrace{\aici - \aoco}_{\text{shift sensitivity}~s}. \label{eq:s}
\end{align}
%
% Again, since we are primarily interested in the robustness of the
% representations of the backbone model $f$ and not the classifier head
% $c$, it is important to isolate the linear head bias from
% effective robustness. 

Thus, we remove the effect of the linear head bias by subtracting
\(b\) from \(\rho\) and reporting the last term in \cref{eq:s}. We
refer to this as {\em shift sensitivity}~$:~s=\rho- b$.
Alternatively, it is the difference between the OOD accuracy using
linear head trained on OOD data, and ID accuracy using linear head
trained on ID data.
Larger $s$ marks higher sensitivity of $f$ to distribution shift,
which is, possibly, dangerous for the deployment of such models.
In summary, for each experiment we report the following three metrics:
OOD linear head accuracy $\aoco$, linear head bias $b$ and shift
sensitivity $s$.

\ifthenelse{\equal{\paperversion}{arxiv}}
{
\section{Experimental results} \label{sec:experiments}
We perform a hyperparameter search on learning rate, scheduler,
optimiser, representation size, etc. for each model. We use the standard SSL augmentations proposed in \citet{he2020moco,chen2020improved} for all models to ensure a fair comparison.
See \cref{sec:app_hp} for details.\looseness=-1 
\subsection{Synthetic distribution shift}  \label{sec:simplicity}
In this section, we evaluate the performance of \gls{SL}, \gls{SSL} and \gls{AE} algorithms on synthetic distribution shift tasks, utilising the MNIST-CIFAR dataset \citep{shah2020pitfalls} and the CdSprites dataset \citep{fish}.
All results are averaged over 5 random seeds.

\subsubsection{MNIST-CIFAR} \label{subsec:mnist_cifar}
\textbf{\emph{Finding:}} \emph{Under this extreme distribution shift setting, \gls{SSL} and \gls{AE} significantly outperform \gls{SL}. The OOD accuracy of SSL and AE can be notably improved by retraining the linear head on OOD data, however the OOD accuracy of SL remains low even with the OOD-trained linear head.}

% \begin{wrapfigure}[10]{r}{0.2\linewidth}
% \vspace{-1.8em}
% \begin{subfigure}{0.99\linewidth}
%     \includegraphics[width=0.95\linewidth]{images/mnist_cifar/ms_id.png}
%     \vspace{-4pt}
%     \subcaption{ID}\label{fig:ms_id_vis}
% \end{subfigure}\\
% \begin{subfigure}{0.99\linewidth}
%     \includegraphics[width=0.95\linewidth]{images/mnist_cifar/ms_ood.png}
%     \vspace{-4pt}
%     \subcaption{OOD}\label{fig:ms_ood_vis}
%     % \vspace{-10pt}
% \end{subfigure}
% \vspace{-0.8em}
% \caption{MNIST-CIFAR dataset.}\label{fig:ms}
% \end{wrapfigure}
%
The MNIST-CIFAR dataset consists of concatenations of images from two
classes of MNIST and CIFAR-10. 
%
% It contains the following splits with varying correlation between the
% labels of the MNIST and CIFAR-10 images 
In each concatenated image, the classes of the two datasets are either
correlated or uncorrelated depending on the split as discussed
below~(See \cref{fig:synthetic_v_real}, \textit{MNIST-CIFAR} for an
example):

\begin{itemize}[nosep,leftmargin=1em,labelwidth=*,align=left]
    \item \textbf{ID train, test}: Correlation between MNIST and
    CIFAR-10 labels is one. Each image belongs to one of the two
    classes: 1) MNIST ``\texttt{0}'' and CIFAR-10
    ``\texttt{automobile}'', and 2) MNIST ``\texttt{1}'' and CIFAR-10
    ``\texttt{plane}'' (\Cref{fig:synthetic_v_real}, \textit{top
    row});
    \item \textbf{OOD train, test}: Zero correlation between MNIST and CIFAR-10 labels, images from the two classes are randomly paired (\Cref{fig:synthetic_v_real}, \textit{bottom row}).
\end{itemize}

Since the MNIST features are much simpler than the CIFAR features, a model trained on the ID train set can use MNIST only to predict the label, even though the CIFAR images are just as predictive \citep{shah2020pitfalls}.
This results in poor performance when predicting the CIFAR label on the OOD test set, where there is no correlation between the MNIST and CIFAR labels.

We train a CNN backbone on the ID train set using the eight \gls{SL}, \gls{SSL} and \gls{AE} algorithms listed in \cref{sec:setup}. At test time, we freeze the backbone and train two linear heads on ID train and OOD train set respectively, and evaluate their performance on the ID and OOD test set to compute 1) OOD linear head accuracy \aoco, 2) shift sensitivity $s$ and, 3) linear head bias $b$.
See results in \cref{tab:mnist_cifar_results}.

We observe that all models achieve near perfect performance when predicting the MNIST label on OOD test set, all with low shift sensitivity and small linear head bias.
However, when predicting the labels of the more complex CIFAR images, unsupervised algorithms have a clear advantage over the supervised one:  SSL achieves the highest OOD accuracy at $86.1\%$, followed by AE at $79.9\%$ and SL at $51.5\%$ (near random).
The shift sensitivity $s$ of the three objectives follow a similar
trend, with SSL and AE scoring significantly lower than SL. This
indicates that unsupervised representations are significantly
less sensitive to distribution shift compared to those from SL, with the latter suffering a drop as large as $47.6\%$.
Interestingly, the classifier head bias $b$ for SSL and AE are
relatively high (around $30\%$), and is very low for SL ($0.8\%$),
indicating that the representations learned from SL is intrinsically un-robust to distribution shift.
That is, while there exist (linearly separable) CIFAR features in
the representations of SSL and AE that can be extracted using a linear
head trained on un-biased (OOD) data, these features are absent from
the representations of SL.

\begin{table}[t]
\centering
\caption{Evaluations on the MNIST-CIFAR dataset. We report accuracy on MNIST and CIFAR trained using OOD linear head ($\text{acc}_o(f,c_o)$), linear head bias ($b$) and shift sensitivity ($s$).}
\vspace*{-0.8\baselineskip}
\scalebox{0.78}{
\begin{tabular}{llccccccc}    \toprule
    \multirow{2}{*}{Regime} &\multirow{2}{*}{Method} & \multicolumn{3}{c}{MNIST (\%)} & & \multicolumn{3}{c}{CIFAR (\%)}  \\
    \cmidrule{3-5} \cmidrule{7-9}
    & &  $\text{acc}_o(f,c_o)\uparrow$ & $s \downarrow$ & $b$ && $\text{acc}_o(f,c_o)\uparrow$ & $s\downarrow$  & $b$ \\
    \midrule \rowcolor{lb}
    & AE  & 99.9 \std{1e-2} & 0.0 \std{1e-2} & 0.0 \std{2e-3} && 81.1 \std{1e+0} & 18.8 \std{1e+0} & 30.2 \std{1e+0}\\  \rowcolor{lb}
    & VAE  & 99.8 \std{8e-3} & -0.1 \std{9e-3} & 0.5 \std{1e-4} && 79.7 \std{4e+0} & 20.2  \std{3e+0} & 29.2  \std{6e+0}\\  \rowcolor{lb}
    & IWAE  & 99.8 \std{9e-3} & 0.0 \std{4e-3} & 0.1 \std{5e-3} && 80.8 \std{2e+0} & 19.0 \std{3e+0} & 30.0  \std{4e+0}\\  \rowcolor{lb}
    \multirow{-4}{*}{AE}
    & $\beta$-VAE  & 99.8 \std{2e-2} & 0.0 \std{4e-2} & -0.1 \std{3e-2} && 78.0 \std{3e+0} & 21.8  \std{4e+0} & 28.0  \std{4e+0}\\  \rowcolor{db}
    \multicolumn{2}{l}{\emph{\textbf{AE average}}} & \emph{\textbf{99.8 \std{1e-2}}}  & \emph{\textbf{0.0 \std{1e-2}}}  & \emph{0.1 \std{9e-3}} && \emph{79.9 \std{3e+0}}  & \emph{20.0  \std{4e+0}} & \emph{29.3  \std{4e+0}} \\
    \midrule \rowcolor{lg}
    & SimCLR  & 99.7 \std{1e-2} & 0.2 \std{1e-3} & -0.2 \std{3e-3} && 85.8 \std{1e+0} & 14.1 \std{2e+0} & 35.5  \std{1e+0}\\ \rowcolor{lg}
    & SimSiam  & 99.8 \std{2e-1} & 0.1 \std{2e-1} & 0.0 \std{9e-2} && 87.8 \std{2e+0} & 12.1  \std{2e+0} & 35.6  \std{4e+0}\\  \rowcolor{lg}
    \multirow{-3}{*}{SSL} 
    & BYOL  & 99.8 \std{4e-2} & 0.0 \std{1e-2} & 0.9 \std{8e-3} && 84.8 \std{9e-1} & 15.0  \std{1e+0} & 33.2  \std{1e+0}\\  \rowcolor{dg}
    \multicolumn{2}{l}{\emph{\textbf{SSL average}}} & \emph{\textbf{99.8 \std{8e-2}}} & \emph{0.1 \std{5e-2}} & \emph{0.2 \std{3e-2}} && \emph{\textbf{86.1 \std{2e+0}}} & \emph{\textbf{13.7 \std{2e+0}}} & \emph{34.8 \std{4e+0}} \\
    \midrule \rowcolor{g}
    SL
    & Supervised  & 97.7 \std{9e-1} & 1.4 \std{1e+0} & -0.3 \std{1e+0} && 51.5 \std{1e+0} & 47.6  \std{1e+0} & 0.8  \std{9e-1}\\ 
    \bottomrule
\end{tabular}
\label{tab:mnist_cifar_results}}
\vspace*{-\baselineskip}
\end{table}

\subsubsection{CdSprites}  \label{subsec:cdsprites_experiment}

% \begin{wrapfigure}[10]{r}{0.3\linewidth}
%      \centering
%      \vspace{-1.5em}
%      \begin{subfigure}[b]{0.45\linewidth}
%          \centering
%          \includegraphics[width=\linewidth]{images/cdsprites/cdsprites_example_corr100.png}
%          \caption{$r = 1$}
%          \label{subfig:cdsprites_example_corr100}
%      \end{subfigure}
%      \begin{subfigure}[b]{0.45\linewidth}
%          \centering
%          \includegraphics[width=\linewidth]{images/cdsprites/cdsprites_example_corr0.png}
%          \caption{$r = 0$}
%          \label{subfig:cdsprites_example_corr0}
%      \end{subfigure}
%      \caption{CdSprites dataset. Each subplot shows 16 samples from the dataset.}
%      \label{fig:cdsprites_example}
%      \vspace{1.0em}
% \end{wrapfigure}

\textbf{\emph{Finding:}} \emph{Similar to MNIST-CIFAR, under extreme
distribution shift,~\gls{SSL} and \gls{AE} are better than
\gls{SL}; when the shift is less extreme, \gls{SSL} and \gls{SL}
achieve comparably strong OOD generalisation performance while
\gls{AE}'s performance is much weaker.}

%  When the shift is less extreme, \gls{SSL} and \gls{SL}
% achieve comparable strong OOD generalisation performance while
% \gls{AE}'s performance is much weaker.}

CdSprites is a colored variant of the popular dSprites dataset \citep{dsprites}, which consists of images of 2D sprites that are procedurally generated from multiple latent factors. 
The CdSprites dataset induces a spurious correlation between the color
and shape of the sprites, by coloring the sprites conditioned on the
shape following a controllable correlation coefficient
$r_{\text{id}}$.
See \cref{fig:synthetic_v_real} for an example: when $r_{\text{id}}=1$ color is completely dependent on shape (\textit{top row}, oval-purple, heart-cyan, square-white), and when $r_{\text{id}}=0$, color and shape are randomly matched (\textit{bottom row}).

\citet{fish} observes that when $r_{\text{id}}$ is high, SL model tend to use color only to predict the label while ignoring shape features due to the texture bias of CNN \citep{texture1, texture2}.
First, we consider the setting of extreme distribution shift similar
to MNIST-CIFAR by setting \(r_{\text{id}}=1\) in the ID train and test
splits. In the OOD train and test splits, the correlation coefficient
is set to zero to investigate how well the model learns both the shape
and the color features.~\Cref{tab:cdsprites_results} reports the three
metrics of interest using the same evaluation protocol as before.

Similar to MNIST-CIFAR, we observe that all models achieve near
perfect performance when predicting the simpler feature, i.e. color on
the OOD test set. However, when predicting shape, the more complex
feature on the OOD test set, SSL~(and also AEs to a lesser extent) is
far superior to SL. Additionally, the shift sensitivity of SSL~(and AE
to a lesser extent) are much smaller than SL, indicating that SSL/AE
models are more robust to extreme distribution shift. The linear head
bias also follows a similar trend as for MNIST-CIFAR, showing that
representations learned using SL methods are inherently not robust to
spurious correlations. This is not the case for SSL and AE algorithms
where a large linear head bias shows that is the ID linear heads and
not the representations that injects the bias.\looseness=-1
\paragraph{Controllable distribution shift}

We extend this experiment to probe the performance of these algorithms
under \emph{varying} degrees of distribution shifts. We generate three
versions of the CdSprites dataset with three different correlation
coefficients $r_{\text{id}} \in \{0, 0.5, 1\}$ of the ID train set. 
As before, the correlation coefficient of the OOD split is set to
zero%
%\footnote{Note that when $r_{\text{id}} = 0$, there is no
% distribution shift between the ID and OOD splits.}. 
The rest of the experimental protocol stays the same.
%
% Similar to MNIST-CIFAR, each version of CdSprites contain 4 splits, including 1) ID train, 2) ID test with $r_{\text{id}}$, and 3) OOD train 4) OOD test with $r_{\text{ood}}$.
% %
% We use the same evaluation protocol as for MNIST-CIFAR to acquire the OOD linear head accuracy \aoco, shift sensitivity $s$ and linear head bias $b$ on the three $r_{\text{id}}$ versions of the dataset.
%
The OOD test accuracy and the shift sensitivity for varying
\(r_{\text{id}}\) is plotted in~\cref{fig:controllable_results} and a
detailed breakdown of results is available in
\cref{app:additional_results}.\looseness=-1

\Cref{fig:controllable_results} shows that despite increasing
distribution shift between the ID and OOD splits (with increasing
$r_{\text{id}}$) the OOD performance of SSL and AE does not suffer.
However, the OOD accuracy of SL plummets and its shift sensitivity
explodes at \(r_{\text{id}}=1\).
%
% This highlights SL's weakness in handling larger distribution shift
% scenarios.
%
Interestingly, SSL maintains a high OOD test accuracy regardless of
the level of distribution shift: when $r_{\text{id}}<1$ its
performance is on par with SL, and when the distribution shift becomes
extreme with $r_{\text{id}}=1$ it significantly outperforms SL both in
terms of accuracy and shift sensitivity.
In comparison, AE models' accuracy lingers around $50\%$, with increasingly higher shift sensitivity as $r_{\text{id}}$ increases.
However, under extreme distribution shift with $r_{\text{id}}=1$ it
still performs better than SL, with slightly higher OOD accuracy and
lower shift sensitivity.\looseness=-1
%
% In terms of the linear head bias, we see a continuous increase for the AE as the distribution shift increases, whereas for SSL and SL the linear head bias is zero for $r_{\text{id}}<1$ and it peaks for all methods at $r_{\text{id}}=1$.

\begin{table}[t]
\vspace{-30pt}
  \centering
  \caption{Evaluations on the CdSprites dataset with $r_{\text{id}}=1.0$. We report accuracy for color and shape classifiers trained using OOD linear head ($\text{acc}_o(f,c_o)$), linear head bias ($b$) and shift sensitivity ($s$).}
  % \vspace*{-0.8\baselineskip}
  \scalebox{0.78}{
  \begin{tabular}{llccccccc}    \toprule
      \multirow{2}{*}{Regime} &\multirow{2}{*}{Method} & \multicolumn{3}{c}{Color classification (\%)} & & \multicolumn{3}{c}{Shape classification (\%)}  \\
      \cmidrule{3-5} \cmidrule{7-9}
      & &  $\text{acc}_o(f,c_o)\uparrow$ & $s \downarrow$ & $b$ && $\text{acc}_o(f,c_o)\uparrow$ & $s\downarrow$  & $b$ \\
      \midrule \rowcolor{lb}
      & AE  & 100.0 \std{0e+0} & 0.0 \std{2e-3} & 0.3 \std{5e-1} && 46.1 \std{6e-1}  &  53.9 \std{6e-1}  & 12.7 \std{5e-1} \\  \rowcolor{lb}
      & VAE  & 99.7 \std{3e-1} & 0.3 \std{3e-1} & -0.3 \std{3e-1} && 52.4 \std{2e+0}  &  47.6 \std{2e+0}   & 18.9 \std{3e+0}  \\  \rowcolor{lb}
      \multirow{-3}{*}{AE}& IWAE  & 100.0 \std{0e+0} & 0.0 \std{2e-3} & 0.4 \std{5e-1} && 58.9 \std{2e+0}  &  41.1 \std{2e+0}  & 25.6 \std{2e+0}  \\
      % & $\beta$-VAE  &  &  &  &&  &  & \\ 
      \rowcolor{db}
      \multicolumn{2}{l}{\emph{\textbf{AE average}}} & \emph{99.9 \std{9e-1}}  & \emph{0.1 \std{9e-2}}  & \emph{0.1 \std{4e-1}} && \emph{52.5 \std{2e+0}}  & \emph{ 47.5 \std{2e+0} } & \emph{19.1 \std{2e+0} } \\
      \midrule \rowcolor{lg}
      & SimCLR  & 100.0 \std{0e+0} & 0.0 \std{0e+0} & 0.0 \std{1e-1} && 87.8 \std{5e-1} &  12.2 \std{5e-1}  & 54.5 \std{5e-1}  \\ \rowcolor{lg}
      & SimSiam  & 100.0 \std{0e+0} & 0.0 \std{0e+0} & 0.1 \std{1e-1}  && 69.2 \std{2e+0}  &  30.8 \std{2e+0}  & 35.6 \std{2e+0}  \\  \rowcolor{lg}
      \multirow{-3}{*}{SSL} 
      & BYOL  & 100.0 \std{0e+0} & 0.0 \std{0e+0} & 0.0 \std{0e+0} && 91.1 \std{4e+0}  &  8.9 \std{4e+0}   & 57.9 \std{4e+0}  \\  \rowcolor{dg}
      \multicolumn{2}{l}{\emph{\textbf{SSL average}}} & \emph{100.0 \std{0e+0}} & \emph{0.0 \std{0e+0}} & \emph{0.1 \std{3e-2}} && \emph{82.7 \std{2e+0} } & \emph{17.3 \std{2e+0} } & \emph{49.3 \std{4e+0} } \\
      \midrule \rowcolor{g}
      SL
      & Supervised  & 100.0 \std{0e+0} & 0.0 \std{0e+0} & 0.0 \std{3e-2} && 44.0 \std{7e-1} & 56.0 \std{7e-1}   & 10.7 \std{7e-1}  \\ 
      \bottomrule
  \end{tabular}
  \label{tab:cdsprites_results}}
   \vspace*{-10pt}
  \end{table}

\subsection{Real-world distribution shift}  \label{sec:dg} 
In this section we investigate the performance of different objectives on real-world distribution shift tasks. 
We use two datasets from WILDS~\citep{koh2021wilds}: 1) Camelyon17, which contains tissue scans acquired from different hospitals, and the task is to determine if a given patch contains breast cancer tissue; and 2) FMoW, which features satellite images of landscapes on five different continents, with the classification target as the type of infrastructure. See examples in \Cref{fig:synthetic_v_real}.
% %
Following the guidelines from WILDS benchmark, we perform 10 random seed runs for all Camelyon17 experiment and 3 random seed runs for FMoW. The error margin in \Cref{fig:controllable_results} represent standard deviation.

\subsubsection{Original WILDS Datasets} \label{sec:wilds_original}
\textbf{\emph{Findings:}} \emph{SL is significantly more sensitive to
distribution shift than SSL and AE; representations from SSL obtain
higher OOD accuracy than SL on Camelyon17 but lower on FMoW. AE is
consistently the least sensitive to distribution shift though it has
the lowest accuracy. The performance of all models significantly
improves by retraining the linear head on a small amount of OOD
data.\looseness=-1}

The original Camelyon17 and FMoW dataset from WILDS benchmark both contains the following three splits: ID train, OOD validation and OOD test.
We further create five splits specified as follows:
\begin{itemize}[nosep,leftmargin=1em,labelwidth=*,align=left]
    \item \textbf{ID train, test}: Contains 90\% and 10\% of the original ID train split, respectively;
    \item \textbf{OOD train, test}: Contains 10\% and 90\% of the original OOD test split, respectively;
    \item \textbf{OOD validation}: Same as the original OOD validation split.
\end{itemize}
Following WILDS, we use OOD validation set to perform early stopping and choose hyperparameters; we also use DenseNet-121 \citep{dense} as the backbone for all models.
We follow similar evaluation protocol as previous experiments, and in addition adopt 10-fold cross-validation for the OOD train and test set.
% Similar to our previous experiments, after training the backbone model on the ID train set, we freeze the backbone and train two linear heads on ID train and OOD train respectively, and evaluate all models on the ID and OOD test set to compute our results.
%
See results in \Cref{tab:camelyon17_results,tab:fmow_results}, where
following WILDS, we report performance on Camelyon17 using standard
average accuracy and on FMoW using worst-group accuracy.

One immediate observation is that in contrast to our previous experiments on synthetic datasets, SL's OOD accuracy is much higher in comparison on realistic distribution shift tasks: it is the best performing model on FMoW with $35.6\%$ worst-group accuracy on OOD test set;
its OOD accuracy is the lowest on Camelyon17, however it is only $3\%$
worse than the highest accuracy achieved by SSL ($89.8\%$). This
highlights the need to study realistic datasets along with synthetic ones.
Nonetheless, we find that SSL is still the best performing method on
Camelyon17 and achieves competitive performance on FMoW with accuracy
$29.6\%$ --- despite learning without labels!
AE has much lower OOD accuracy on FMoW compared to the other two methods: we believe this is due to its reconstruction-based objective wasting modelling capacity on high frequency details, a phenomenon frequently observed in prior work \citep{bao2021beit,ramesh2021zero}.
Note that the standard deviation for all three methods are quite high for Camelyon17: this is a known property of the dataset and similar pattern is observed across most methods on WILDS benchmark \citep{koh2021wilds}.

In terms of shift sensitivity, unsupervised objectives including SSL
and AE consistently outperforms SL --- this stands out the most on
FMoW, where the shift sensitivity of SSL and AE are $9.1\%$ and
$5.0\%$ respectively, while SL is as high as $37.7\%$. 
This observation further validates our previous finding on synthetic
datasets, that SSL and AE's ID accuracy is a relatively reliable
indication of their generalisation performance, while SL can undergo a
huge performance drop under distribution shift, which can be dangerous
for the deployment of such models. We highlight that, in sensitive
application domains, a low shift sensitivity is an important criterion
as it implies that the model's performance will remain consistent when
the distribution shifts.
Another interesting observation here is that for all objectives on both datasets, the classifier bias $b$ is consistently high.
This indicates the bias of the linear classification head plays a significant role even for real world distribution shifts, and that it is possible to mitigate this effect by training the linear head using a small amount of OOD data (in this case $10\%$ of the original OOD test set).

\begin{table}[t]
\parbox{.48\linewidth}{
\centering
\caption{Evaluations on test set of Camelyon17, all metrics computed using average accuracy.}
\vspace*{-0.8\baselineskip}
\scalebox{0.7}{
\begin{tabular}{llccc}    \toprule
    \multirow{2}{*}{Regime} &\multirow{2}{*}{Method} & \multicolumn{3}{c}{Metrics (\%)} \\
    \cmidrule{3-5} 
    & &  $\text{acc}_o(f,c_o)\uparrow$ & $s \downarrow$ & $b$ \\
    \midrule \rowcolor{lb}
    & AE          & 84.4 \std{2e+0} & -0.6 \std{1e+0}  &12.7 \std{2e+0} \\  \rowcolor{lb}
    & VAE         & 88.1 \std{2e+0} & 0.5 \std{2e+0} &39.0 \std{2e+0} \\  \rowcolor{lb}
    & IWAE        & 88.1 \std{1e+0} & -0.9 \std{3e+0}  &39.1 \std{4e+0} \\  \rowcolor{lb}
    \multirow{-4}{*}{AE}
    & $\beta$-VAE & 87.1 \std{4e+0} &0.2 \std{4e+0} & 36.0 \std{5e+0} \\  \rowcolor{db}
    \multicolumn{2}{l}{\emph{\textbf{AE average}}} & \emph{86.9 \std{2e+0}}  & \emph{\textbf{-0.2 \std{3e+0}}}  & \emph{31.7 \std{3e+0}} \\
    \midrule \rowcolor{lg}
    & SimCLR      & 92.7 \std{2e+0} & 0.4 \std{1e+0} & 8.3 \std{1e+0} \\ \rowcolor{lg}
    & SimSiam     & 86.7 \std{1e+0} &3.1 \std{1e+0} & 7.9 \std{3e+0}\\  \rowcolor{lg}
    \multirow{-3}{*}{SSL} 
    & BYOL        & 89.9 \std{1e+0} &1.4 \std{1e+0} & 10.3 \std{2e+0}\\  \rowcolor{dg}
    \multicolumn{2}{l}{\emph{\textbf{SSL average}}} & \emph{\textbf{89.8 \std{1e+0}}} & \emph{1.6 \std{1e+0}} & \emph{8.8 \std{2e+0}}\\
    \midrule \rowcolor{g}
    SL
    & Supervised  & 86.8 \std{2e+0} & 3.5 \std{1e+0} &  7.4 \std{3e+0} \\ 
    \bottomrule
\end{tabular}
\label{tab:camelyon17_results}}
}
\hspace{10pt}
\parbox{.48\linewidth}{
\centering
\caption{Evaluations on test set of FMoW, all metrics computed using worst-group accuracy.}
\vspace*{-0.8\baselineskip}
\scalebox{0.7}{
\begin{tabular}{llccc}    \toprule
    \multirow{2}{*}{Regime} &\multirow{2}{*}{Method} & \multicolumn{3}{c}{Metrics (\%)} \\
    \cmidrule{3-5} 
    & &  $\text{acc}_o(f,c_o)\uparrow$ & $s \downarrow$ & $b$ \\
    \midrule \rowcolor{lb}
    & AE          & 26.9 \std{9e-3} &6.4 \std{6e-3}  & 5.8 \std{1e-2} \\  \rowcolor{lb}
    & VAE         & 21.7 \std{6e-3} & 4.7 \std{4e-3} & 8.0 \std{2e-2} \\  \rowcolor{lb}
    & IWAE        & 20.9 \std{2e-2} &5.5 \std{1e-2}  & 7.8 \std{1e-2} \\  \rowcolor{lb}
    \multirow{-4}{*}{AE}
    & $\beta$-VAE & 21.7 \std{5e-3} &3.4 \std{6e-3} & 7.6  \std{8e-3} \\  \rowcolor{db}
    \multicolumn{2}{l}{\emph{\textbf{AE average}}} & \emph{22.8 \std{3e-2}}  & \emph{\textbf{5.0  \std{7e-3}}}  & \emph{7.3  \std{1e-2}} \\
    \midrule \rowcolor{lg}
    & SimCLR      & 29.9 \std{6e-3} & 10.7 \std{6e-3} & 7.6 \std{7e-3} \\ \rowcolor{lg}
    & SimSiam     & 27.8 \std{2e-2} &4.6 \std{1e-2} & 6.3 \std{2e-2}\\  \rowcolor{lg}
    \multirow{-3}{*}{SSL} 
    & BYOL        & 31.3 \std{1e-2} &12.1 \std{7e-3} & 7.9  \std{1e-2}\\  \rowcolor{dg}
    \multicolumn{2}{l}{\emph{\textbf{SSL average}}} & \emph{29.6 \std{2e-2}} & \emph{9.1 \std{9e-3}} & \emph{7.3 \std{1e-2}}\\
    \midrule \rowcolor{g}
    SL
    & Supervised  & \textbf{35.6 \std{7e-3}} &37.7 \std{4e-2} &  6.9 \std{9e-3} \\ 
    \bottomrule
\end{tabular}
\label{tab:fmow_results}}
}
\vspace{-10pt}
\end{table}

\subsubsection{WILDS Datasets with Controllable Shift} \label{sec:wilds_controllable}

\textbf{\emph{Findings:}} \emph{SL's OOD accuracy drops as more the distribution shift becomes more challenging, with SSL being the best performing model when the distribution shift is the most extreme. The shift sensitivity of SSL and AE are consistently lower than SL regardless of the level of shift.}

To examine models' generalisation performance under different levels of distribution shift, we create versions of these realistic datasets with {\em controllable shifts}, which we name Camelyon17-CS and FMoW-CS.
Specifically, we subsample the ID train set of these datasets to artificially create spurious correlation between the domain and label.
For instance, given dataset with domain \texttt{A}, \texttt{B} and label \texttt{0}, \texttt{1}, to create a version of the dataset where the spurious correlation is 1 we would sample only examples with label \texttt{0} from domain \texttt{A} and label \texttt{1} from domain \texttt{B}.
See \Cref{app:controllable_datasets} for further details.

Similar to CdSprites, we create three versions of both of these
datasets with the spurious correlation coefficient $r_{\text{id}} \in
\{0, 0.5,1\}$ in ID~(train and test) sets.
The OOD train, test and validation set remains unchanged\footnote{Note
that even when $r_{\text{id}}=0$, distribution shift between the ID
and OOD splits exists, as the spurious correlation is not the only
source of the distribution shift.}.
Using identical experimental setup as in \cref{sec:wilds_original}, we
plot the trend of performance with increasing $r$ for Camelyon17-CS and FMoW-CS in
\Cref{fig:controllable_results} with detailed numerical breakdown in \Cref{tab:camelyon17corr50_results,tab:fmowcorr50_results,tab:camelyon17corr100_results,tab:fmowcorr100_results}.

\begin{table}
\parbox{.48\linewidth}{
\centering
\caption{Evaluations on test set of Camelyon17-C with $r_{\text{id}}=0.5$, all metrics computed using average accuracy.}
\vspace*{-0.8\baselineskip}
\scalebox{0.7}{
\begin{tabular}{llccc}    \toprule
    \multirow{2}{*}{Regime} &\multirow{2}{*}{Method} & \multicolumn{3}{c}{Metrics (\%)} \\
    \cmidrule{3-5} 
    & &  $\text{acc}_o(f,c_o)\uparrow$ & $s\downarrow$ & $b$ \\
    \midrule \rowcolor{lb}
    & AE          & 80.4	\std{3e+0} &	6.0	\std{2e+0} &	19.0	\std{4e+0} \\  \rowcolor{lb}
    & VAE         & 88.6	\std{2e+0} & -0.5	\std{1e+0} &	17.8	\std{6e+0} \\  \rowcolor{lb}
    & IWAE        & 87.8	\std{1e+0} & -0.2	\std{1e+0} &	26.4	\std{6e+0} \\  \rowcolor{lb}
    \multirow{-4}{*}{AE}
    & $\beta$-VAE & 88.5	\std{2e+0} &	0.1	\std{9e-1} &	19.7	\std{6e+0} \\  \rowcolor{db}
    \multicolumn{2}{l}{\emph{\textbf{AE average}}} & \emph{86.3	\std{2e+0}} &	\emph{1.4	\std{1e+0}} &	\emph{20.7	\std{5e+0}} \\
    \midrule \rowcolor{lg}
    & SimCLR      & 84.5	\std{2e+0} &	8.0	\std{1e+0} &	6.6 	\std{2e+0} \\ \rowcolor{lg}
    & SimSiam     & 86.1	\std{2e+0} &	5.7	\std{1e+0} &	8.3 	\std{4e+0}\\  \rowcolor{lg}
    \multirow{-3}{*}{SSL} 
    & BYOL        & 86.4	\std{2e+0} &	4.5	\std{2e+0} &	8.8 	\std{4e+0}\\  \rowcolor{dg}
    \multicolumn{2}{l}{\emph{\textbf{SSL average}}} & \emph{85.7	\std{2e+0}} &	\emph{6.1	\std{2e+0}} &	\emph{7.9 	\std{3e+0}} \\
    \midrule \rowcolor{g}
    SL
    & Supervised  & 81.5	\std{5e+0} & 13.2	\std{3e+0} &	3.4 	\std{4e+0}\\ 
    \bottomrule
\end{tabular}
\label{tab:camelyon17corr50_results}}
}
\hspace{10pt}
\parbox{.48\linewidth}{
\centering
\caption{Evaluations on test set of FMoW-C with $r_{\text{id}}=0.5$, all metrics computed using worst-group accuracy.}
\vspace*{-0.8\baselineskip}
\scalebox{0.7}{
\begin{tabular}{llccc}    \toprule
    \multirow{2}{*}{Regime} &\multirow{2}{*}{Method} & \multicolumn{3}{c}{Metrics (\%)} \\
    \cmidrule{3-5} 
    & &  $\text{acc}_o(f,c_o)\uparrow$ & $s\downarrow$ & $b$ \\
    \midrule \rowcolor{lb}
    & AE          &23.4 \std{1e+0} &	8.6  	\std{6e-1}	& 4.2	\std{8e-1} \\  \rowcolor{lb}
    & VAE         &18.7 \std{1e+0} &	7.7  	\std{6e-1}	& 1.5	\std{6e-1} \\  \rowcolor{lb}
    & IWAE        &18.5 \std{2e+0} &	7.3  	\std{2e+0}	& 2.2	\std{1e+0} \\  \rowcolor{lb}
    \multirow{-4}{*}{AE}
    & $\beta$-VAE &21.4 \std{3e-1} &	4.0  	\std{4e-1}	& 3.9	\std{7e-1} \\  \rowcolor{db}
    \multicolumn{2}{l}{\emph{\textbf{AE average}}} & \emph{20.5 \std{2e+0}} &	\emph{6.9  	\std{8e-1}}	& \emph{3.0	\std{9e-1}} \\
    \midrule \rowcolor{lg}
    & SimCLR      &29.5 \std{9e-1} &	9.2  	\std{6e-1}	& 6.9	\std{7e-1}\\ \rowcolor{lg}
    & SimSiam     &27.9 \std{2e+0} &	7.5  	\std{1e+0}	& 4.6	\std{1e+0}\\  \rowcolor{lg}
    \multirow{-3}{*}{SSL} 
    & BYOL        &32.6 \std{3e+0} &	7.9  	\std{2e+0}	& 8.5	\std{2e+0}\\  \rowcolor{dg}
    \multicolumn{2}{l}{\emph{\textbf{SSL average}}} & \emph{30.0 \std{2e+0}} &	\emph{8.2 \std{1e+0}} & \emph{6.7 \std{1e+0}}\\
    \midrule \rowcolor{g}
    SL
    & Supervised  & 32.3 \std{3e+0} &	25.1  	\std{3e+0}	& 6.0	\std{2e+0}\\ 
    \bottomrule
\end{tabular}
\label{tab:fmowcorr50_results}}
}
\end{table}

\begin{table}
\parbox{.48\linewidth}{
\centering
\caption{Evaluations on test set of Camelyon17-C with $r_{\text{id}}=1$, all metrics computed using average accuracy.}
\vspace*{-0.8\baselineskip}
\scalebox{0.7}{
\begin{tabular}{llccc}    \toprule
    \multirow{2}{*}{Regime} &\multirow{2}{*}{Method} & \multicolumn{3}{c}{Metrics (\%)} \\
    \cmidrule{3-5} 
    & &  $\text{acc}_o(f,c_o)\uparrow$ & $s\downarrow$ & $b$ \\
    \midrule \rowcolor{lb}
    & AE          &75.7	\std{5e+0}	& 7.3	 \std{2e+0} &	35.1	\std{4e+0} \\  \rowcolor{lb}
    & VAE         &86.0	\std{3e+0}	& 2.7	 \std{1e+0} &	12.4	\std{4e+0} \\  \rowcolor{lb}
    & IWAE        &86.1	\std{1e+0}	& 2.6	 \std{7e-1} &	9.1 	\std{3e+0} \\  \rowcolor{lb}
    \multirow{-4}{*}{AE}
    & $\beta$-VAE &84.7	\std{2e+0}	& 3.8	 \std{1e+0} &	15.5	\std{4e+0} \\  \rowcolor{db}
    \multicolumn{2}{l}{\emph{\textbf{AE average}}} & \emph{83.1	\std{3e+0}}	& \emph{4.1	 \std{1e+0}} &	\emph{18.0	\std{4e+0}}\\
    \midrule \rowcolor{lg}
    & SimCLR      &85.8	\std{8e+1}	& 2.8	 \std{4e-1} &	6.2 	\std{2e+0} \\ \rowcolor{lg}
    & SimSiam     &82.1	\std{1e+0}	& 6.0	 \std{7e-1} &	8.3 	\std{4e+0}\\  \rowcolor{lg}
    \multirow{-3}{*}{SSL} 
    & BYOL        &74.8	\std{5e+0}	& 10.3 \std{2e+0} &	-2.2	    \std{4e+0}\\  \rowcolor{dg}
    \multicolumn{2}{l}{\emph{\textbf{SSL average}}} &\emph{80.9	\std{2e+0}}	& \emph{6.3	 \std{1e+0}} &	\emph{4.1 	\std{3e+0}} \\
    \midrule \rowcolor{g}
    SL
    & Supervised  &73.4	\std{6e+0}	& 10.7 \std{3e+0} &	5.9 	    \std{8e+0}\\ 
    \bottomrule
\end{tabular}
\label{tab:camelyon17corr100_results}}
}
\hspace{10pt}
\parbox{.48\linewidth}{
\centering
\caption{Evaluations on test set of FMoW-C with $r_{\text{id}}=1$, all metrics computed using worst-group accuracy.}
\vspace*{-0.8\baselineskip}
\scalebox{0.7}{
\begin{tabular}{llccc}    \toprule
    \multirow{2}{*}{Regime} &\multirow{2}{*}{Method} & \multicolumn{3}{c}{Metrics (\%)} \\
    \cmidrule{3-5} 
    & &  $\text{acc}_o(f,c_o)\uparrow$ & $s\downarrow$ & $b$ \\
    \midrule \rowcolor{lb}
    & AE          &22.4	\std{1e+0} &   10.0	\std{6e-1} &	3.8	\std{7e-1} \\  \rowcolor{lb}
    & VAE         &16.6	\std{9e-1} &	8.6	\std{1e+0} &	2.8	\std{8e-1} \\  \rowcolor{lb}
    & IWAE        &17.2	\std{5e-1} &	8.7	\std{6e-1} &	3.8	\std{5e-1} \\  \rowcolor{lb}
    \multirow{-4}{*}{AE}
    & $\beta$-VAE &16.7	\std{3e-1} &	7.9	\std{4e-1} &	3.2	\std{4e-1} \\  \rowcolor{db}
    \multicolumn{2}{l}{\emph{\textbf{AE average}}} & \emph{18.3	\std{3e+0}} &	\emph{8.8	\std{7e-1}} &	\emph{3.4	\std{6e-1}}\\
    \midrule \rowcolor{lg}
    & SimCLR      &26.3	\std{1e+0} &   10.6	\std{8e-1} &	7.2	\std{1e+0}\\ \rowcolor{lg}
    & SimSiam     &27.1	\std{5e-1} &	6.3	\std{7e-1} &	7.8	\std{3e-1}\\  \rowcolor{lg}
    \multirow{-3}{*}{SSL} 
    & BYOL        &27.4	\std{2e+0} &   11.1	\std{1e+0} &	6.5	\std{2e+0}\\  \rowcolor{dg}
    \multicolumn{2}{l}{\emph{\textbf{SSL average}}} &\emph{26.9	\std{6e-1}} &	\emph{9.3	\std{1e+0}} &	\emph{7.2	\std{1e+0}}\\
    \midrule \rowcolor{g}
    SL
    & Supervised  &25.7	\std{5e-1} &   12.3	\std{2e+0} &	5.3	\std{1e+0}\\ 
    \bottomrule
\end{tabular}
\label{tab:fmowcorr100_results}}
}
\end{table}

\begin{figure*}[t]
  \centering
  \vspace{-3em}
  \captionsetup[subfigure]{belowskip=1ex}
  \begin{subfigure}{0.9\linewidth}
    \centering
    \includegraphics[width=0.32\linewidth]{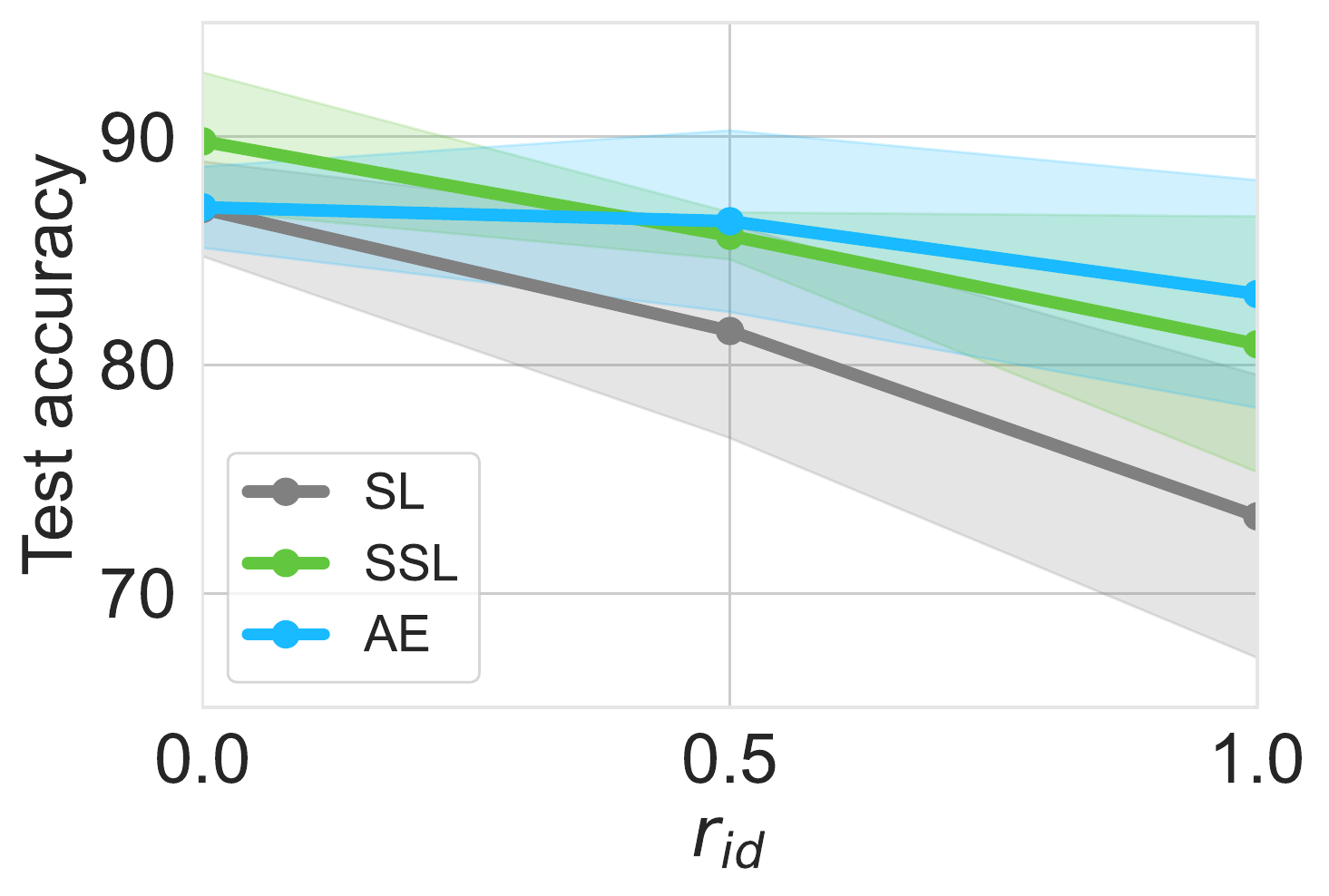}
    \includegraphics[width=0.32\linewidth]{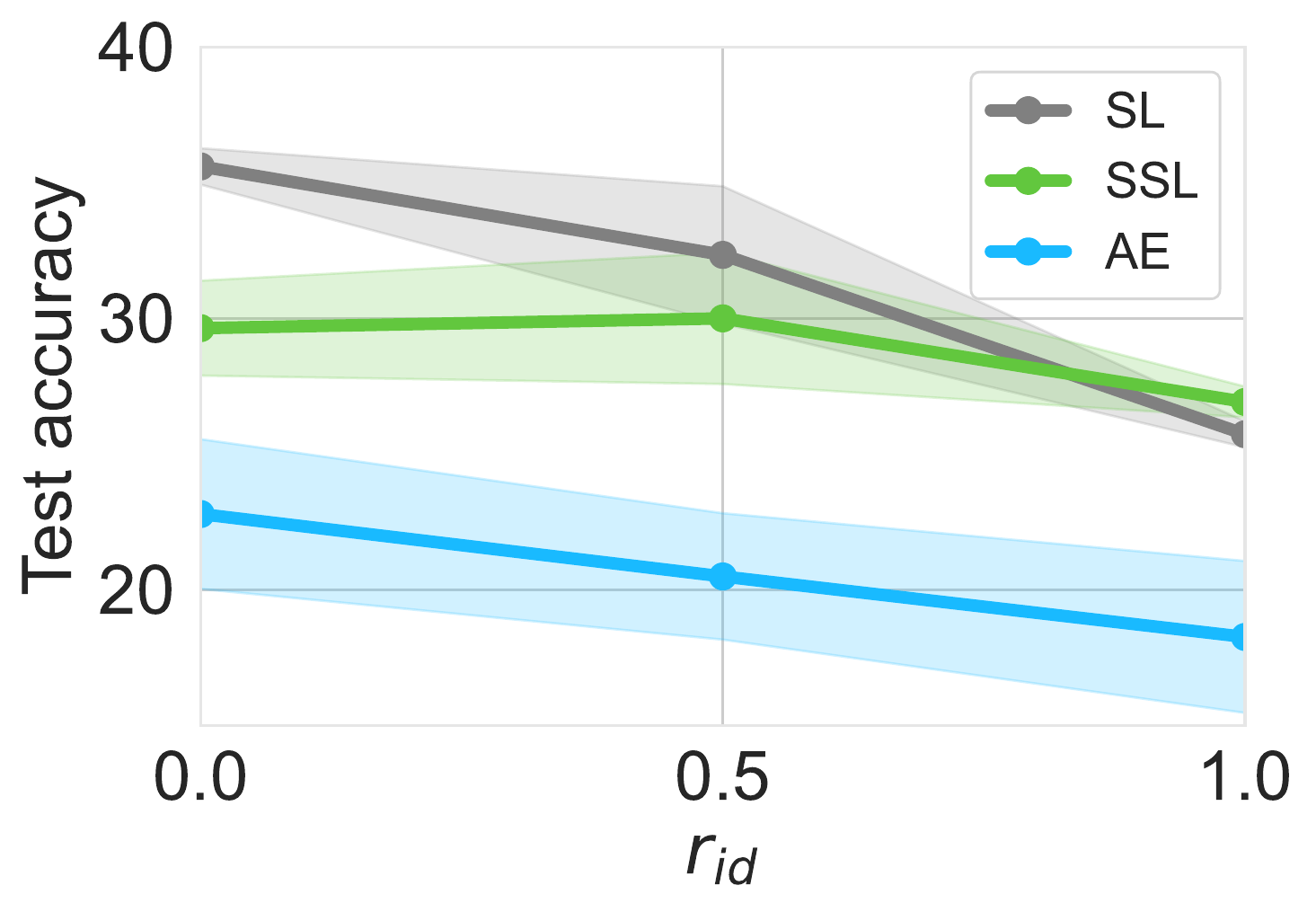}
    \includegraphics[width=0.32\linewidth]{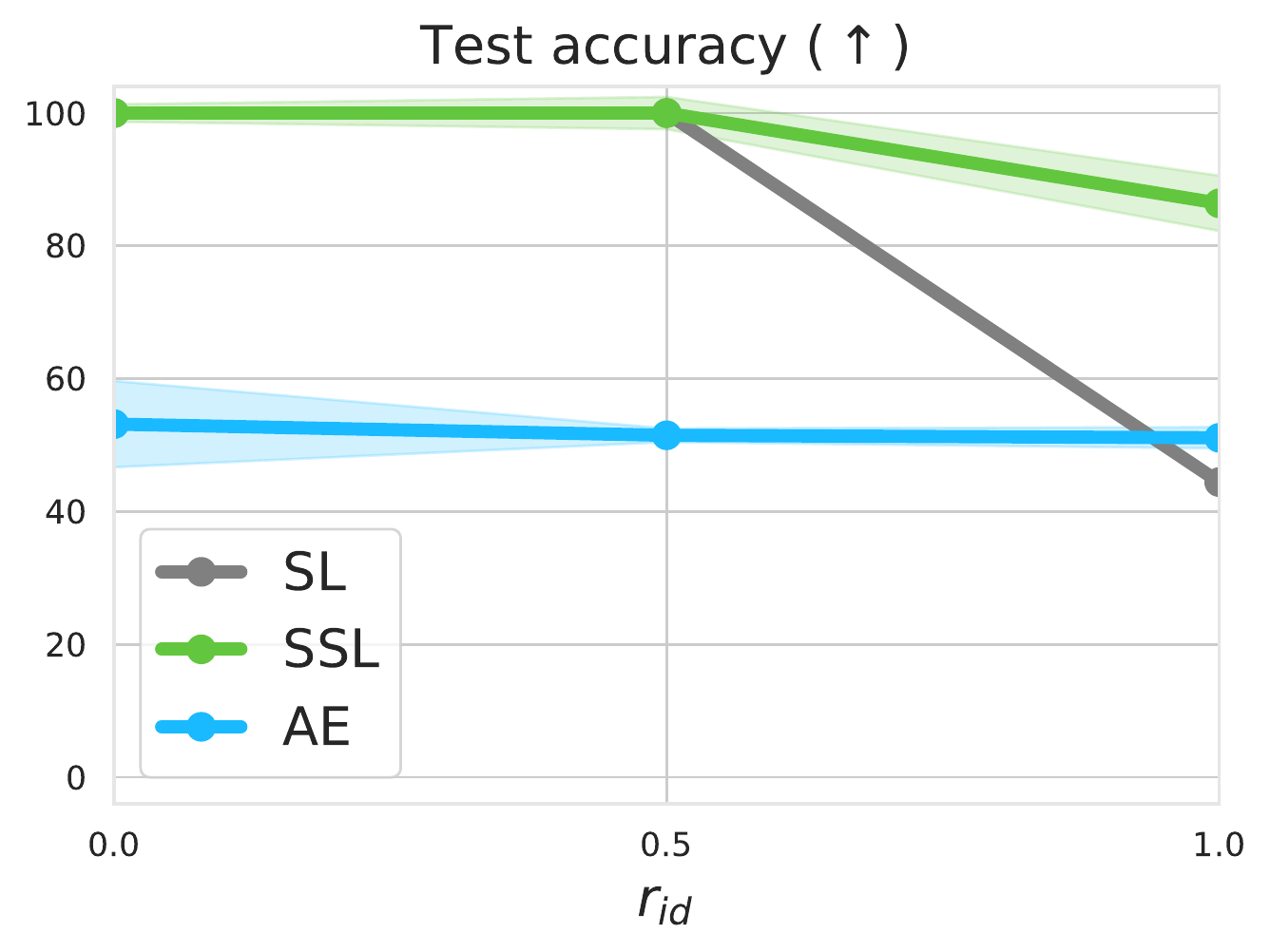}
  % \end{subfigure}
  % \begin{subfigure}
    % 
    % \includegraphics[width=0.32\linewidth]{images/camelyon_fmow_c/b_camelyon17.pdf}\vspace{-8pt}
    % \caption{Camelyon17-CS  (\%).}\label{fig:camelyon17_results}
  \end{subfigure} 
  \begin{subfigure}{0.9\linewidth}
    \centering
    \begin{subfigure}{0.32\linewidth}
    \includegraphics[width=0.99\linewidth]{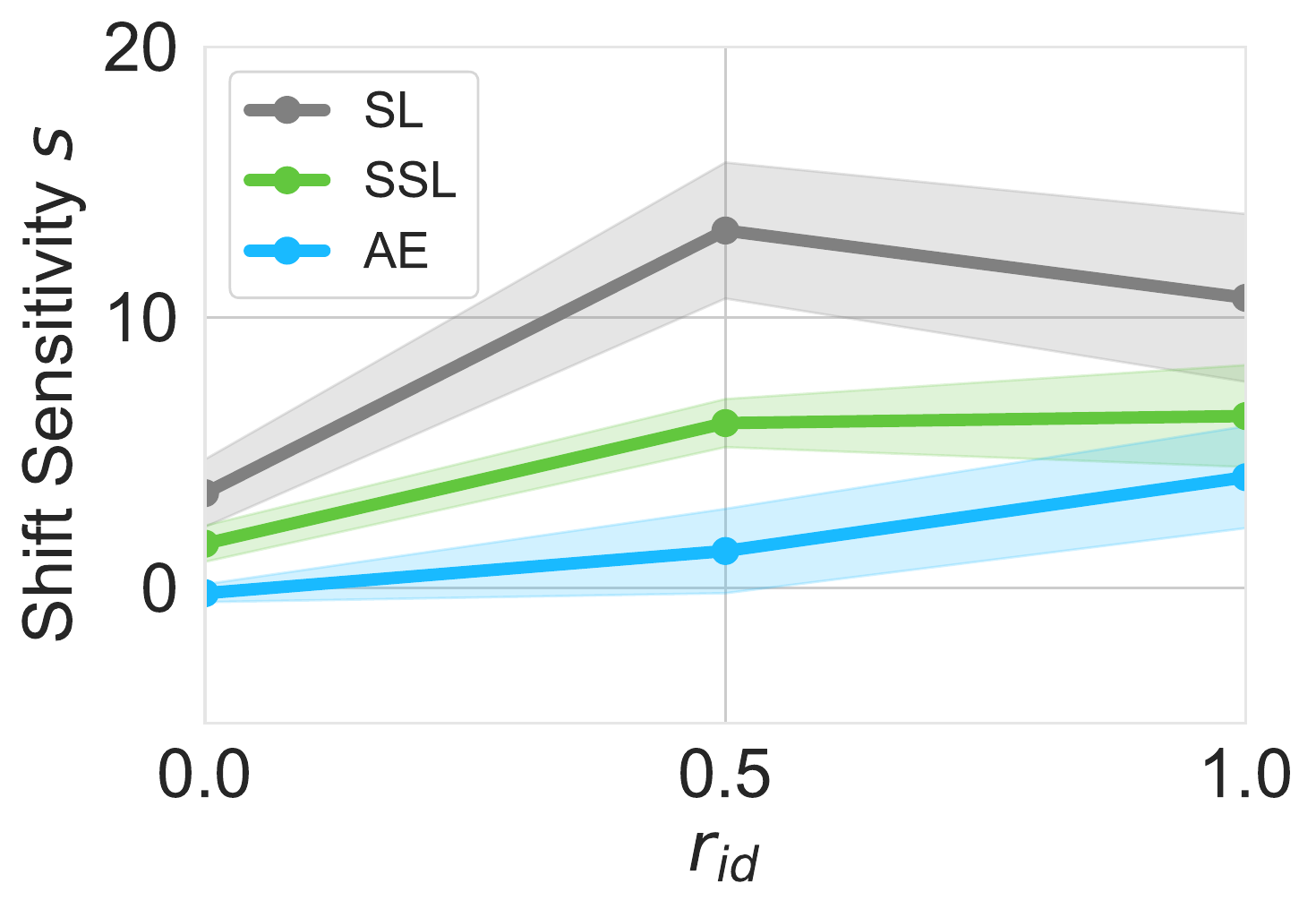}
    \subcaption{Camelyon17-CS}
    \end{subfigure}
    \begin{subfigure}{0.32\linewidth}
    \includegraphics[width=0.99\linewidth]{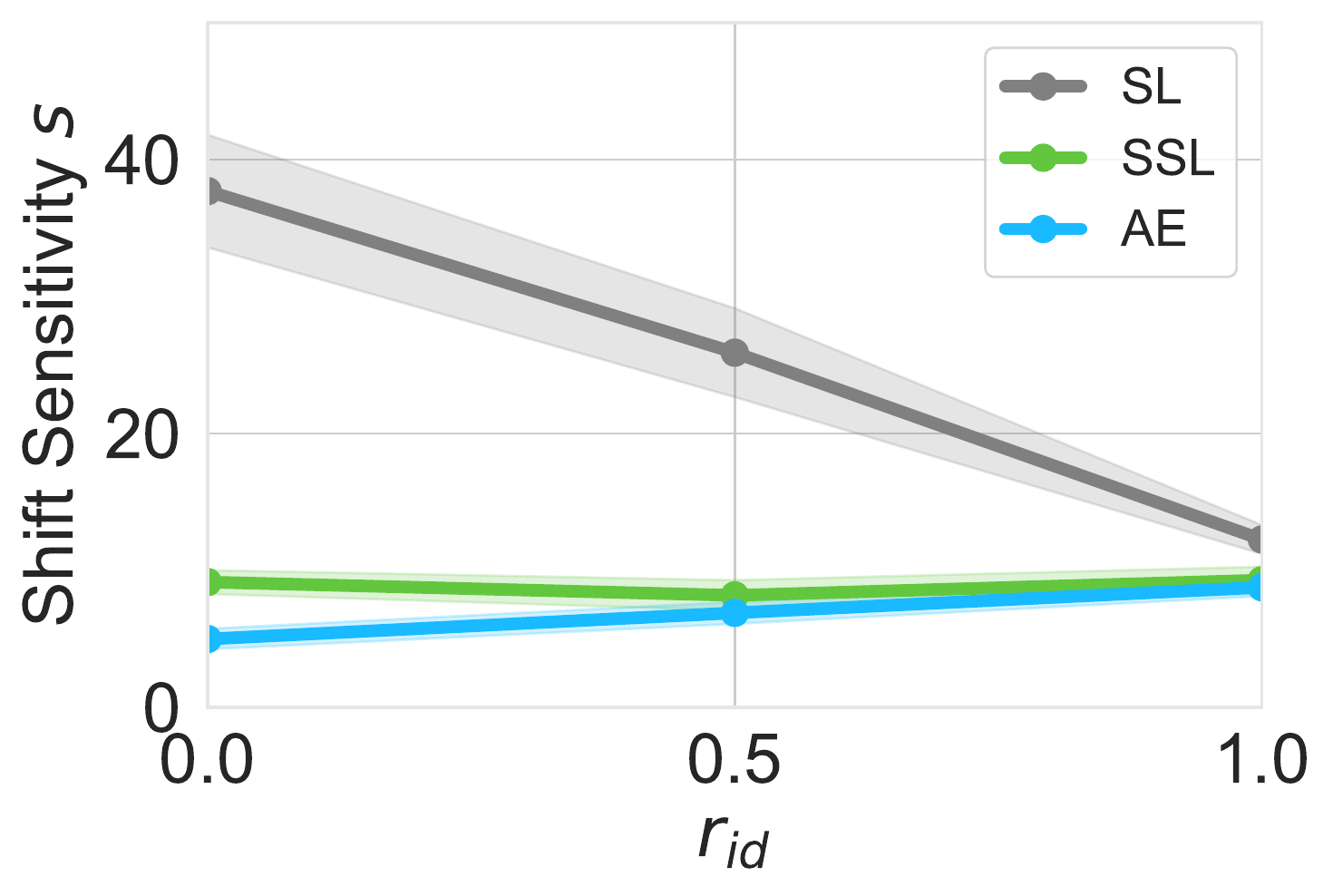}
    \subcaption{FMoW-CS}
  \end{subfigure}
    \begin{subfigure}{0.32\linewidth}
    \includegraphics[width=0.99\linewidth]{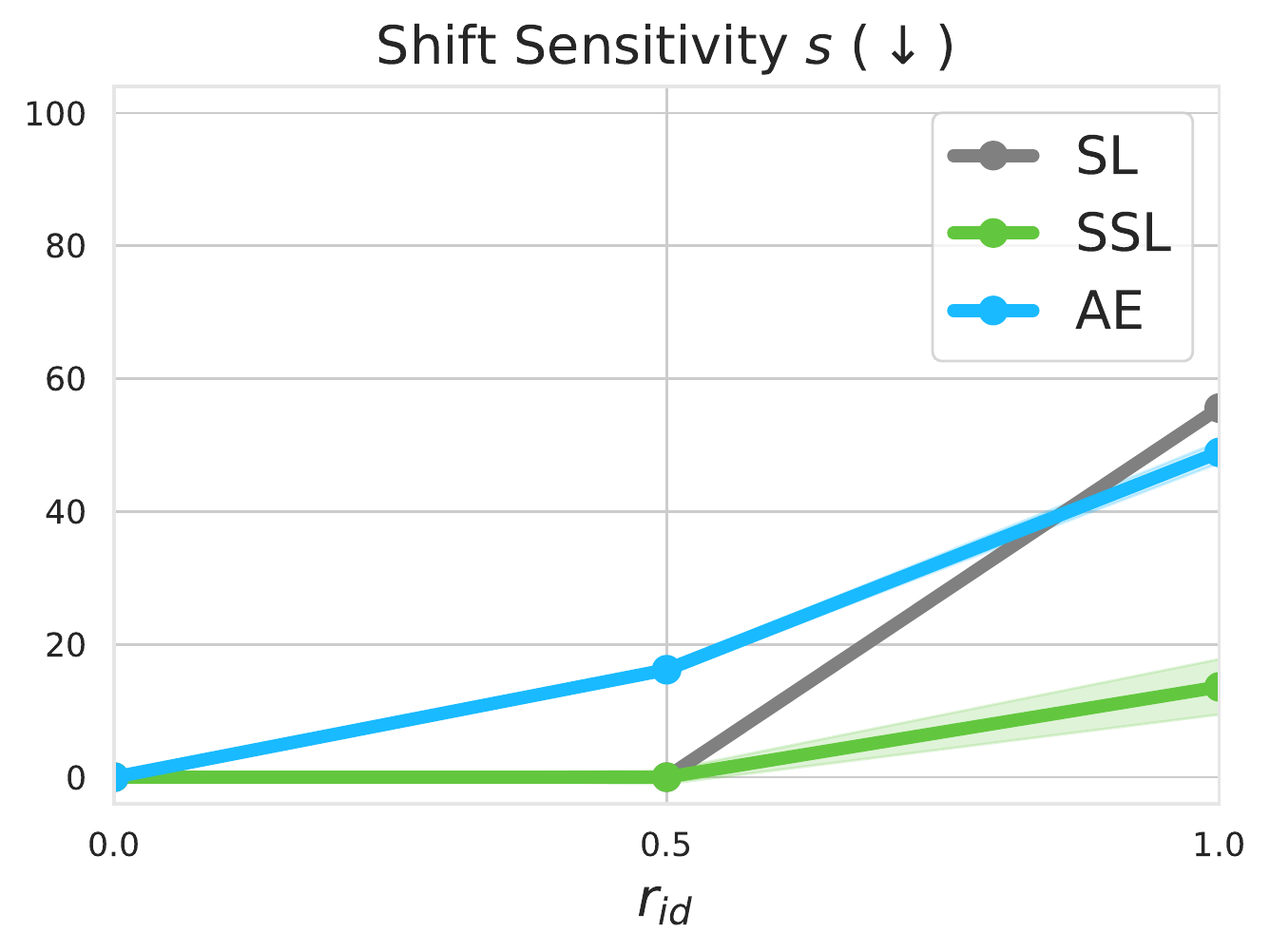}
    \subcaption{CdSprites}
  \end{subfigure}
    % \includegraphics[width=0.32\linewidth]{images/camelyon_fmow_c/b_fmow.pdf}
    % \caption{FMoW-CS  (\%).}\label{fig:fmow_results}
  \end{subfigure} 
  \vspace{-1em}
  \caption{Evaluations on Camelyon17-CS, FMoW-CS, and CdSprites with
  $r_{\text{id}}\in \{0, 0.5, 1.0\}$. We report OOD test accuracy
  using OOD-trained linear head ($\text{acc}_o(f,c_o)$) and shift
  sensitivity ($s$). Blue lines are results averaged over \gls{AE}
  models, green lines are \gls{SSL} models and grey is \gls{SL}.}
  \vspace{-1.5em}
  \label{fig:controllable_results}
\end{figure*}

For both datasets, the OOD test accuracy of all models drop as the
spurious correlation $r_{\text{id}}$ increases (\textit{top row}
in~\Cref{fig:controllable_results}).
However, this drop is the far more obvious in SL than in SSL and AE:
when $r_{\text{id}}=1$, SL's accuracy is $10\%$ lower than SSL on Camelyon17 and $2\%$ lower on FMoW --- a significant drop from its original $3\%$ lag on Camelyon17 and $5\%$ lead on FMoW.
This demonstrates that SL is less capable of dealing with more challenging distribution shift settings compared to SSL and AE.
In terms of shift sensitivity (\textit{bottom row,} \cref{fig:controllable_results}), SL's remains the highest regardless of $r_{\text{id}}$;
curiously, we see a decrease in SL's shift sensitivity as $r_{\text{id}}$ increases in FMoW-CS, however this has more to do with the ID test set accuracy decreasing due to the subsampling of the dataset. 
%
% The linear head bias (\textit{rightmost column,} \cref{fig:controllable_results}) does not seem to be affected in any consistent manner by the varying of $r_{\text{id}}$, again demonstrating the benefit of re-training the linear classifier on small amount of OOD data even for realistic distribution shift problems.
}
{
\section{Experimental results} \label{sec:experiments}
We perform a hyperparameter search on learning rate, scheduler,
optimiser, representation size, etc. for each model. We use the standard SSL augmentations proposed in \citet{he2020moco,chen2020improved} for all models to ensure a fair comparison.
See \cref{sec:app_hp} for details.\looseness=-1 
\subsection{Synthetic distribution shift}  \label{sec:simplicity}
In this section, we evaluate the performance of \gls{SL}, \gls{SSL} and \gls{AE} algorithms on synthetic distribution shift tasks, utilising the MNIST-CIFAR dataset \citep{shah2020pitfalls} and the CdSprites dataset \citep{fish}.
All results are averaged over 5 random seeds.

\subsubsection{MNIST-CIFAR} \label{subsec:mnist_cifar}
\textbf{\emph{Finding:}} \emph{Under this extreme distribution shift setting, \gls{SSL} and \gls{AE} significantly outperform \gls{SL}. The OOD accuracy of SSL and AE can be notably improved by retraining the linear head on OOD data, however the OOD accuracy of SL remains low even with the OOD-trained linear head.}

% \begin{wrapfigure}[10]{r}{0.2\linewidth}
% \vspace{-1.8em}
% \begin{subfigure}{0.99\linewidth}
%     \includegraphics[width=0.95\linewidth]{images/mnist_cifar/ms_id.png}
%     \vspace{-4pt}
%     \subcaption{ID}\label{fig:ms_id_vis}
% \end{subfigure}\\
% \begin{subfigure}{0.99\linewidth}
%     \includegraphics[width=0.95\linewidth]{images/mnist_cifar/ms_ood.png}
%     \vspace{-4pt}
%     \subcaption{OOD}\label{fig:ms_ood_vis}
%     % \vspace{-10pt}
% \end{subfigure}
% \vspace{-0.8em}
% \caption{MNIST-CIFAR dataset.}\label{fig:ms}
% \end{wrapfigure}
%
The MNIST-CIFAR dataset consists of concatenations of images from two
classes of MNIST and CIFAR-10. 
%
% It contains the following splits with varying correlation between the
% labels of the MNIST and CIFAR-10 images 
In each concatenated image, the classes of the two datasets are either
correlated or uncorrelated depending on the split as discussed
below~(See \cref{fig:synthetic_v_real}, \textit{MNIST-CIFAR} for an
example):

\begin{itemize}[nosep,leftmargin=1em,labelwidth=*,align=left]
    \item \textbf{ID train, test}: Correlation between MNIST and
    CIFAR-10 labels is one. Each image belongs to one of the two
    classes: 1) MNIST ``\texttt{0}'' and CIFAR-10
    ``\texttt{automobile}'', and 2) MNIST ``\texttt{1}'' and CIFAR-10
    ``\texttt{plane}'' (\Cref{fig:synthetic_v_real}, \textit{top
    row});
    \item \textbf{OOD train, test}: Zero correlation between MNIST and CIFAR-10 labels, images from the two classes are randomly paired (\Cref{fig:synthetic_v_real}, \textit{bottom row}).
\end{itemize}

Since the MNIST features are much simpler than the CIFAR features, a model trained on the ID train set can use MNIST only to predict the label, even though the CIFAR images are just as predictive \citep{shah2020pitfalls}.
This results in poor performance when predicting the CIFAR label on the OOD test set, where there is no correlation between the MNIST and CIFAR labels.

We train a CNN backbone on the ID train set using the eight \gls{SL}, \gls{SSL} and \gls{AE} algorithms listed in \cref{sec:setup}. At test time, we freeze the backbone and train two linear heads on ID train and OOD train set respectively, and evaluate their performance on the ID and OOD test set to compute 1) OOD linear head accuracy \aoco, 2) shift sensitivity $s$ and, 3) linear head bias $b$.
See results in \cref{tab:mnist_cifar_results}.

We observe that all models achieve near perfect performance when predicting the MNIST label on OOD test set, all with low shift sensitivity and small linear head bias.
However, when predicting the labels of the more complex CIFAR images, unsupervised algorithms have a clear advantage over the supervised one:  SSL achieves the highest OOD accuracy at $86.1\%$, followed by AE at $79.9\%$ and SL at $51.5\%$ (near random).
The shift sensitivity $s$ of the three objectives follow a similar
trend, with SSL and AE scoring significantly lower than SL. This
indicates that unsupervised representations are significantly
less sensitive to distribution shift compared to those from SL, with the latter suffering a drop as large as $47.6\%$.
Interestingly, the classifier head bias $b$ for SSL and AE are
relatively high (around $30\%$), and is very low for SL ($0.8\%$),
indicating that the representations learned from SL is intrinsically un-robust to distribution shift.
That is, while there exist (linearly separable) CIFAR features in
the representations of SSL and AE that can be extracted using a linear
head trained on un-biased (OOD) data, these features are absent from
the representations of SL.

\begin{table}[t]
\centering
\caption{Evaluations on the MNIST-CIFAR dataset. We report accuracy on MNIST and CIFAR trained using OOD linear head ($\text{acc}_o(f,c_o)$), linear head bias ($b$) and shift sensitivity ($s$).}
\vspace*{-0.8\baselineskip}
\scalebox{0.78}{
\begin{tabular}{llccccccc}    \toprule
    \multirow{2}{*}{Regime} &\multirow{2}{*}{Method} & \multicolumn{3}{c}{MNIST (\%)} & & \multicolumn{3}{c}{CIFAR (\%)}  \\
    \cmidrule{3-5} \cmidrule{7-9}
    & &  $\text{acc}_o(f,c_o)\uparrow$ & $s \downarrow$ & $b$ && $\text{acc}_o(f,c_o)\uparrow$ & $s\downarrow$  & $b$ \\
    \midrule \rowcolor{lb}
    & AE  & 99.9 \std{1e-2} & 0.0 \std{1e-2} & 0.0 \std{2e-3} && 81.1 \std{1e+0} & 18.8 \std{1e+0} & 30.2 \std{1e+0}\\  \rowcolor{lb}
    & VAE  & 99.8 \std{8e-3} & -0.1 \std{9e-3} & 0.5 \std{1e-4} && 79.7 \std{4e+0} & 20.2  \std{3e+0} & 29.2  \std{6e+0}\\  \rowcolor{lb}
    & IWAE  & 99.8 \std{9e-3} & 0.0 \std{4e-3} & 0.1 \std{5e-3} && 80.8 \std{2e+0} & 19.0 \std{3e+0} & 30.0  \std{4e+0}\\  \rowcolor{lb}
    \multirow{-4}{*}{AE}
    & $\beta$-VAE  & 99.8 \std{2e-2} & 0.0 \std{4e-2} & -0.1 \std{3e-2} && 78.0 \std{3e+0} & 21.8  \std{4e+0} & 28.0  \std{4e+0}\\  \rowcolor{db}
    \multicolumn{2}{l}{\emph{\textbf{AE average}}} & \emph{\textbf{99.8 \std{1e-2}}}  & \emph{\textbf{0.0 \std{1e-2}}}  & \emph{0.1 \std{9e-3}} && \emph{79.9 \std{3e+0}}  & \emph{20.0  \std{4e+0}} & \emph{29.3  \std{4e+0}} \\
    \midrule \rowcolor{lg}
    & SimCLR  & 99.7 \std{1e-2} & 0.2 \std{1e-3} & -0.2 \std{3e-3} && 85.8 \std{1e+0} & 14.1 \std{2e+0} & 35.5  \std{1e+0}\\ \rowcolor{lg}
    & SimSiam  & 99.8 \std{2e-1} & 0.1 \std{2e-1} & 0.0 \std{9e-2} && 87.8 \std{2e+0} & 12.1  \std{2e+0} & 35.6  \std{4e+0}\\  \rowcolor{lg}
    \multirow{-3}{*}{SSL} 
    & BYOL  & 99.8 \std{4e-2} & 0.0 \std{1e-2} & 0.9 \std{8e-3} && 84.8 \std{9e-1} & 15.0  \std{1e+0} & 33.2  \std{1e+0}\\  \rowcolor{dg}
    \multicolumn{2}{l}{\emph{\textbf{SSL average}}} & \emph{\textbf{99.8 \std{8e-2}}} & \emph{0.1 \std{5e-2}} & \emph{0.2 \std{3e-2}} && \emph{\textbf{86.1 \std{2e+0}}} & \emph{\textbf{13.7 \std{2e+0}}} & \emph{34.8 \std{4e+0}} \\
    \midrule \rowcolor{g}
    SL
    & Supervised  & 97.7 \std{9e-1} & 1.4 \std{1e+0} & -0.3 \std{1e+0} && 51.5 \std{1e+0} & 47.6  \std{1e+0} & 0.8  \std{9e-1}\\ 
    \bottomrule
\end{tabular}
\label{tab:mnist_cifar_results}}
\vspace*{-\baselineskip}
\end{table}

\subsubsection{CdSprites}  \label{subsec:cdsprites_experiment}

% \begin{wrapfigure}[10]{r}{0.3\linewidth}
%      \centering
%      \vspace{-1.5em}
%      \begin{subfigure}[b]{0.45\linewidth}
%          \centering
%          \includegraphics[width=\linewidth]{images/cdsprites/cdsprites_example_corr100.png}
%          \caption{$r = 1$}
%          \label{subfig:cdsprites_example_corr100}
%      \end{subfigure}
%      \begin{subfigure}[b]{0.45\linewidth}
%          \centering
%          \includegraphics[width=\linewidth]{images/cdsprites/cdsprites_example_corr0.png}
%          \caption{$r = 0$}
%          \label{subfig:cdsprites_example_corr0}
%      \end{subfigure}
%      \caption{CdSprites dataset. Each subplot shows 16 samples from the dataset.}
%      \label{fig:cdsprites_example}
%      \vspace{1.0em}
% \end{wrapfigure}

\textbf{\emph{Finding:}} \emph{Similar to MNIST-CIFAR, under extreme
distribution shift,~\gls{SSL} and \gls{AE} are better than
\gls{SL}; when the shift is less extreme, \gls{SSL} and \gls{SL}
achieve comparably strong OOD generalisation performance while
\gls{AE}'s performance is much weaker.}

%  When the shift is less extreme, \gls{SSL} and \gls{SL}
% achieve comparable strong OOD generalisation performance while
% \gls{AE}'s performance is much weaker.}

CdSprites is a colored variant of the popular dSprites dataset \citep{dsprites}, which consists of images of 2D sprites that are procedurally generated from multiple latent factors. 
The CdSprites dataset induces a spurious correlation between the color
and shape of the sprites, by coloring the sprites conditioned on the
shape following a controllable correlation coefficient
$r_{\text{id}}$.
See \cref{fig:synthetic_v_real} for an example: when $r_{\text{id}}=1$ color is completely dependent on shape (\textit{top row}, oval-purple, heart-cyan, square-white), and when $r_{\text{id}}=0$, color and shape are randomly matched (\textit{bottom row}).

\citet{fish} observes that when $r_{\text{id}}$ is high, SL model tend to use color only to predict the label while ignoring shape features due to the texture bias of CNN \citep{texture1, texture2}.
First, we consider the setting of extreme distribution shift similar
to MNIST-CIFAR by setting \(r_{\text{id}}=1\) in the ID train and test
splits. In the OOD train and test splits, the correlation coefficient
is set to zero to investigate how well the model learns both the shape
and the color features.~\Cref{tab:cdsprites_results} reports the three
metrics of interest using the same evaluation protocol as before.

Similar to MNIST-CIFAR, we observe that all models achieve near
perfect performance when predicting the simpler feature, i.e. color on
the OOD test set. However, when predicting shape, the more complex
feature on the OOD test set, SSL~(and also AEs to a lesser extent) is
far superior to SL. Additionally, the shift sensitivity of SSL~(and AE
to a lesser extent) are much smaller than SL, indicating that SSL/AE
models are more robust to extreme distribution shift. The linear head
bias also follows a similar trend as for MNIST-CIFAR, showing that
representations learned using SL methods are inherently not robust to
spurious correlations. This is not the case for SSL and AE algorithms
where a large linear head bias shows that is the ID linear heads and
not the representations that injects the bias.\looseness=-1
\paragraph{Controllable distribution shift}

We extend this experiment to probe the performance of these algorithms
under \emph{varying} degrees of distribution shifts. We generate three
versions of the CdSprites dataset with three different correlation
coefficients $r_{\text{id}} \in \{0, 0.5, 1\}$ of the ID train set. 
As before, the correlation coefficient of the OOD split is set to
zero%
%\footnote{Note that when $r_{\text{id}} = 0$, there is no
% distribution shift between the ID and OOD splits.}. 
The rest of the experimental protocol stays the same.
%
% Similar to MNIST-CIFAR, each version of CdSprites contain 4 splits, including 1) ID train, 2) ID test with $r_{\text{id}}$, and 3) OOD train 4) OOD test with $r_{\text{ood}}$.
% %
% We use the same evaluation protocol as for MNIST-CIFAR to acquire the OOD linear head accuracy \aoco, shift sensitivity $s$ and linear head bias $b$ on the three $r_{\text{id}}$ versions of the dataset.
%
The OOD test accuracy and the shift sensitivity for varying
\(r_{\text{id}}\) is plotted in~\cref{fig:controllable_results} and a
detailed breakdown of results is available in
\cref{app:additional_results}.\looseness=-1

\Cref{fig:controllable_results} shows that despite increasing
distribution shift between the ID and OOD splits (with increasing
$r_{\text{id}}$) the OOD performance of SSL and AE does not suffer.
However, the OOD accuracy of SL plummets and its shift sensitivity
explodes at \(r_{\text{id}}=1\).
%
% This highlights SL's weakness in handling larger distribution shift
% scenarios.
%
Interestingly, SSL maintains a high OOD test accuracy regardless of
the level of distribution shift: when $r_{\text{id}}<1$ its
performance is on par with SL, and when the distribution shift becomes
extreme with $r_{\text{id}}=1$ it significantly outperforms SL both in
terms of accuracy and shift sensitivity.
In comparison, AE models' accuracy lingers around $50\%$, with increasingly higher shift sensitivity as $r_{\text{id}}$ increases.
However, under extreme distribution shift with $r_{\text{id}}=1$ it
still performs better than SL, with slightly higher OOD accuracy and
lower shift sensitivity.\looseness=-1
%
% In terms of the linear head bias, we see a continuous increase for the AE as the distribution shift increases, whereas for SSL and SL the linear head bias is zero for $r_{\text{id}}<1$ and it peaks for all methods at $r_{\text{id}}=1$.

\begin{table}[t]
\vspace{-30pt}
  \centering
  \caption{Evaluations on the CdSprites dataset with $r_{\text{id}}=1.0$. We report accuracy for color and shape classifiers trained using OOD linear head ($\text{acc}_o(f,c_o)$), linear head bias ($b$) and shift sensitivity ($s$).}
  % \vspace*{-0.8\baselineskip}
  \scalebox{0.78}{
  \begin{tabular}{llccccccc}    \toprule
      \multirow{2}{*}{Regime} &\multirow{2}{*}{Method} & \multicolumn{3}{c}{Color classification (\%)} & & \multicolumn{3}{c}{Shape classification (\%)}  \\
      \cmidrule{3-5} \cmidrule{7-9}
      & &  $\text{acc}_o(f,c_o)\uparrow$ & $s \downarrow$ & $b$ && $\text{acc}_o(f,c_o)\uparrow$ & $s\downarrow$  & $b$ \\
      \midrule \rowcolor{lb}
      & AE  & 100.0 \std{0e+0} & 0.0 \std{2e-3} & 0.3 \std{5e-1} && 46.1 \std{6e-1}  &  53.9 \std{6e-1}  & 12.7 \std{5e-1} \\  \rowcolor{lb}
      & VAE  & 99.7 \std{3e-1} & 0.3 \std{3e-1} & -0.3 \std{3e-1} && 52.4 \std{2e+0}  &  47.6 \std{2e+0}   & 18.9 \std{3e+0}  \\  \rowcolor{lb}
      \multirow{-3}{*}{AE}& IWAE  & 100.0 \std{0e+0} & 0.0 \std{2e-3} & 0.4 \std{5e-1} && 58.9 \std{2e+0}  &  41.1 \std{2e+0}  & 25.6 \std{2e+0}  \\
      % & $\beta$-VAE  &  &  &  &&  &  & \\ 
      \rowcolor{db}
      \multicolumn{2}{l}{\emph{\textbf{AE average}}} & \emph{99.9 \std{9e-1}}  & \emph{0.1 \std{9e-2}}  & \emph{0.1 \std{4e-1}} && \emph{52.5 \std{2e+0}}  & \emph{ 47.5 \std{2e+0} } & \emph{19.1 \std{2e+0} } \\
      \midrule \rowcolor{lg}
      & SimCLR  & 100.0 \std{0e+0} & 0.0 \std{0e+0} & 0.0 \std{1e-1} && 87.8 \std{5e-1} &  12.2 \std{5e-1}  & 54.5 \std{5e-1}  \\ \rowcolor{lg}
      & SimSiam  & 100.0 \std{0e+0} & 0.0 \std{0e+0} & 0.1 \std{1e-1}  && 69.2 \std{2e+0}  &  30.8 \std{2e+0}  & 35.6 \std{2e+0}  \\  \rowcolor{lg}
      \multirow{-3}{*}{SSL} 
      & BYOL  & 100.0 \std{0e+0} & 0.0 \std{0e+0} & 0.0 \std{0e+0} && 91.1 \std{4e+0}  &  8.9 \std{4e+0}   & 57.9 \std{4e+0}  \\  \rowcolor{dg}
      \multicolumn{2}{l}{\emph{\textbf{SSL average}}} & \emph{100.0 \std{0e+0}} & \emph{0.0 \std{0e+0}} & \emph{0.1 \std{3e-2}} && \emph{82.7 \std{2e+0} } & \emph{17.3 \std{2e+0} } & \emph{49.3 \std{4e+0} } \\
      \midrule \rowcolor{g}
      SL
      & Supervised  & 100.0 \std{0e+0} & 0.0 \std{0e+0} & 0.0 \std{3e-2} && 44.0 \std{7e-1} & 56.0 \std{7e-1}   & 10.7 \std{7e-1}  \\ 
      \bottomrule
  \end{tabular}
  \label{tab:cdsprites_results}}
   \vspace*{-10pt}
  \end{table}

\subsection{Real-world distribution shift}  \label{sec:dg} 
In this section we investigate the performance of different objectives on real-world distribution shift tasks. 
We use two datasets from WILDS~\citep{koh2021wilds}: 1) Camelyon17, which contains tissue scans acquired from different hospitals, and the task is to determine if a given patch contains breast cancer tissue; and 2) FMoW, which features satellite images of landscapes on five different continents, with the classification target as the type of infrastructure. See examples in \Cref{fig:synthetic_v_real}.
% %
Following the guidelines from WILDS benchmark, we perform 10 random seed runs for all Camelyon17 experiment and 3 random seed runs for FMoW. The error margin in \Cref{fig:controllable_results} represent standard deviation.

\subsubsection{Original WILDS Datasets} \label{sec:wilds_original}
\textbf{\emph{Findings:}} \emph{SL is significantly more sensitive to
distribution shift than SSL and AE; representations from SSL obtain
higher OOD accuracy than SL on Camelyon17 but lower on FMoW. AE is
consistently the least sensitive to distribution shift though it has
the lowest accuracy. The performance of all models significantly
improves by retraining the linear head on a small amount of OOD
data.\looseness=-1}

The original Camelyon17 and FMoW dataset from WILDS benchmark both contains the following three splits: ID train, OOD validation and OOD test.
We further create five splits specified as follows:
\begin{itemize}[nosep,leftmargin=1em,labelwidth=*,align=left]
    \item \textbf{ID train, test}: Contains 90\% and 10\% of the original ID train split, respectively;
    \item \textbf{OOD train, test}: Contains 10\% and 90\% of the original OOD test split, respectively;
    \item \textbf{OOD validation}: Same as the original OOD validation split.
\end{itemize}
Following WILDS, we use OOD validation set to perform early stopping and choose hyperparameters; we also use DenseNet-121 \citep{dense} as the backbone for all models.
We follow similar evaluation protocol as previous experiments, and in addition adopt 10-fold cross-validation for the OOD train and test set.
% Similar to our previous experiments, after training the backbone model on the ID train set, we freeze the backbone and train two linear heads on ID train and OOD train respectively, and evaluate all models on the ID and OOD test set to compute our results.
%
See results in \Cref{tab:camelyon17_results,tab:fmow_results}, where
following WILDS, we report performance on Camelyon17 using standard
average accuracy and on FMoW using worst-group accuracy.

One immediate observation is that in contrast to our previous experiments on synthetic datasets, SL's OOD accuracy is much higher in comparison on realistic distribution shift tasks: it is the best performing model on FMoW with $35.6\%$ worst-group accuracy on OOD test set;
its OOD accuracy is the lowest on Camelyon17, however it is only $3\%$
worse than the highest accuracy achieved by SSL ($89.8\%$). This
highlights the need to study realistic datasets along with synthetic ones.
Nonetheless, we find that SSL is still the best performing method on
Camelyon17 and achieves competitive performance on FMoW with accuracy
$29.6\%$ --- despite learning without labels!
AE has much lower OOD accuracy on FMoW compared to the other two methods: we believe this is due to its reconstruction-based objective wasting modelling capacity on high frequency details, a phenomenon frequently observed in prior work \citep{bao2021beit,ramesh2021zero}.
Note that the standard deviation for all three methods are quite high for Camelyon17: this is a known property of the dataset and similar pattern is observed across most methods on WILDS benchmark \citep{koh2021wilds}.

In terms of shift sensitivity, unsupervised objectives including SSL
and AE consistently outperforms SL --- this stands out the most on
FMoW, where the shift sensitivity of SSL and AE are $9.1\%$ and
$5.0\%$ respectively, while SL is as high as $37.7\%$. 
This observation further validates our previous finding on synthetic
datasets, that SSL and AE's ID accuracy is a relatively reliable
indication of their generalisation performance, while SL can undergo a
huge performance drop under distribution shift, which can be dangerous
for the deployment of such models. We highlight that, in sensitive
application domains, a low shift sensitivity is an important criterion
as it implies that the model's performance will remain consistent when
the distribution shifts.
Another interesting observation here is that for all objectives on both datasets, the classifier bias $b$ is consistently high.
This indicates the bias of the linear classification head plays a significant role even for real world distribution shifts, and that it is possible to mitigate this effect by training the linear head using a small amount of OOD data (in this case $10\%$ of the original OOD test set).

\begin{table}[t]
\parbox{.48\linewidth}{
\centering
\caption{Evaluations on test set of Camelyon17, all metrics computed using average accuracy.}
\vspace*{-0.8\baselineskip}
\scalebox{0.7}{
\begin{tabular}{llccc}    \toprule
    \multirow{2}{*}{Regime} &\multirow{2}{*}{Method} & \multicolumn{3}{c}{Metrics (\%)} \\
    \cmidrule{3-5} 
    & &  $\text{acc}_o(f,c_o)\uparrow$ & $s \downarrow$ & $b$ \\
    \midrule \rowcolor{lb}
    & AE          & 84.4 \std{2e+0} & -0.6 \std{1e+0}  &12.7 \std{2e+0} \\  \rowcolor{lb}
    & VAE         & 88.1 \std{2e+0} & 0.5 \std{2e+0} &39.0 \std{2e+0} \\  \rowcolor{lb}
    & IWAE        & 88.1 \std{1e+0} & -0.9 \std{3e+0}  &39.1 \std{4e+0} \\  \rowcolor{lb}
    \multirow{-4}{*}{AE}
    & $\beta$-VAE & 87.1 \std{4e+0} &0.2 \std{4e+0} & 36.0 \std{5e+0} \\  \rowcolor{db}
    \multicolumn{2}{l}{\emph{\textbf{AE average}}} & \emph{86.9 \std{2e+0}}  & \emph{\textbf{-0.2 \std{3e+0}}}  & \emph{31.7 \std{3e+0}} \\
    \midrule \rowcolor{lg}
    & SimCLR      & 92.7 \std{2e+0} & 0.4 \std{1e+0} & 8.3 \std{1e+0} \\ \rowcolor{lg}
    & SimSiam     & 86.7 \std{1e+0} &3.1 \std{1e+0} & 7.9 \std{3e+0}\\  \rowcolor{lg}
    \multirow{-3}{*}{SSL} 
    & BYOL        & 89.9 \std{1e+0} &1.4 \std{1e+0} & 10.3 \std{2e+0}\\  \rowcolor{dg}
    \multicolumn{2}{l}{\emph{\textbf{SSL average}}} & \emph{\textbf{89.8 \std{1e+0}}} & \emph{1.6 \std{1e+0}} & \emph{8.8 \std{2e+0}}\\
    \midrule \rowcolor{g}
    SL
    & Supervised  & 86.8 \std{2e+0} & 3.5 \std{1e+0} &  7.4 \std{3e+0} \\ 
    \bottomrule
\end{tabular}
\label{tab:camelyon17_results}}
}
\hspace{10pt}
\parbox{.48\linewidth}{
\centering
\caption{Evaluations on test set of FMoW, all metrics computed using worst-group accuracy.}
\vspace*{-0.8\baselineskip}
\scalebox{0.7}{
\begin{tabular}{llccc}    \toprule
    \multirow{2}{*}{Regime} &\multirow{2}{*}{Method} & \multicolumn{3}{c}{Metrics (\%)} \\
    \cmidrule{3-5} 
    & &  $\text{acc}_o(f,c_o)\uparrow$ & $s \downarrow$ & $b$ \\
    \midrule \rowcolor{lb}
    & AE          & 26.9 \std{9e-3} &6.4 \std{6e-3}  & 5.8 \std{1e-2} \\  \rowcolor{lb}
    & VAE         & 21.7 \std{6e-3} & 4.7 \std{4e-3} & 8.0 \std{2e-2} \\  \rowcolor{lb}
    & IWAE        & 20.9 \std{2e-2} &5.5 \std{1e-2}  & 7.8 \std{1e-2} \\  \rowcolor{lb}
    \multirow{-4}{*}{AE}
    & $\beta$-VAE & 21.7 \std{5e-3} &3.4 \std{6e-3} & 7.6  \std{8e-3} \\  \rowcolor{db}
    \multicolumn{2}{l}{\emph{\textbf{AE average}}} & \emph{22.8 \std{3e-2}}  & \emph{\textbf{5.0  \std{7e-3}}}  & \emph{7.3  \std{1e-2}} \\
    \midrule \rowcolor{lg}
    & SimCLR      & 29.9 \std{6e-3} & 10.7 \std{6e-3} & 7.6 \std{7e-3} \\ \rowcolor{lg}
    & SimSiam     & 27.8 \std{2e-2} &4.6 \std{1e-2} & 6.3 \std{2e-2}\\  \rowcolor{lg}
    \multirow{-3}{*}{SSL} 
    & BYOL        & 31.3 \std{1e-2} &12.1 \std{7e-3} & 7.9  \std{1e-2}\\  \rowcolor{dg}
    \multicolumn{2}{l}{\emph{\textbf{SSL average}}} & \emph{29.6 \std{2e-2}} & \emph{9.1 \std{9e-3}} & \emph{7.3 \std{1e-2}}\\
    \midrule \rowcolor{g}
    SL
    & Supervised  & \textbf{35.6 \std{7e-3}} &37.7 \std{4e-2} &  6.9 \std{9e-3} \\ 
    \bottomrule
\end{tabular}
\label{tab:fmow_results}}
}
\vspace{-10pt}
\end{table}

\subsubsection{WILDS Datasets with Controllable Shift} \label{sec:wilds_controllable}

\textbf{\emph{Findings:}} \emph{SL's OOD accuracy drops as more the distribution shift becomes more challenging, with SSL being the best performing model when the distribution shift is the most extreme. The shift sensitivity of SSL and AE are consistently lower than SL regardless of the level of shift.}

To examine models' generalisation performance under different levels of distribution shift, we create versions of these realistic datasets with {\em controllable shifts}, which we name Camelyon17-CS and FMoW-CS.
Specifically, we subsample the ID train set of these datasets to artificially create spurious correlation between the domain and label.
For instance, given dataset with domain \texttt{A}, \texttt{B} and label \texttt{0}, \texttt{1}, to create a version of the dataset where the spurious correlation is 1 we would sample only examples with label \texttt{0} from domain \texttt{A} and label \texttt{1} from domain \texttt{B}.
See \Cref{app:controllable_datasets} for further details.

Similar to CdSprites, we create three versions of both of these
datasets with the spurious correlation coefficient $r_{\text{id}} \in
\{0, 0.5,1\}$ in ID~(train and test) sets.
The OOD train, test and validation set remains unchanged\footnote{Note
that even when $r_{\text{id}}=0$, distribution shift between the ID
and OOD splits exists, as the spurious correlation is not the only
source of the distribution shift.}.
Using identical experimental setup as in \cref{sec:wilds_original}, we
report the results for Camelyon17-CS and FMoW-CS in
\Cref{fig:controllable_results} with detailed numerical results in \Cref{app:wilds_additional}.

% \begin{figure*}[t]
%   \centering
%   \captionsetup[subfigure]{belowskip=1ex}
%   \begin{subfigure}{\linewidth}
%     \centering
%     \includegraphics[width=0.32\linewidth]{images/camelyon_fmow_c/accood_clsood_camelyon17.pdf}
%     \includegraphics[width=0.32\linewidth]{images/camelyon_fmow_c/rho_camelyon17.pdf}
%     \includegraphics[width=0.32\linewidth]{images/camelyon_fmow_c/b_camelyon17.pdf}\vspace{-8pt}
%     \caption{Camelyon17-CS  (\%).}\label{fig:camelyon17_results}
%   \end{subfigure} 
%   \begin{subfigure}{\linewidth}
%     \centering
%     \includegraphics[width=0.32\linewidth]{images/camelyon_fmow_c/accood_clsood_fmow.pdf}
%     \includegraphics[width=0.32\linewidth]{images/camelyon_fmow_c/rho_fmow.pdf}
%     \includegraphics[width=0.32\linewidth]{images/camelyon_fmow_c/b_fmow.pdf}
%     \vspace{-8pt}
%     \caption{FMoW-CS  (\%).}\label{fig:fmow_results}
%   \end{subfigure} 
%   \vspace{-1em}
%   \caption{Evaluations on Camelyon17-CS and FMoW-CS, with $r_{\text{id}}\in \{0, 0.5, 1.0\}$. We report OOD test accuracy using OOD-trained linear head ($\text{acc}_o(f,c_o)$), shift sensitivity ($s$) and linear head bias ($b$). Blue lines are results averaged over \gls{AE} models, green lines are \gls{SSL} models and grey is \gls{SL}.}
%   \vspace{-1.5em}
%   \label{fig:controllable_results}
% \end{figure*}

\begin{figure*}[t]
  \centering
  \vspace{-3em}
  \captionsetup[subfigure]{belowskip=1ex}
  \begin{subfigure}{0.9\linewidth}
    \centering
    \includegraphics[width=0.32\linewidth]{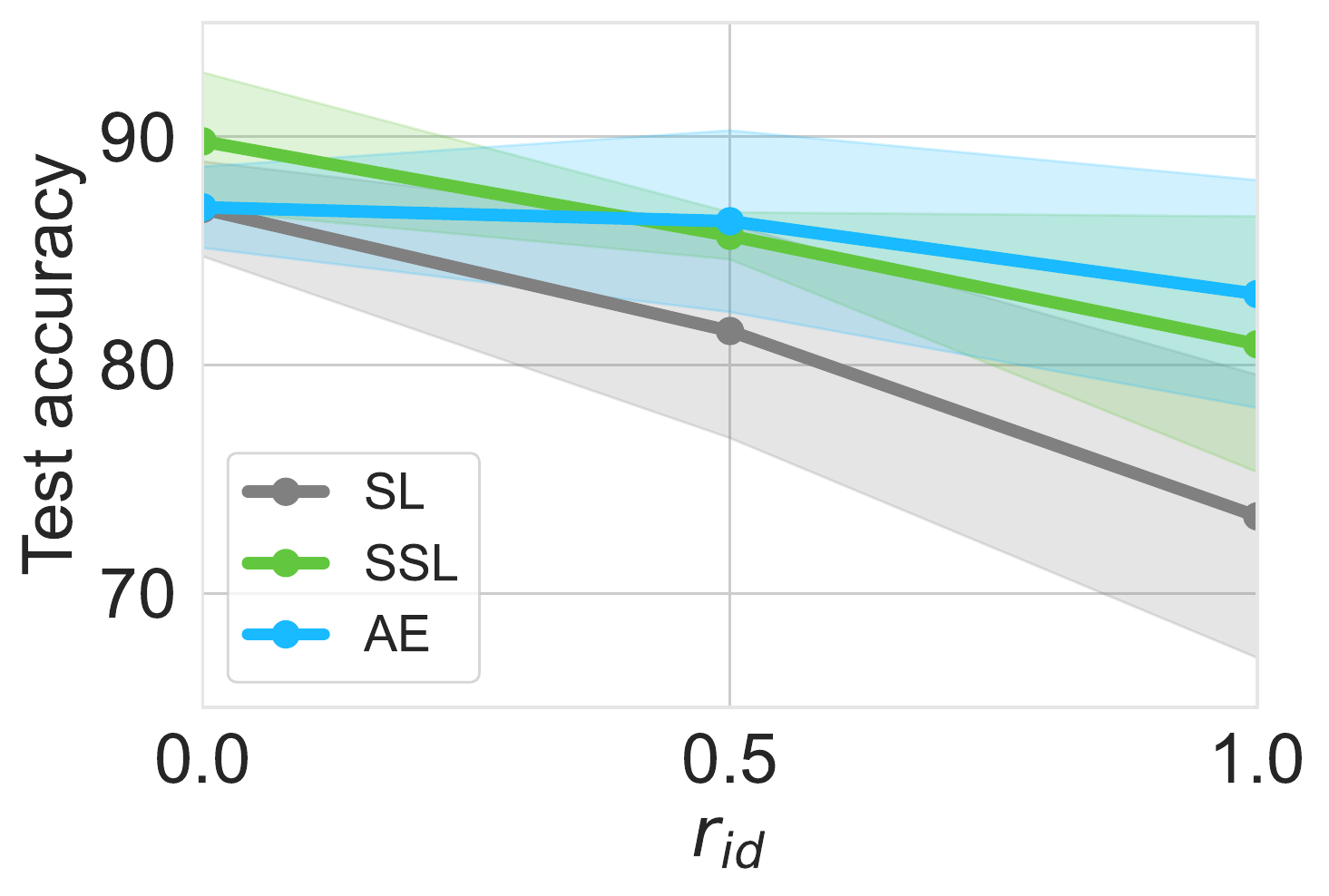}
    \includegraphics[width=0.32\linewidth]{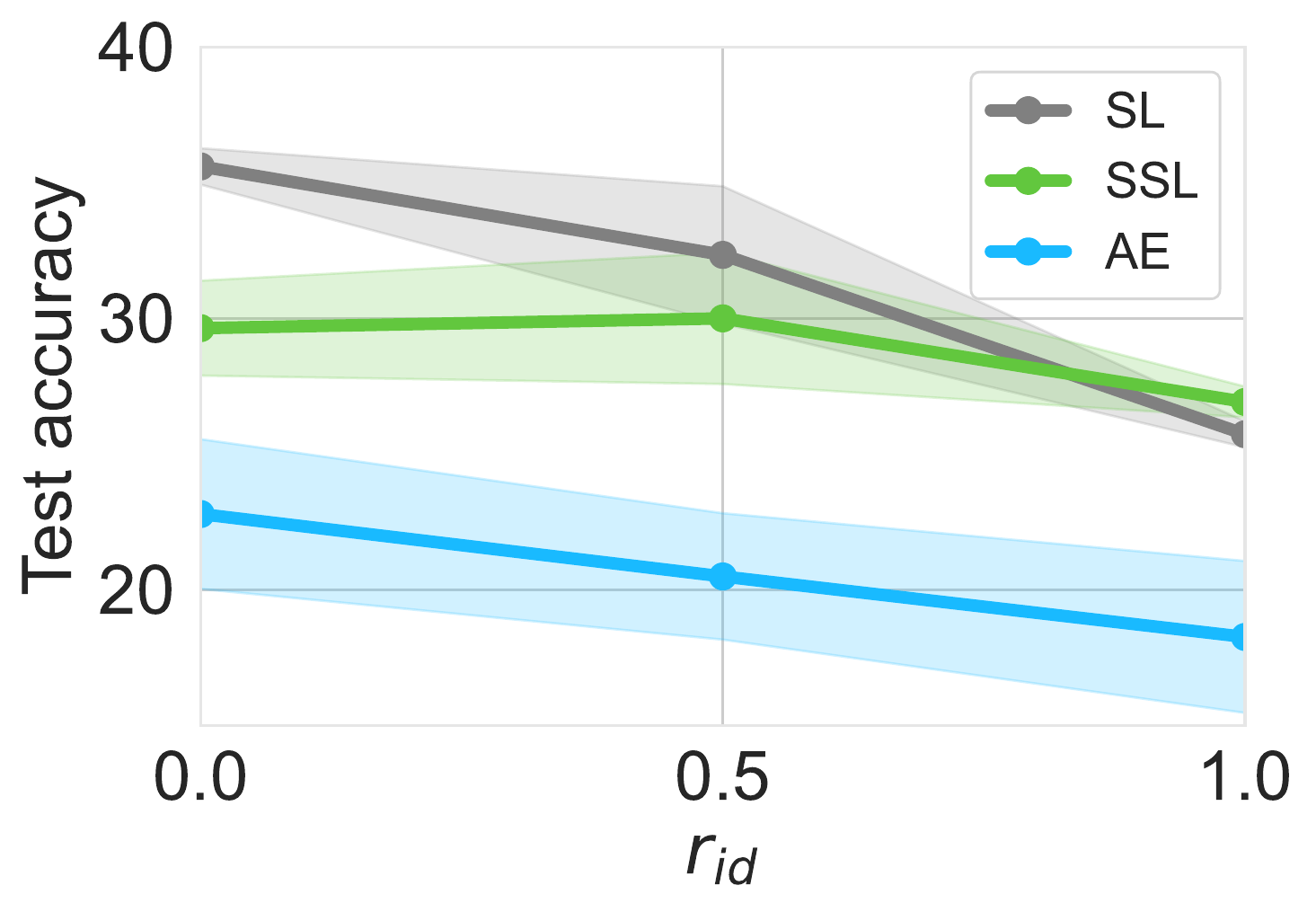}
    \includegraphics[width=0.32\linewidth]{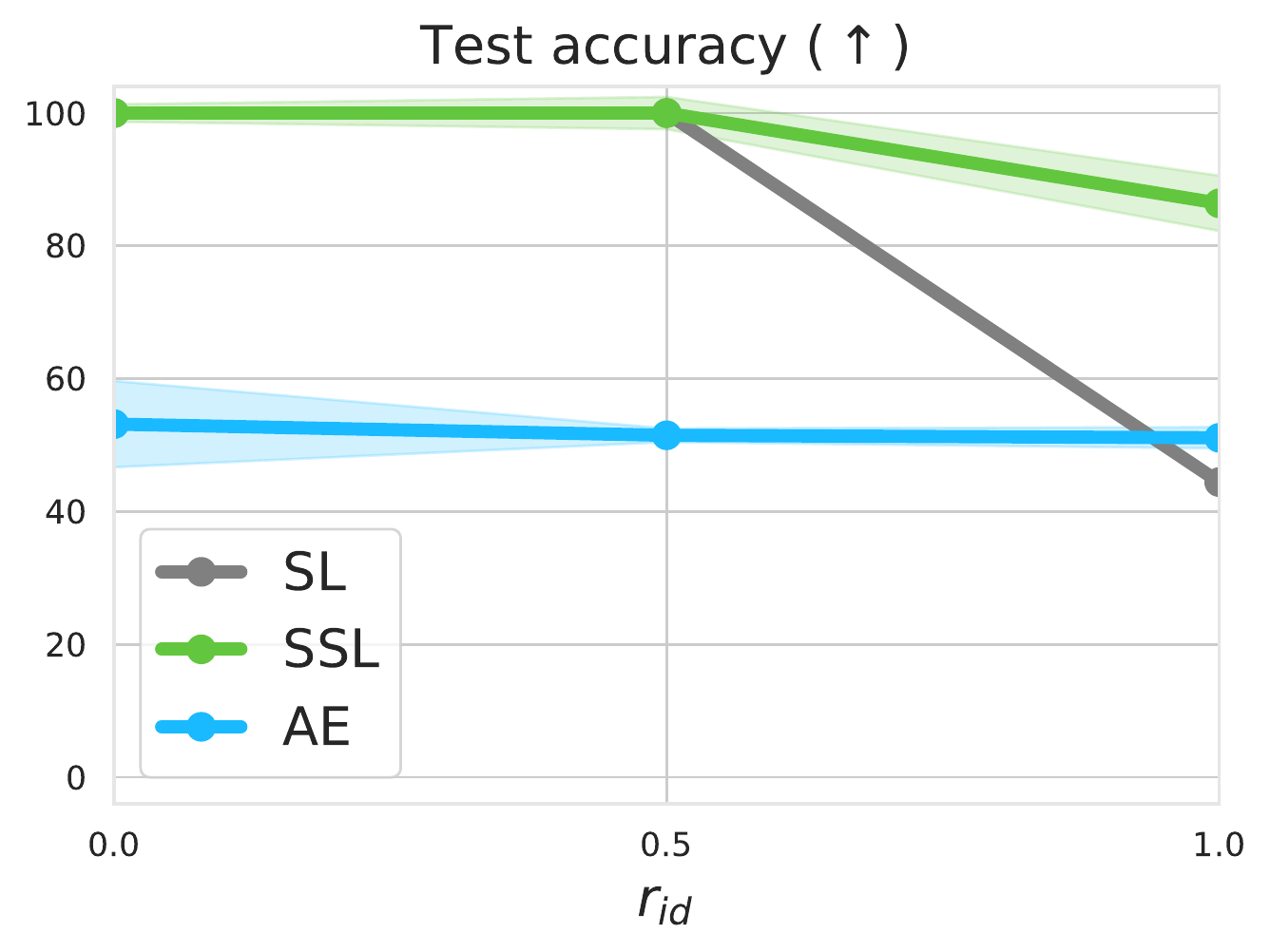}
  % \end{subfigure}
  % \begin{subfigure}
    % 
    % \includegraphics[width=0.32\linewidth]{images/camelyon_fmow_c/b_camelyon17.pdf}\vspace{-8pt}
    % \caption{Camelyon17-CS  (\%).}\label{fig:camelyon17_results}
  \end{subfigure} 
  \begin{subfigure}{0.9\linewidth}
    \centering
    \begin{subfigure}{0.32\linewidth}
    \includegraphics[width=0.99\linewidth]{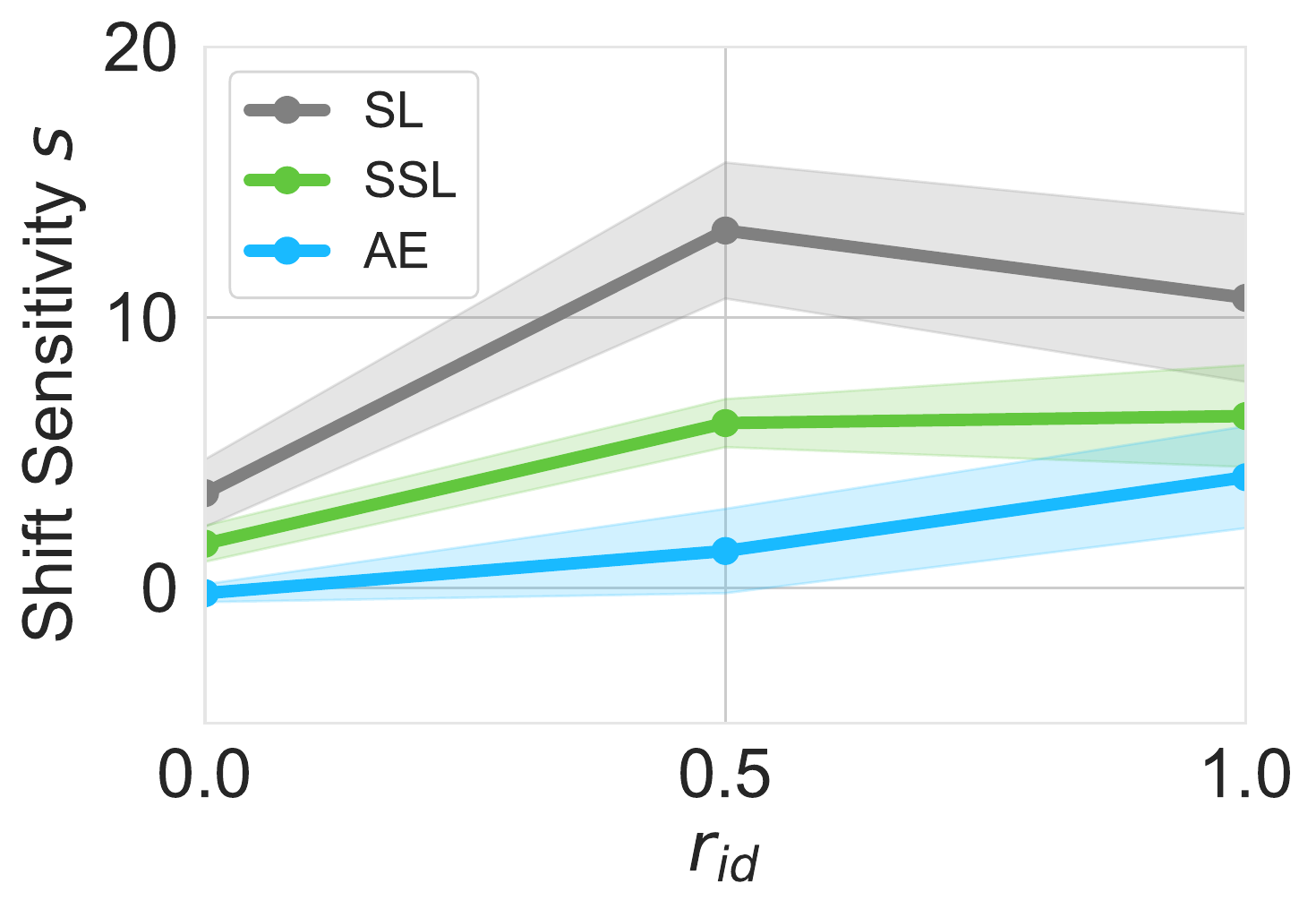}
    \subcaption{Camelyon17-CS}
    \end{subfigure}
    \begin{subfigure}{0.32\linewidth}
    \includegraphics[width=0.99\linewidth]{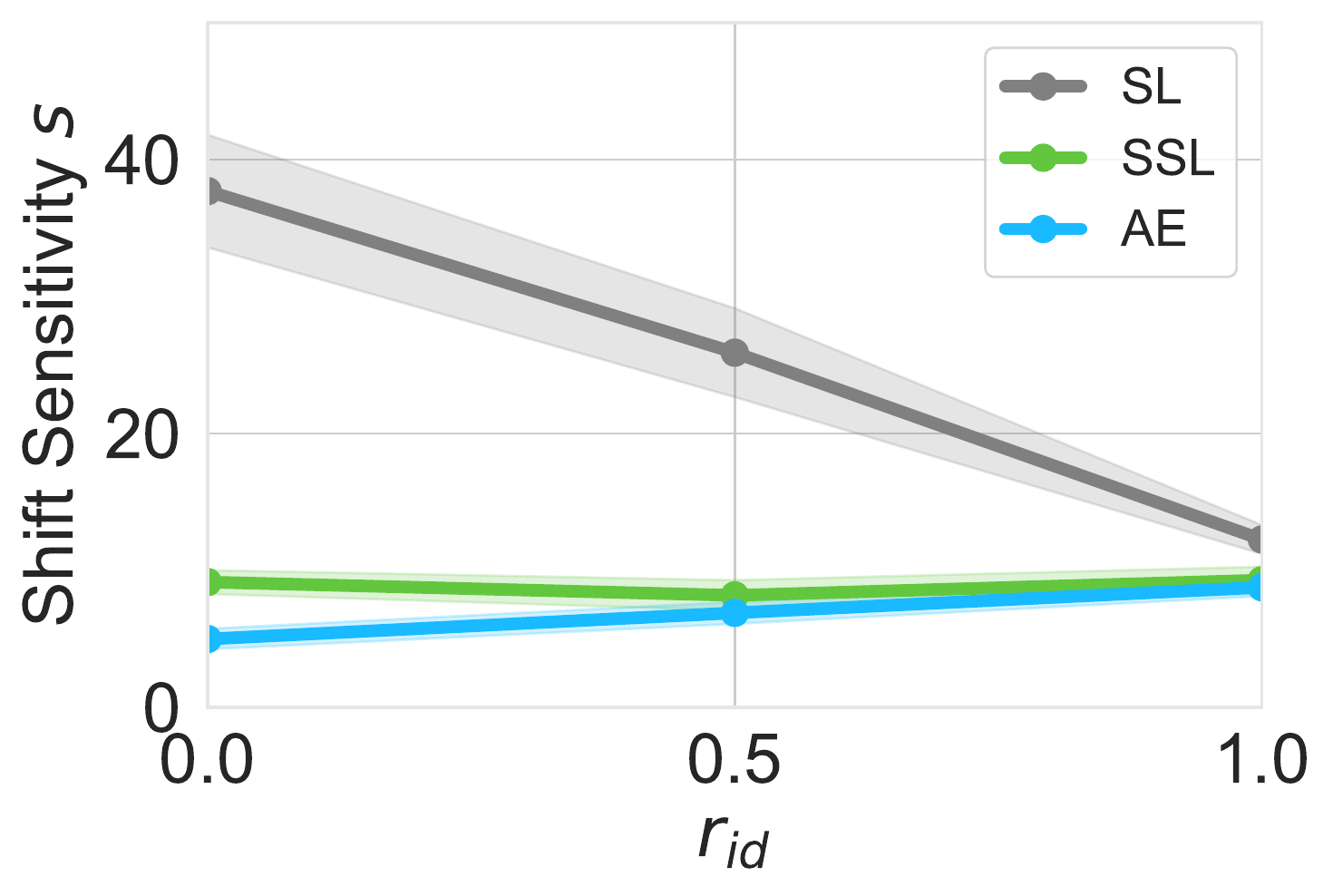}
    \subcaption{FMoW-CS}
  \end{subfigure}
    \begin{subfigure}{0.32\linewidth}
    \includegraphics[width=0.99\linewidth]{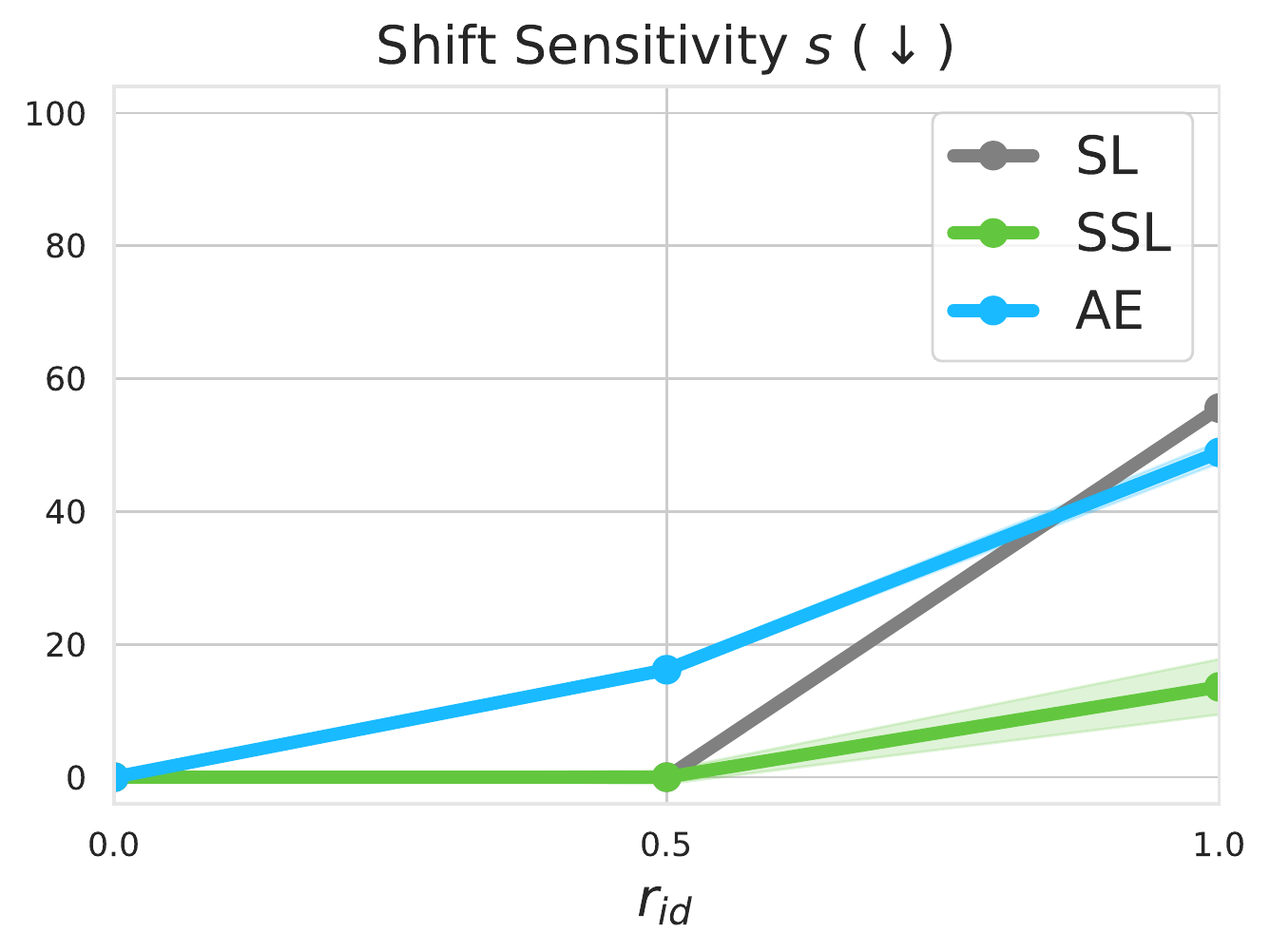}
    \subcaption{CdSprites}
  \end{subfigure}
    % \includegraphics[width=0.32\linewidth]{images/camelyon_fmow_c/b_fmow.pdf}
    % \caption{FMoW-CS  (\%).}\label{fig:fmow_results}
  \end{subfigure} 
  \vspace{-1em}
  \caption{Evaluations on Camelyon17-CS, FMoW-CS, and CdSprites with
  $r_{\text{id}}\in \{0, 0.5, 1.0\}$. We report OOD test accuracy
  using OOD-trained linear head ($\text{acc}_o(f,c_o)$) and shift
  sensitivity ($s$). Blue lines are results averaged over \gls{AE}
  models, green lines are \gls{SSL} models and grey is \gls{SL}.}
  \vspace{-1.5em}
  \label{fig:controllable_results}
\end{figure*}

For both datasets, the OOD test accuracy of all models drop as the
spurious correlation $r_{\text{id}}$ increases (\textit{top row}
in~\Cref{fig:controllable_results}).
However, this drop is the far more obvious in SL than in SSL and AE:
when $r_{\text{id}}=1$, SL's accuracy is $10\%$ lower than SSL on Camelyon17 and $2\%$ lower on FMoW --- a significant drop from its original $3\%$ lag on Camelyon17 and $5\%$ lead on FMoW.
This demonstrates that SL is less capable of dealing with more challenging distribution shift settings compared to SSL and AE.
In terms of shift sensitivity (\textit{bottom row,} \cref{fig:controllable_results}), SL's remains the highest regardless of $r_{\text{id}}$;
curiously, we see a decrease in SL's shift sensitivity as $r_{\text{id}}$ increases in FMoW-CS, however this has more to do with the ID test set accuracy decreasing due to the subsampling of the dataset. 
%
% The linear head bias (\textit{rightmost column,} \cref{fig:controllable_results}) does not seem to be affected in any consistent manner by the varying of $r_{\text{id}}$, again demonstrating the benefit of re-training the linear classifier on small amount of OOD data even for realistic distribution shift problems.
}

\section{Related Work}

While we are the first to systematically evaluate the OOD generalisation performance of unsupervised learning algorithms, there are other insightful work that considers the robustness to distribution shift of other existing, non-specialised methods/techniques.
%including important deep learning characteristics such as training procedure and architecture.
%
For instance, \citet{liu2022empirical} studies the impact of different pre-training set-ups to distribution shift robustness, including dataset, objective and data augmentation.
\citet{ghosal2022vision} focuses on architecture, and found that Vision Transformers are more robust to spurious correlations than ConvNets when using larger models and are given more training data;
further, \citet{liu2021self} found that SSL is more robust to data imbalance.
\citet{azizi2022robust} also performed extensive studies on the generalisation performance of SSL algorithms on medical data.
Interestingly, \citet{robinson2021simplicity} also investigates the robustness of contrastive-SSL methods against extreme spurious correlation (i.e.simplicity bias). 
However, their work did not consider the linear head bias found in \citep{kirichenko2022last,kang2019decoupling} and led to opposing conclusions.
%
% without training the linear head on OOD data, their finding is opposite to ours --- that SSL methods are not able to avoid shortcut solutions.
%
In contrast, our work investigates the distribution shift performance of unsupervised algorithms, with experiments on both synthetic and realistic settings that go beyond the data imbalance regime.
By isolating the linear head bias in our experiments, we find that
unsupervised, especially SSL-learned representations, achieves similar
if not better generalisation performance than SL under a wide range of
distribution shift settings. See~ \Cref{app:related_work} for a more
detailed discussion on distribution shift problems\looseness=-1.
%  such as simplicity
% bias, shortcut learning and domain generalisation as well as popular
% methods in these fields.
\section{Conclusion and future work}

In this paper, we investigate the robustness of both unsupervised (AE, SSL) and supervised (SL) objectives for distribution shift. 
Through extensive and principled experiments on both synthetic and
realistic distribution shift tasks, we find unsupervised
representation learning algorithms to consistently outperform SL when
the distribution shift is extreme. In addition, we see that SSL's OOD
accuracy is comparable, if not better to SL in all experiments.
This is particularly crucial, as most work studying distribution shift
for images are developed in the SL regime.
%However our studies suggest
% significant robustness gains in using unsupervised, especially SSL
% objectives.
%
We hope that these results inspire more future work on
unsupervised/semi-supervised representation learning methods for OOD
generalisation.
Another important finding is that unsupervised models' performance remains relatively stable under distribution shift.
This is especially crucial for the real-world application of these
machine learning systems, as this indicates that the ID performance of
SSL/AE algorithms are a more reliable indicator of how they would
perform in different environments at deployment, while that of SL is
not.
It is also worth noting that while models trained with AE objectives are consistently the least sensitive to distribution shift on realistic datasets, their OOD performance can be low especially when presented with complex data (such as FMoW).
This is consistent to the observation in prior work that these models can waste modelling capacity on high frequency details, and suggests that one should be careful about employing AE algorithms on large scale, complex tasks.
Finally, a key contribution of this work is establishing the existence of linear head bias even for realistic distribution shift problems.
We believe that using an OOD-trained linear head is necessary to be
able to make comparisons between various algorithms irrespective of
the final downstream task, and on the other hand, more efforts in the
field of distribution shift could be devoted into re-balancing the
linear layer.\looseness=-1

\subsection*{Acknowledgements}
YS and PHST were supported by the UKRI grant: Turing AI Fellowship EP/W002981/1 and EPSRC/MURI grant: EP/N019474/1. We would also like to thank the Royal Academy of Engineering and FiveAI.
YS was additionally supported by Remarkdip through their PhD Scholarship Programme. 
ID was supported by the SNSF grant \texttt{\#200021\_188466}.
AS was partially supported by the ETH AI Center postdoctoral fellowship.
Special thanks to Alain Ryser for suggesting the design of controllable versions of the WILDS dataset, and to Josh Dillon for helpful suggestions in the early stage of this project.

% %
\bibliographystyle{plainnat}
\bibliography{main}

\begin{thebibliography}{55}
\providecommand{\natexlab}[1]{#1}
\providecommand{\url}[1]{\texttt{#1}}
\expandafter\ifx\csname urlstyle\endcsname\relax
  \providecommand{\doi}[1]{doi: #1}\else
  \providecommand{\doi}{doi: \begingroup \urlstyle{rm}\Url}\fi

\bibitem[Alemi et~al.(2017)Alemi, Fischer, Dillon, and Murphy]{vib}
Alexander~A. Alemi, Ian Fischer, Joshua~V. Dillon, and Kevin Murphy.
\newblock Deep variational information bottleneck.
\newblock In \emph{{International Conference on Learning Representations}},
  2017.

\bibitem[Arjovsky et~al.(2019)Arjovsky, Bottou, Gulrajani, and
  Lopez-Paz]{arjovsky2019irm}
Martin Arjovsky, L{\'e}on Bottou, Ishaan Gulrajani, and David Lopez-Paz.
\newblock Invariant risk minimization.
\newblock \emph{arXiv:1907.02893}, 2019.

\bibitem[Azizi et~al.(2022)Azizi, Culp, Freyberg, Mustafa, Baur, Kornblith,
  Chen, MacWilliams, Mahdavi, Wulczyn, et~al.]{azizi2022robust}
Shekoofeh Azizi, Laura Culp, Jan Freyberg, Basil Mustafa, Sebastien Baur, Simon
  Kornblith, Ting Chen, Patricia MacWilliams, S~Sara Mahdavi, Ellery Wulczyn,
  et~al.
\newblock Robust and efficient medical imaging with self-supervision.
\newblock \emph{arXiv:2205.09723}, 2022.

\bibitem[Bao et~al.(2021)Bao, Dong, and Wei]{bao2021beit}
Hangbo Bao, Li~Dong, and Furu Wei.
\newblock Beit: Bert pre-training of image transformers.
\newblock \emph{ArXiv preprint}, abs/2106.08254, 2021.

\bibitem[Ben-David et~al.(2010)Ben-David, Blitzer, Crammer, Kulesza, Pereira,
  and Vaughan]{ben2010theory}
Shai Ben-David, John Blitzer, Koby Crammer, Alex Kulesza, Fernando Pereira, and
  Jennifer~Wortman Vaughan.
\newblock A theory of learning from different domains.
\newblock \emph{Machine learning}, 2010.

\bibitem[Brendel and Bethge(2019)]{texture2}
Wieland Brendel and Matthias Bethge.
\newblock Approximating cnns with bag-of-local-features models works
  surprisingly well on imagenet.
\newblock In \emph{{International Conference on Learning Representations}},
  2019.

\bibitem[Burda et~al.(2016)Burda, Grosse, and Salakhutdinov]{burda2015iwae}
Yuri Burda, Roger~B. Grosse, and Ruslan Salakhutdinov.
\newblock Importance weighted autoencoders.
\newblock In \emph{{International Conference on Learning Representations}},
  2016.

\bibitem[Chen et~al.(2020{\natexlab{a}})Chen, Kornblith, Norouzi, and
  Hinton]{chen2020simclr}
Ting Chen, Simon Kornblith, Mohammad Norouzi, and Geoffrey~E. Hinton.
\newblock A simple framework for contrastive learning of visual
  representations.
\newblock In \emph{{International Conference on Machine Learning}},
  2020{\natexlab{a}}.

\bibitem[{Chen} and {He}(2021)]{chen2021simsiam}
Xinlei {Chen} and Kaiming {He}.
\newblock Exploring simple siamese representation learning.
\newblock In \emph{{IEEE Conference on Computer Vision and Pattern
  Recognition}}, 2021.

\bibitem[Chen et~al.(2020{\natexlab{b}})Chen, Fan, Girshick, and
  He]{chen2020improved}
Xinlei Chen, Haoqi Fan, Ross Girshick, and Kaiming He.
\newblock Improved baselines with momentum contrastive learning.
\newblock \emph{arXiv preprint arXiv:2003.04297}, 2020{\natexlab{b}}.

\bibitem[da~Costa et~al.(2022)da~Costa, Fini, Nabi, Sebe, and
  Ricci]{solo-learn}
Victor Guilherme~Turrisi da~Costa, Enrico Fini, Moin Nabi, Nicu Sebe, and Elisa
  Ricci.
\newblock solo-learn: A library of self-supervised methods for visual
  representation learning.
\newblock \emph{Journal of Machine Learning Research}, 2022.

\bibitem[Dock{\`e}s et~al.(2021)Dock{\`e}s, Varoquaux, and
  Poline]{dockes2021preventing}
J{\'e}r{\^o}me Dock{\`e}s, Ga{\"e}l Varoquaux, and Jean-Baptiste Poline.
\newblock Preventing dataset shift from breaking machine-learning biomarkers.
\newblock \emph{GigaScience}, 2021.

\bibitem[Ganin et~al.(2016)Ganin, Ustinova, Ajakan, Germain, Larochelle,
  Laviolette, Marchand, and Lempitsky]{ganin2016dann}
Yaroslav Ganin, Evgeniya Ustinova, Hana Ajakan, Pascal Germain, Hugo
  Larochelle, Fran{\c{c}}ois Laviolette, Mario Marchand, and Victor Lempitsky.
\newblock Domain-adversarial training of neural networks.
\newblock \emph{The journal of machine learning research}, 2016.

\bibitem[Geirhos et~al.(2019)Geirhos, Rubisch, Michaelis, Bethge, Wichmann, and
  Brendel]{texture1}
Robert Geirhos, Patricia Rubisch, Claudio Michaelis, Matthias Bethge, Felix~A.
  Wichmann, and Wieland Brendel.
\newblock Imagenet-trained cnns are biased towards texture; increasing shape
  bias improves accuracy and robustness.
\newblock In \emph{{International Conference on Learning Representations}},
  2019.

\bibitem[Geirhos et~al.(2020)Geirhos, Jacobsen, Michaelis, Zemel, Brendel,
  Bethge, and Wichmann]{geirhos2020shortcut}
Robert Geirhos, J{\"o}rn-Henrik Jacobsen, Claudio Michaelis, Richard Zemel,
  Wieland Brendel, Matthias Bethge, and Felix~A Wichmann.
\newblock Shortcut learning in deep neural networks.
\newblock \emph{Nature Machine Intelligence}, 2020.

\bibitem[Ghosal et~al.(2022)Ghosal, Ming, and Li]{ghosal2022vision}
Soumya~Suvra Ghosal, Yifei Ming, and Yixuan Li.
\newblock Are vision transformers robust to spurious correlations?
\newblock \emph{arXiv:2203.09125}, 2022.

\bibitem[Glocker et~al.(2019)Glocker, Robinson, Castro, Dou, and
  Konukoglu]{glocker2019machine}
Ben Glocker, Robert Robinson, Daniel~C Castro, Qi~Dou, and Ender Konukoglu.
\newblock Machine learning with multi-site imaging data: an empirical study on
  the impact of scanner effects.
\newblock \emph{arXiv:1910.04597}, 2019.

\bibitem[Gretton et~al.(2009{\natexlab{a}})Gretton, Smola, Huang, Schmittfull,
  Borgwardt, and Sch{\"o}lkopf]{gretton2008covariate}
A.~Gretton, AJ. Smola, J.~Huang, M.~Schmittfull, KM. Borgwardt, and
  B.~Sch{\"o}lkopf.
\newblock \emph{Covariate shift and local learning by distribution matching}.
\newblock 2009{\natexlab{a}}.

\bibitem[Gretton et~al.(2009{\natexlab{b}})Gretton, Smola, Huang, Schmittfull,
  Borgwardt, and Sch{\"o}lkopf]{gretton2009covariate}
Arthur Gretton, Alex Smola, Jiayuan Huang, Marcel Schmittfull, Karsten
  Borgwardt, and Bernhard Sch{\"o}lkopf.
\newblock Covariate shift by kernel mean matching.
\newblock \emph{Dataset shift in machine learning}, 2009{\natexlab{b}}.

\bibitem[Grill et~al.(2020)Grill, Strub, Altch{\'{e}}, Tallec, Richemond,
  Buchatskaya, Doersch, Pires, Guo, Azar, Piot, Kavukcuoglu, Munos, and
  Valko]{grill2020byol}
Jean{-}Bastien Grill, Florian Strub, Florent Altch{\'{e}}, Corentin Tallec,
  Pierre~H. Richemond, Elena Buchatskaya, Carl Doersch, Bernardo~{\'{A}}vila
  Pires, Zhaohan Guo, Mohammad~Gheshlaghi Azar, Bilal Piot, Koray Kavukcuoglu,
  R{\'{e}}mi Munos, and Michal Valko.
\newblock Bootstrap your own latent - {A} new approach to self-supervised
  learning.
\newblock In \emph{{Conference on Neural Information Processing Systems}},
  2020.

\bibitem[Gulrajani and Lopez-Paz(2020)]{domainbed}
Ishaan Gulrajani and David Lopez-Paz.
\newblock In search of lost domain generalization.
\newblock \emph{arXiv:2007.01434}, 2020.

\bibitem[Harary et~al.(2022)Harary, Schwartz, Arbelle, Staar, Abu-Hussein,
  Amrani, Herzig, Alfassy, Giryes, Kuehne, et~al.]{harary2022unsupervised}
Sivan Harary, Eli Schwartz, Assaf Arbelle, Peter Staar, Shady Abu-Hussein, Elad
  Amrani, Roei Herzig, Amit Alfassy, Raja Giryes, Hilde Kuehne, et~al.
\newblock Unsupervised domain generalization by learning a bridge across
  domains.
\newblock In \emph{{IEEE Conference on Computer Vision and Pattern
  Recognition}}, 2022.

\bibitem[He et~al.(2020)He, Fan, Wu, Xie, and Girshick]{he2020moco}
Kaiming He, Haoqi Fan, Yuxin Wu, Saining Xie, and Ross~B. Girshick.
\newblock Momentum contrast for unsupervised visual representation learning.
\newblock In \emph{{IEEE Conference on Computer Vision and Pattern
  Recognition}}, 2020.

\bibitem[Henrich et~al.(2010)Henrich, Heine, and Norenzayan]{henrich2010most}
Joseph Henrich, Steven~J Heine, and Ara Norenzayan.
\newblock Most people are not weird.
\newblock \emph{Nature}, 2010.

\bibitem[Hermann and Lampinen(2020)]{hermann2020shapes}
Katherine Hermann and Andrew Lampinen.
\newblock What shapes feature representations? exploring datasets,
  architectures, and training.
\newblock \emph{{Conference on Neural Information Processing Systems}}, 2020.

\bibitem[Higgins et~al.(2017)Higgins, Matthey, Pal, Burgess, Glorot, Botvinick,
  Mohamed, and Lerchner]{higgins2016betavae}
Irina Higgins, Lo{\"{\i}}c Matthey, Arka Pal, Christopher Burgess, Xavier
  Glorot, Matthew Botvinick, Shakir Mohamed, and Alexander Lerchner.
\newblock beta-vae: Learning basic visual concepts with a constrained
  variational framework.
\newblock In \emph{{International Conference on Learning Representations}},
  2017.

\bibitem[{Hu} et~al.(2018){Hu}, {Niu}, {Sato}, and {Sugiyama}]{groupdro0hu}
Weihua {Hu}, Gang {Niu}, Issei {Sato}, and Masashi {Sugiyama}.
\newblock Does distributionally robust supervised learning give robust
  classifiers.
\newblock In \emph{{International Conference on Machine Learning}}, 2018.

\bibitem[Huang et~al.(2017)Huang, Liu, van~der Maaten, and Weinberger]{dense}
Gao Huang, Zhuang Liu, Laurens van~der Maaten, and Kilian~Q. Weinberger.
\newblock Densely connected convolutional networks.
\newblock In \emph{{IEEE Conference on Computer Vision and Pattern
  Recognition}}, 2017.

\bibitem[Kalimeris et~al.(2019)Kalimeris, Kaplun, Nakkiran, Edelman, Yang,
  Barak, and Zhang]{kalimeris2019sgd}
Dimitris Kalimeris, Gal Kaplun, Preetum Nakkiran, Benjamin Edelman, Tristan
  Yang, Boaz Barak, and Haofeng Zhang.
\newblock Sgd on neural networks learns functions of increasing complexity.
\newblock \emph{{Conference on Neural Information Processing Systems}}, 2019.

\bibitem[Kang et~al.(2020)Kang, Xie, Rohrbach, Yan, Gordo, Feng, and
  Kalantidis]{kang2019decoupling}
Bingyi Kang, Saining Xie, Marcus Rohrbach, Zhicheng Yan, Albert Gordo, Jiashi
  Feng, and Yannis Kalantidis.
\newblock Decoupling representation and classifier for long-tailed recognition.
\newblock In \emph{{International Conference on Learning Representations}},
  2020.

\bibitem[Kim et~al.(2019)Kim, Kim, Kim, Kim, and Kim]{kim2019datasetbias2}
Byungju Kim, Hyunwoo Kim, Kyungsu Kim, Sungjin Kim, and Junmo Kim.
\newblock Learning not to learn: Training deep neural networks with biased
  data.
\newblock In \emph{{IEEE Conference on Computer Vision and Pattern
  Recognition}}, 2019.

\bibitem[Kingma and Welling(2014)]{kingma2013vae}
Diederik~P. Kingma and Max Welling.
\newblock Auto-encoding variational bayes.
\newblock In \emph{{International Conference on Learning Representations}},
  2014.

\bibitem[Kirichenko et~al.(2022)Kirichenko, Izmailov, and
  Wilson]{kirichenko2022last}
Polina Kirichenko, Pavel Izmailov, and Andrew~Gordon Wilson.
\newblock Last layer re-training is sufficient for robustness to spurious
  correlations.
\newblock \emph{arXiv:2204.02937}, 2022.

\bibitem[Koh et~al.(2021)Koh, Sagawa, Marklund, Xie, Zhang, Balsubramani, Hu,
  Yasunaga, Phillips, Gao, Lee, David, Stavness, Guo, Earnshaw, Haque, Beery,
  Leskovec, Kundaje, Pierson, Levine, Finn, and Liang]{koh2021wilds}
Pang~Wei Koh, Shiori Sagawa, Henrik Marklund, Sang~Michael Xie, Marvin Zhang,
  Akshay Balsubramani, Weihua Hu, Michihiro Yasunaga, Richard~Lanas Phillips,
  Irena Gao, Tony Lee, Etienne David, Ian Stavness, Wei Guo, Berton Earnshaw,
  Imran Haque, Sara~M. Beery, Jure Leskovec, Anshul Kundaje, Emma Pierson,
  Sergey Levine, Chelsea Finn, and Percy Liang.
\newblock {WILDS:} {A} benchmark of in-the-wild distribution shifts.
\newblock In \emph{{International Conference on Machine Learning}}, 2021.

\bibitem[{Koyama} and {Yamaguchi}(2021)]{koyama2021out}
Masanori {Koyama} and Shoichiro {Yamaguchi}.
\newblock Out-of-distribution generalization with maximal invariant predictor.
\newblock In \emph{arXiv e-prints}, 2021.

\bibitem[Lapuschkin et~al.(2019)Lapuschkin, W{\"a}ldchen, Binder, Montavon,
  Samek, and M{\"u}ller]{lapuschkin2019shortcut2}
Sebastian Lapuschkin, Stephan W{\"a}ldchen, Alexander Binder, Gr{\'e}goire
  Montavon, Wojciech Samek, and Klaus-Robert M{\"u}ller.
\newblock Unmasking clever hans predictors and assessing what machines really
  learn.
\newblock \emph{Nature communications}, 2019.

\bibitem[Le~Bras et~al.(2020)Le~Bras, Swayamdipta, Bhagavatula, Zellers,
  Peters, Sabharwal, and Choi]{le2020datasetbias3}
Ronan Le~Bras, Swabha Swayamdipta, Chandra Bhagavatula, Rowan Zellers, Matthew
  Peters, Ashish Sabharwal, and Yejin Choi.
\newblock Adversarial filters of dataset biases.
\newblock In \emph{{International Conference on Machine Learning}}, 2020.

\bibitem[Liu et~al.(2021)Liu, HaoChen, Gaidon, and Ma]{liu2021self}
Hong Liu, Jeff~Z HaoChen, Adrien Gaidon, and Tengyu Ma.
\newblock Self-supervised learning is more robust to dataset imbalance.
\newblock \emph{arXiv:2110.05025}, 2021.

\bibitem[Liu et~al.(2022)Liu, Xu, Xu, Qian, Li, Jin, Ji, and
  Chan]{liu2022empirical}
Ziquan Liu, Yi~Xu, Yuanhong Xu, Qi~Qian, Hao Li, Rong Jin, Xiangyang Ji, and
  Antoni~B Chan.
\newblock An empirical study on distribution shift robustness from the
  perspective of pre-training and data augmentation.
\newblock \emph{arXiv:2205.12753}, 2022.

\bibitem[{Lopez-Paz} and {Ranzato}(2017)]{lopez-paz2017gradient}
David {Lopez-Paz} and Marc'Aurelio {Ranzato}.
\newblock Gradient episodic memory for continual learning.
\newblock In \emph{{Conference on Neural Information Processing Systems}},
  2017.

\bibitem[Luo et~al.(2021)Luo, Wei, Wen, Yang, Xie, Xu, and
  Tian]{luo2021rectifying}
Xu~Luo, Longhui Wei, Liangjian Wen, Jinrong Yang, Lingxi Xie, Zenglin Xu, and
  Qi~Tian.
\newblock Rectifying the shortcut learning of background for few-shot learning.
\newblock \emph{{Conference on Neural Information Processing Systems}}, 2021.

\bibitem[Matthey et~al.(2017)Matthey, Higgins, Hassabis, and
  Lerchner]{dsprites}
Loic Matthey, Irina Higgins, Demis Hassabis, and Alexander Lerchner.
\newblock dsprites: Disentanglement testing sprites dataset.
\newblock https://github.com/deepmind/dsprites-dataset/, 2017.

\bibitem[Menon et~al.(2020)Menon, Rawat, and
  Kumar]{menon2020overparameterisation}
Aditya~Krishna Menon, Ankit~Singh Rawat, and Sanjiv Kumar.
\newblock Overparameterisation and worst-case generalisation: friend or foe?
\newblock In \emph{{International Conference on Learning Representations}},
  2020.

\bibitem[Ramesh et~al.(2021)Ramesh, Pavlov, Goh, Gray, Voss, Radford, Chen, and
  Sutskever]{ramesh2021zero}
Aditya Ramesh, Mikhail Pavlov, Gabriel Goh, Scott Gray, Chelsea Voss, Alec
  Radford, Mark Chen, and Ilya Sutskever.
\newblock Zero-shot text-to-image generation.
\newblock In \emph{{International Conference on Machine Learning}}, 2021.

\bibitem[Robinson et~al.(2021)Robinson, Sun, Yu, Batmanghelich, Jegelka, and
  Sra]{robinson2021simplicity}
Joshua Robinson, Li~Sun, Ke~Yu, Kayhan Batmanghelich, Stefanie Jegelka, and
  Suvrit Sra.
\newblock Can contrastive learning avoid shortcut solutions?
\newblock \emph{{Conference on Neural Information Processing Systems}}, 2021.

\bibitem[Rumelhart et~al.(1985)Rumelhart, Hinton, and
  Williams]{rumelhart1985ae}
David~E Rumelhart, Geoffrey~E Hinton, and Ronald~J Williams.
\newblock Learning internal representations by error propagation.
\newblock Technical report, California Univ San Diego La Jolla Inst for
  Cognitive Science, 1985.

\bibitem[Sagawa et~al.(2020)Sagawa, Koh, Hashimoto, and Liang]{Sagawa2020dro}
Shiori Sagawa, Pang~Wei Koh, Tatsunori~B. Hashimoto, and Percy Liang.
\newblock Distributionally robust neural networks.
\newblock In \emph{{International Conference on Learning Representations}},
  2020.

\bibitem[Shah et~al.(2020)Shah, Tamuly, Raghunathan, Jain, and
  Netrapalli]{shah2020pitfalls}
Harshay Shah, Kaustav Tamuly, Aditi Raghunathan, Prateek Jain, and Praneeth
  Netrapalli.
\newblock The pitfalls of simplicity bias in neural networks.
\newblock In \emph{{Conference on Neural Information Processing Systems}},
  2020.

\bibitem[Shi et~al.(2022)Shi, Seely, Torr, Siddharth, Hannun, Usunier, and
  Synnaeve]{fish}
Yuge Shi, Jeffrey Seely, Philip~HS Torr, N~Siddharth, Awni Hannun, Nicolas
  Usunier, and Gabriel Synnaeve.
\newblock Gradient matching for domain generalization.
\newblock In \emph{{International Conference on Learning Representations}},
  2022.

\bibitem[Sun and Saenko(2016)]{sun2016coral}
Baochen Sun and Kate Saenko.
\newblock Deep coral: Correlation alignment for deep domain adaptation.
\newblock In \emph{European conference on computer vision}, pages 443--450.
  Springer, 2016.

\bibitem[Taori et~al.(2020)Taori, Dave, Shankar, Carlini, Recht, and
  Schmidt]{taori2020measuring}
Rohan Taori, Achal Dave, Vaishaal Shankar, Nicholas Carlini, Benjamin Recht,
  and Ludwig Schmidt.
\newblock Measuring robustness to natural distribution shifts in image
  classification.
\newblock \emph{Advances in Neural Information Processing Systems},
  33:\penalty0 18583--18599, 2020.

\bibitem[Teney et~al.(2022)Teney, Abbasnejad, Lucey, and van~den
  Hengel]{teney2022evading}
Damien Teney, Ehsan Abbasnejad, Simon Lucey, and Anton van~den Hengel.
\newblock Evading the simplicity bias: Training a diverse set of models
  discovers solutions with superior ood generalization.
\newblock In \emph{Proceedings of the IEEE/CVF Conference on Computer Vision
  and Pattern Recognition}, pages 16761--16772, 2022.

\bibitem[Torralba and Efros(2011)]{torralba2011datasetbias1}
Antonio Torralba and Alexei~A Efros.
\newblock Unbiased look at dataset bias.
\newblock In \emph{CVPR 2011}, pages 1521--1528. IEEE, 2011.

\bibitem[{Wang} et~al.(2019){Wang}, {Ge}, {Lipton}, and
  {Xing}]{wang2019learning}
Haohan {Wang}, Songwei {Ge}, Zachary~C. {Lipton}, and Eric~P. {Xing}.
\newblock Learning robust global representations by penalizing local predictive
  power.
\newblock In \emph{Advances in Neural Information Processing Systems},
  volume~32, pages 10506--10518, 2019.

\bibitem[Zhang et~al.(2022)Zhang, Zhou, Xu, Cui, Shen, and
  Liu]{zhang2022towards}
Xingxuan Zhang, Linjun Zhou, Renzhe Xu, Peng Cui, Zheyan Shen, and Haoxin Liu.
\newblock Towards unsupervised domain generalization.
\newblock In \emph{Proceedings of the IEEE/CVF Conference on Computer Vision
  and Pattern Recognition}, pages 4910--4920, 2022.

\end{thebibliography}
\ifthenelse{\equal{\paperversion}{arxiv}}
{\clearpage
\appendix

\section{Additional Experimental Results} \label{app:additional_results}
\subsection{CdSprites}\label{app:cdsprites_additional}

% from main text:

In addition to our results in \cref{tab:cdsprites_results}, where we use a dataset with perfectly correlated features ($r_{\text{id}} = 1$) to train the backbones, in  \cref{fig:cdsprites_individual} we vary $r_{\text{id}}$ to analyse the effect of imperfectly correlated features. Notably, with imperfect correlation ($r_{\text{id}} < 1$), the OOD linear heads trained on top of the \gls{SL} and \gls{SSL} backbones perform perfectly. For the \gls{AE}, we observe that the performance of the OOD linear head does not depend on the correlation $r_{\text{id}}$ in the data used to train the backbones. Our results suggest that with imperfect correlation between features, \gls{SL} and \gls{SSL} models learn a linearly separable representation of the features, whereas \gls{AE} does not.

In \cref{fig:cdsprites_corrx_oody} we provide an ablation where we also vary $r_{\text{ood}}$, the correlation in the data used to train and evaluate the linear head.~\Cref{subfig:cdsprites_corrx_oody_sup,subfig:cdsprites_corrx_oody_ssl} corroborate our results that \gls{SSL} performs on par with \gls{SL} for $r_{\text{id}}<1$ and strictly better when $r_{\text{id}}=1$.
For the \gls{AE} (\Cref{subfig:cdsprites_corrx_oody_ae}), we observe an interesting pattern where the performance of the OOD linear head depends on the OOD correlation $r_{\text{ood}}$, but not on the correlation $r_{\text{id}}$ in the data used to train the backbones. Hence, the ablation corroborates our result that \gls{SL} and \gls{SSL} models learn a linearly separable representation of the shape and color features when there is an imperfect correlation between the features, whereas \gls{AE} does not.

% \begin{figure*}[t]
%   \centering
%   \captionsetup[subfigure]{belowskip=1ex}
%   \begin{subfigure}{0.48\linewidth}
%     \centering
%     \includegraphics[width=0.49\linewidth]{images/cdsprites/id_color_v.pdf}
%     \includegraphics[width=0.49\linewidth]{images/cdsprites/ood_color_v.pdf}
%     \caption{Color classification, OOD acc (\%).}\label{fig:cdsprites_color}
%   \end{subfigure} \vspace{20pt}
%   \begin{subfigure}{0.48\linewidth}
%     \centering
%     \includegraphics[width=0.49\linewidth]{images/cdsprites/id_shape_v.pdf}
%     \includegraphics[width=0.49\linewidth]{images/cdsprites/ood_shape_v.pdf}
%     \caption{Shape classification, OOD acc (\%).}\label{fig:cdsprites_shape}
%   \end{subfigure}
%   \vspace{-3em}
%   \caption{Individual model results on CdSprites. Blue colors are \gls{AE} models, green colors are \gls{SSL} models and grey is \gls{SL}. {\textbf{Left:}} Color classification accuracy on \gls{OOD} test set with linear head trained on ID data; {\textbf{Right:}} Shape classification accuracy on \gls{OOD} test set with linear head trained on OOD data.
%   }
%   \vspace{1em}
%   \label{fig:cdsprites_individual}
% \end{figure*}

\begin{figure*}[t]
  \centering
  \captionsetup[subfigure]{belowskip=1ex}
  \begin{subfigure}{0.32\linewidth}
    \centering
    \includegraphics[width=\linewidth]{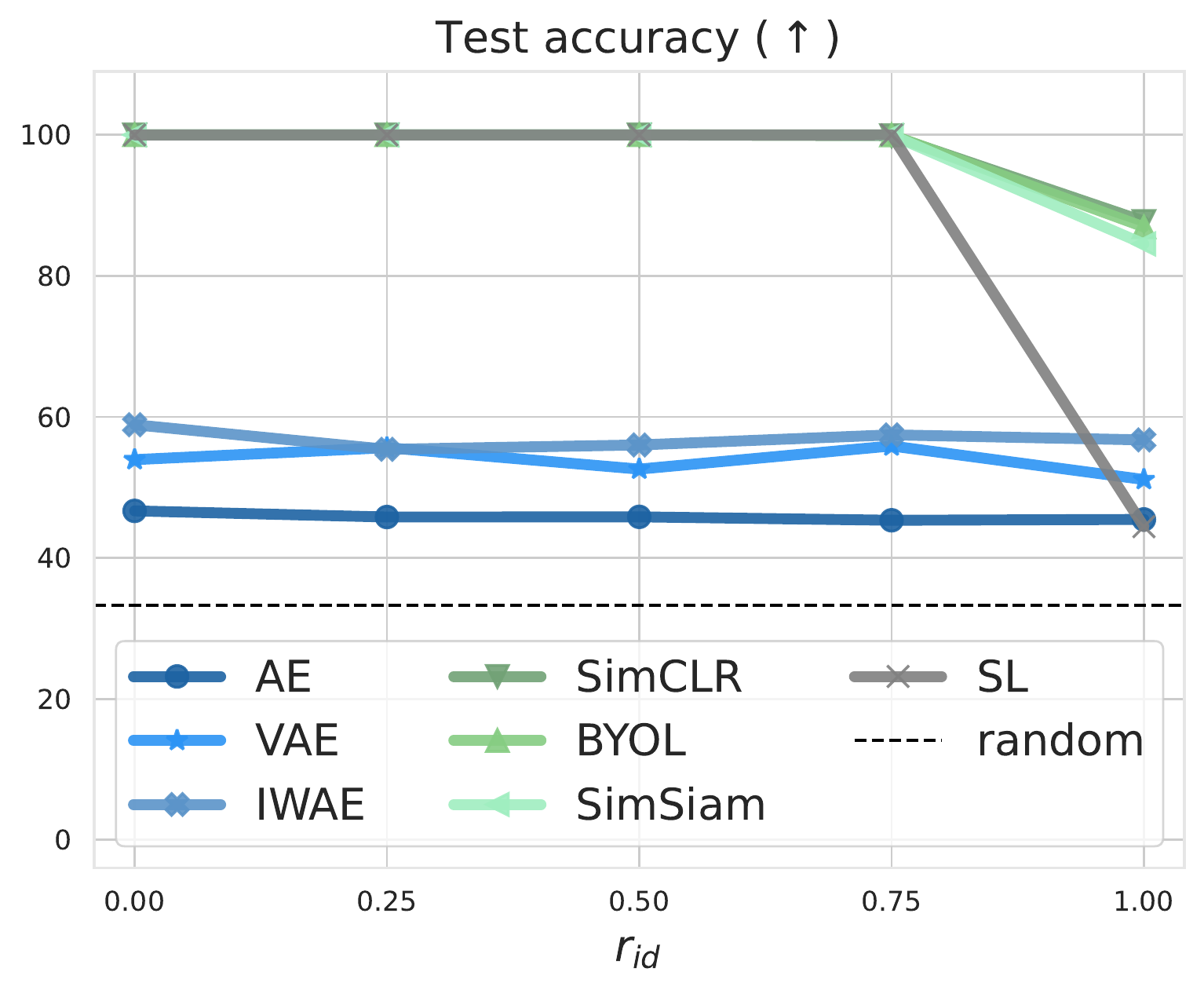}
    \caption{\aoco  ~(\%).}\label{fig:cdsprites_aoco_individual}
  \end{subfigure} 
  \begin{subfigure}{0.32\linewidth}
    \centering
    \includegraphics[width=\linewidth]{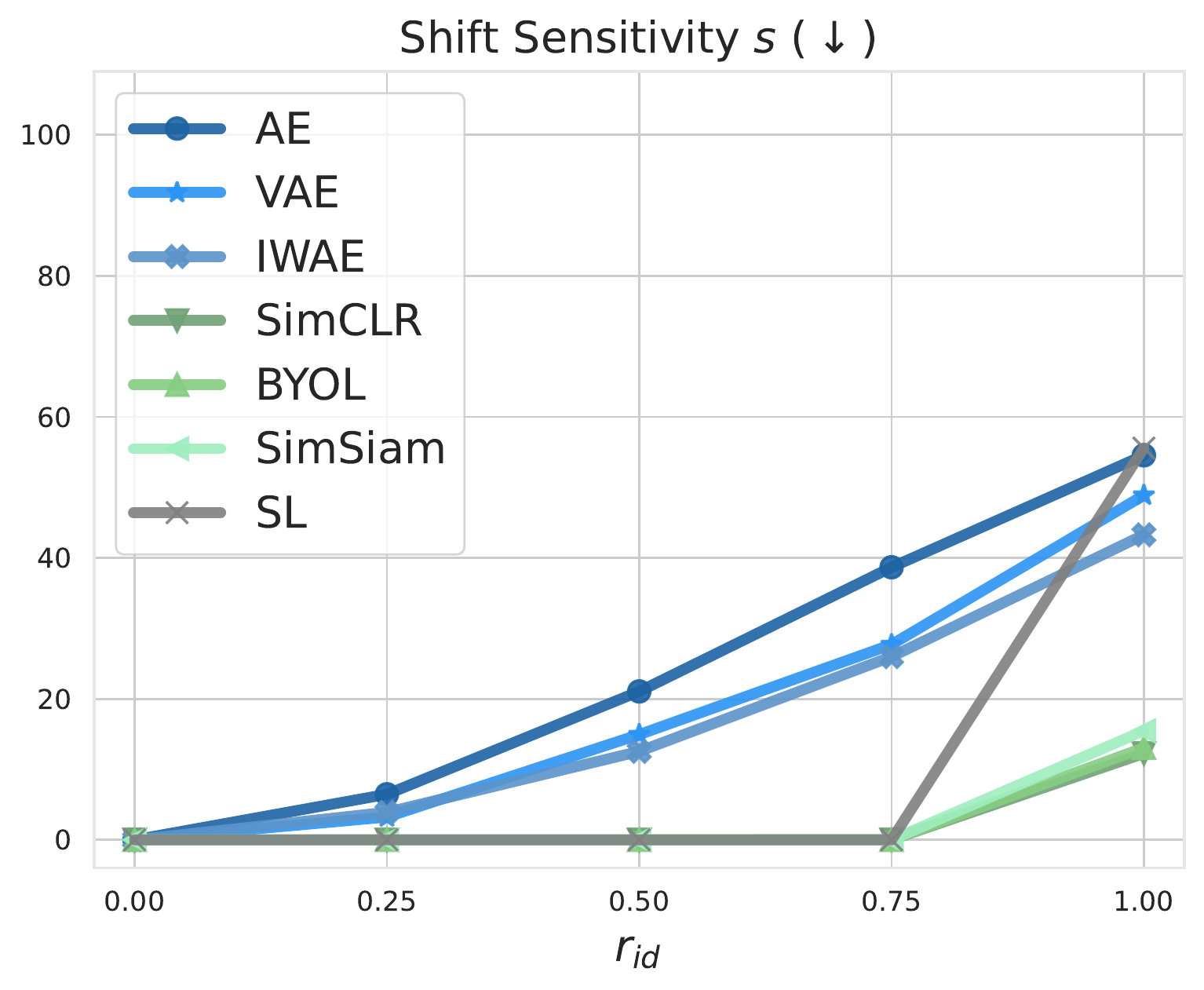}
    \caption{Shift sensitivity $s$  (\%).}\label{fig:cdsprites_rho_individual}
  \end{subfigure}
  \begin{subfigure}{0.32\linewidth}
    \centering
    \includegraphics[width=\linewidth]{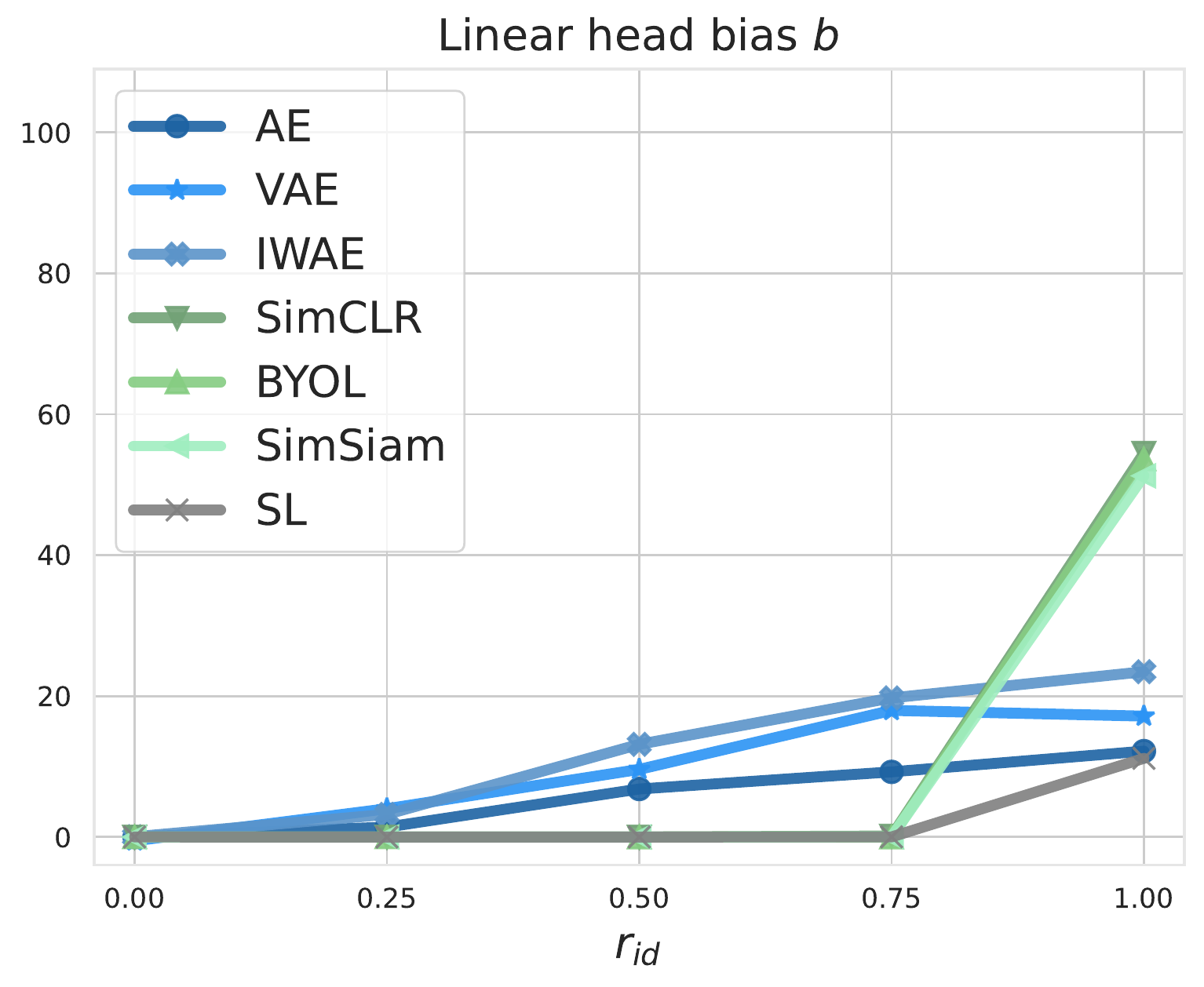}
    \caption{Linear head bias $b$ (\%).}\label{fig:cdsprites_b_individual}
  \end{subfigure}
  \vspace{-10pt}
  \caption{Evaluations on the CdSprites dataset with $r_{\text{id}}\in \{0.25, 0.5, 0.75, 1.0\}$. We report shape classification accuracy using OOD-trained linear head ($\text{acc}_o(f,c_o)$), shift sensitivity $s$, and linear head bias $b$. Results are shown for \emph{individual} models from the class of \gls{AE} (blue), \gls{SSL} (green), and \gls{SL} (grey) algorithms. The black horizontal line denotes the random baseline (33.3\% for three classes).}
  % Shape classification accuracy as a function of the correlation between shape and color features ($r_{\text{id}}$, x-axis) in the ID train split used to pre-train the respective backbone. Linear heads were trained on top of the frozen backbones using the OOD train split. 
%   \vspace{-12pt}
  \label{fig:cdsprites_individual}
\end{figure*}

\begin{figure}[H]
     \centering
     \begin{subfigure}[b]{0.32\linewidth}
         \centering
         \includegraphics[width=\linewidth]{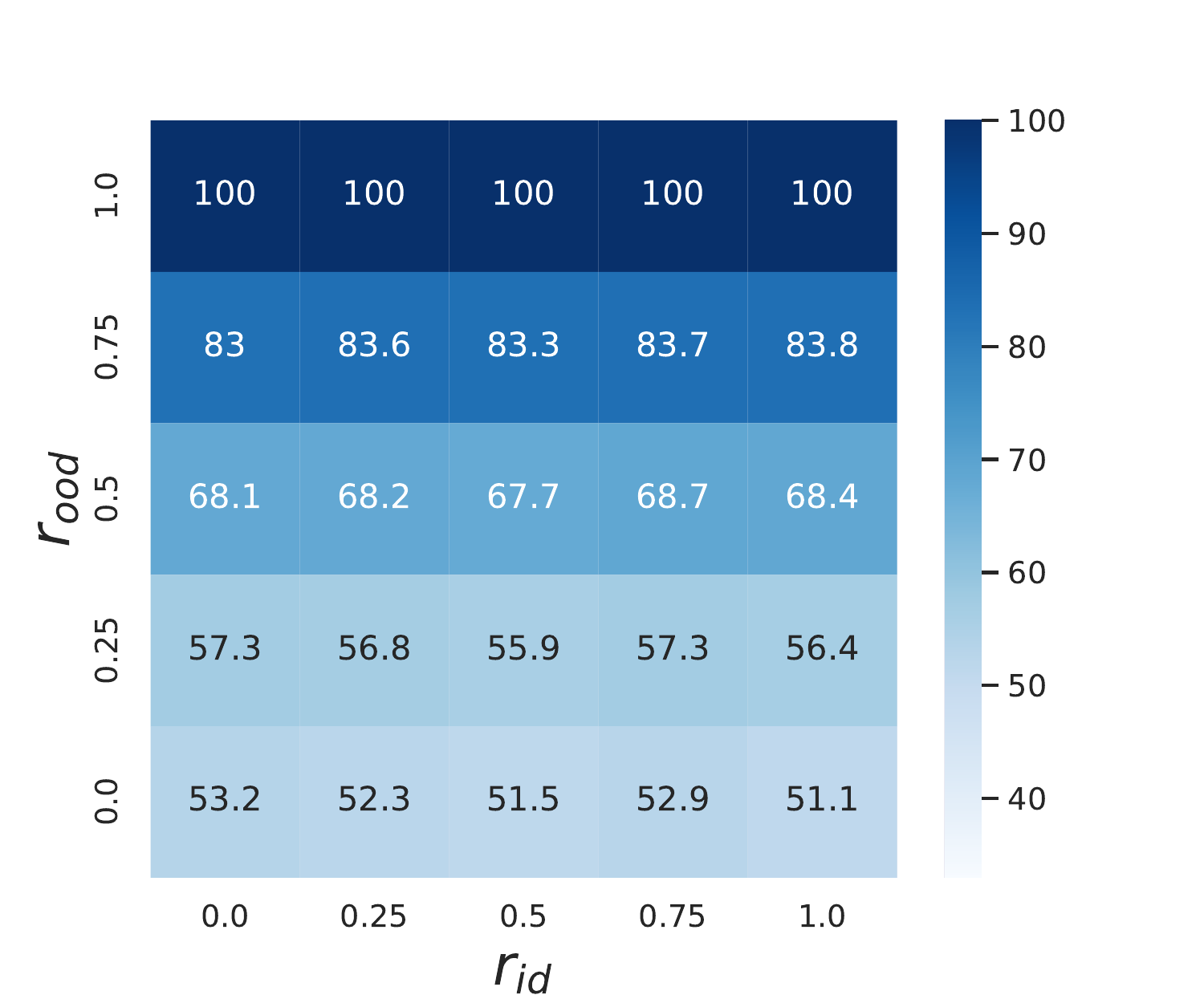}
         \caption{AE}
         \label{subfig:cdsprites_corrx_oody_ae}
     \end{subfigure}
     \begin{subfigure}[b]{0.32\linewidth}
         \centering
         \includegraphics[width=\linewidth]{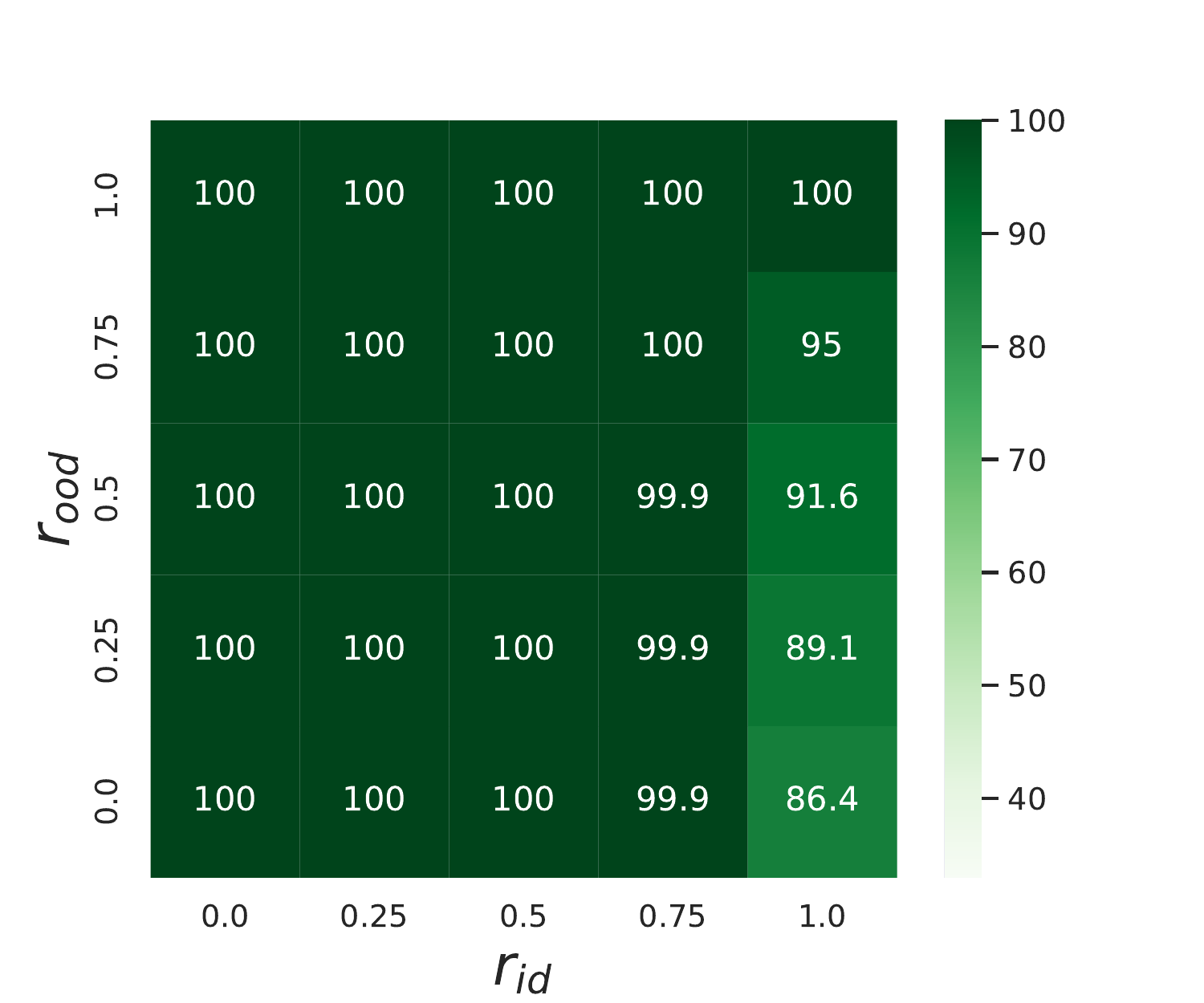}
         \caption{SSL}
         \label{subfig:cdsprites_corrx_oody_ssl}
     \end{subfigure}
     \begin{subfigure}[b]{0.32\linewidth}
         \centering
         \includegraphics[width=\linewidth]{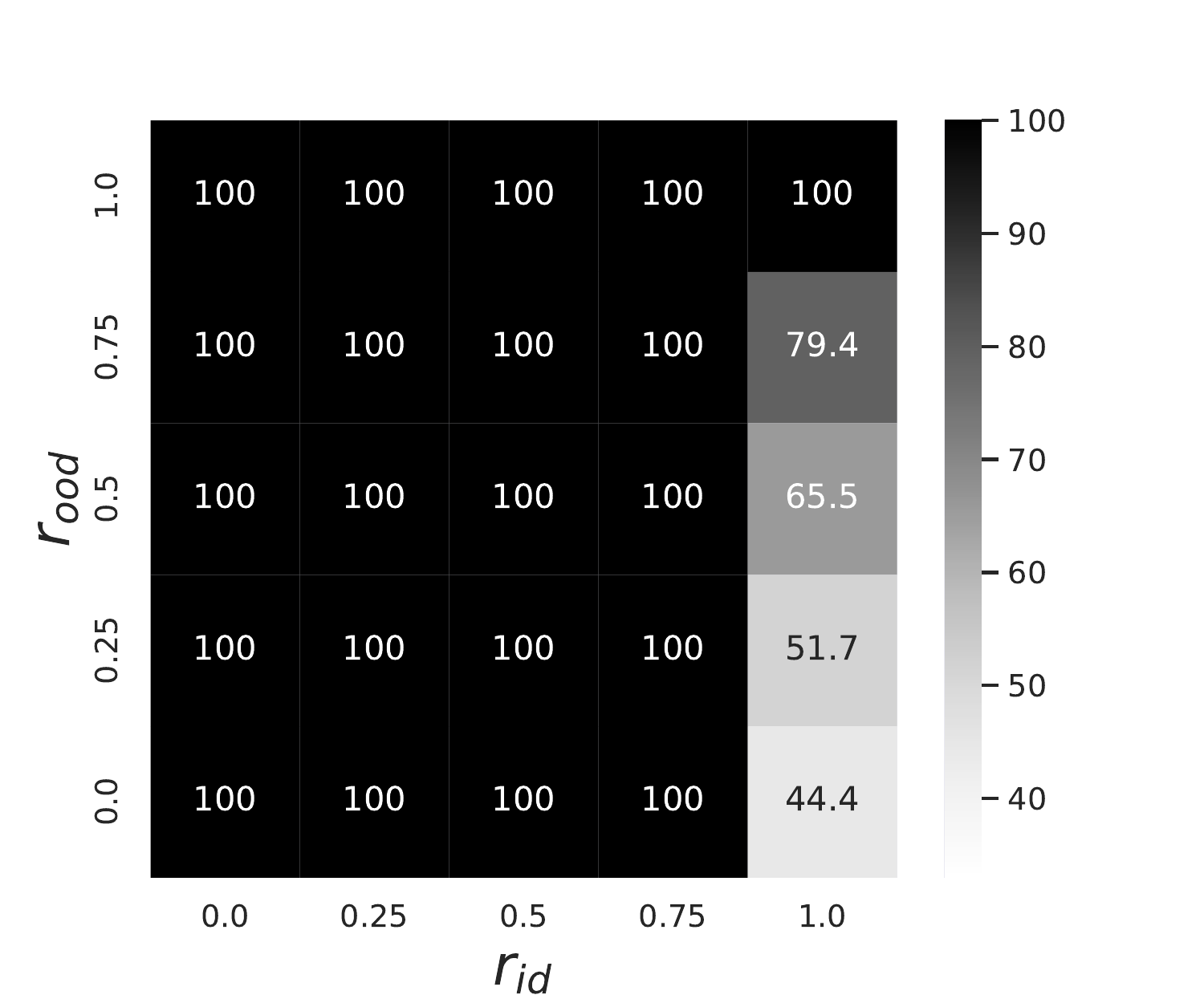}
         \caption{SL}
         \label{subfig:cdsprites_corrx_oody_sup}
     \end{subfigure}
     \caption{Correlation coefficient ablation for CdSprites. Shape classification accuracy for the CdSprites experiment with varying correlation of the ID training data ($r_{\text{id}}$, x-axis) and OOD training and test data ($r_{\text{ood}}$, y-axis). Backbones were trained on data with correlation $r_{\text{id}}$ and linear classifiers trained and evaluated on top of the frozen backbones with correlation $r_{\text{ood}}$.}
     \label{fig:cdsprites_corrx_oody}
\end{figure}

}
{\clearpage
\appendix

\section{Additional Experimental Results} \label{app:additional_results}
\subsection{CdSprites}\label{app:cdsprites_additional}

% from main text:

In addition to our results in \cref{tab:cdsprites_results}, where we use a dataset with perfectly correlated features ($r_{\text{id}} = 1$) to train the backbones, in  \cref{fig:cdsprites_individual} we vary $r_{\text{id}}$ to analyse the effect of imperfectly correlated features. Notably, with imperfect correlation ($r_{\text{id}} < 1$), the OOD linear heads trained on top of the \gls{SL} and \gls{SSL} backbones perform perfectly. For the \gls{AE}, we observe that the performance of the OOD linear head does not depend on the correlation $r_{\text{id}}$ in the data used to train the backbones. Our results suggest that with imperfect correlation between features, \gls{SL} and \gls{SSL} models learn a linearly separable representation of the features, whereas \gls{AE} does not.

In \cref{fig:cdsprites_corrx_oody} we provide an ablation where we also vary $r_{\text{ood}}$, the correlation in the data used to train and evaluate the linear head.~\Cref{subfig:cdsprites_corrx_oody_sup,subfig:cdsprites_corrx_oody_ssl} corroborate our results that \gls{SSL} performs on par with \gls{SL} for $r_{\text{id}}<1$ and strictly better when $r_{\text{id}}=1$.
For the \gls{AE} (\Cref{subfig:cdsprites_corrx_oody_ae}), we observe an interesting pattern where the performance of the OOD linear head depends on the OOD correlation $r_{\text{ood}}$, but not on the correlation $r_{\text{id}}$ in the data used to train the backbones. Hence, the ablation corroborates our result that \gls{SL} and \gls{SSL} models learn a linearly separable representation of the shape and color features when there is an imperfect correlation between the features, whereas \gls{AE} does not.

% \begin{figure*}[t]
%   \centering
%   \captionsetup[subfigure]{belowskip=1ex}
%   \begin{subfigure}{0.48\linewidth}
%     \centering
%     \includegraphics[width=0.49\linewidth]{images/cdsprites/id_color_v.pdf}
%     \includegraphics[width=0.49\linewidth]{images/cdsprites/ood_color_v.pdf}
%     \caption{Color classification, OOD acc (\%).}\label{fig:cdsprites_color}
%   \end{subfigure} \vspace{20pt}
%   \begin{subfigure}{0.48\linewidth}
%     \centering
%     \includegraphics[width=0.49\linewidth]{images/cdsprites/id_shape_v.pdf}
%     \includegraphics[width=0.49\linewidth]{images/cdsprites/ood_shape_v.pdf}
%     \caption{Shape classification, OOD acc (\%).}\label{fig:cdsprites_shape}
%   \end{subfigure}
%   \vspace{-3em}
%   \caption{Individual model results on CdSprites. Blue colors are \gls{AE} models, green colors are \gls{SSL} models and grey is \gls{SL}. {\textbf{Left:}} Color classification accuracy on \gls{OOD} test set with linear head trained on ID data; {\textbf{Right:}} Shape classification accuracy on \gls{OOD} test set with linear head trained on OOD data.
%   }
%   \vspace{1em}
%   \label{fig:cdsprites_individual}
% \end{figure*}

\begin{figure*}[t]
  \centering
  \captionsetup[subfigure]{belowskip=1ex}
  \begin{subfigure}{0.32\linewidth}
    \centering
    \includegraphics[width=\linewidth]{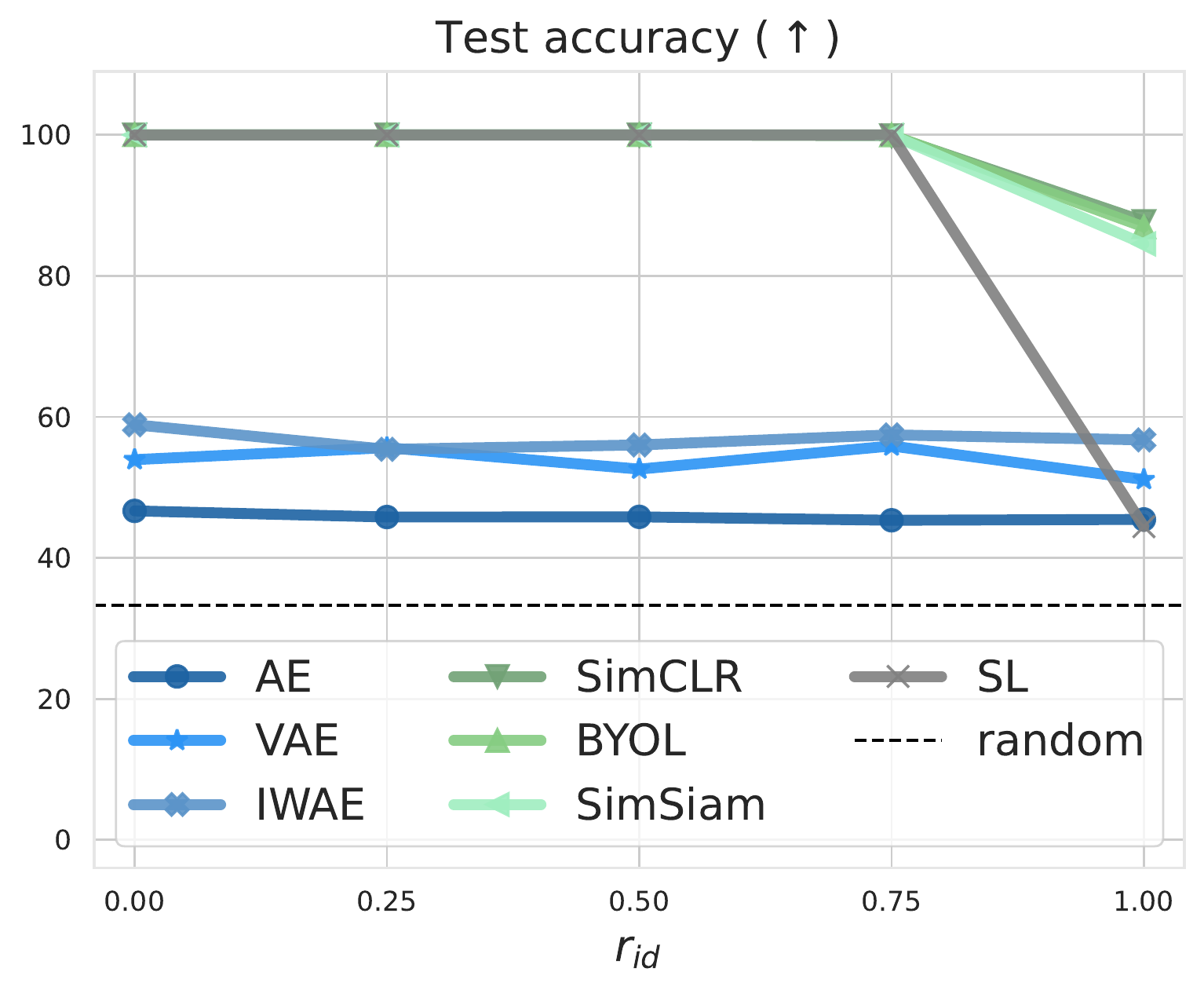}
    \caption{\aoco  ~(\%).}\label{fig:cdsprites_aoco_individual}
  \end{subfigure} 
  \begin{subfigure}{0.32\linewidth}
    \centering
    \includegraphics[width=\linewidth]{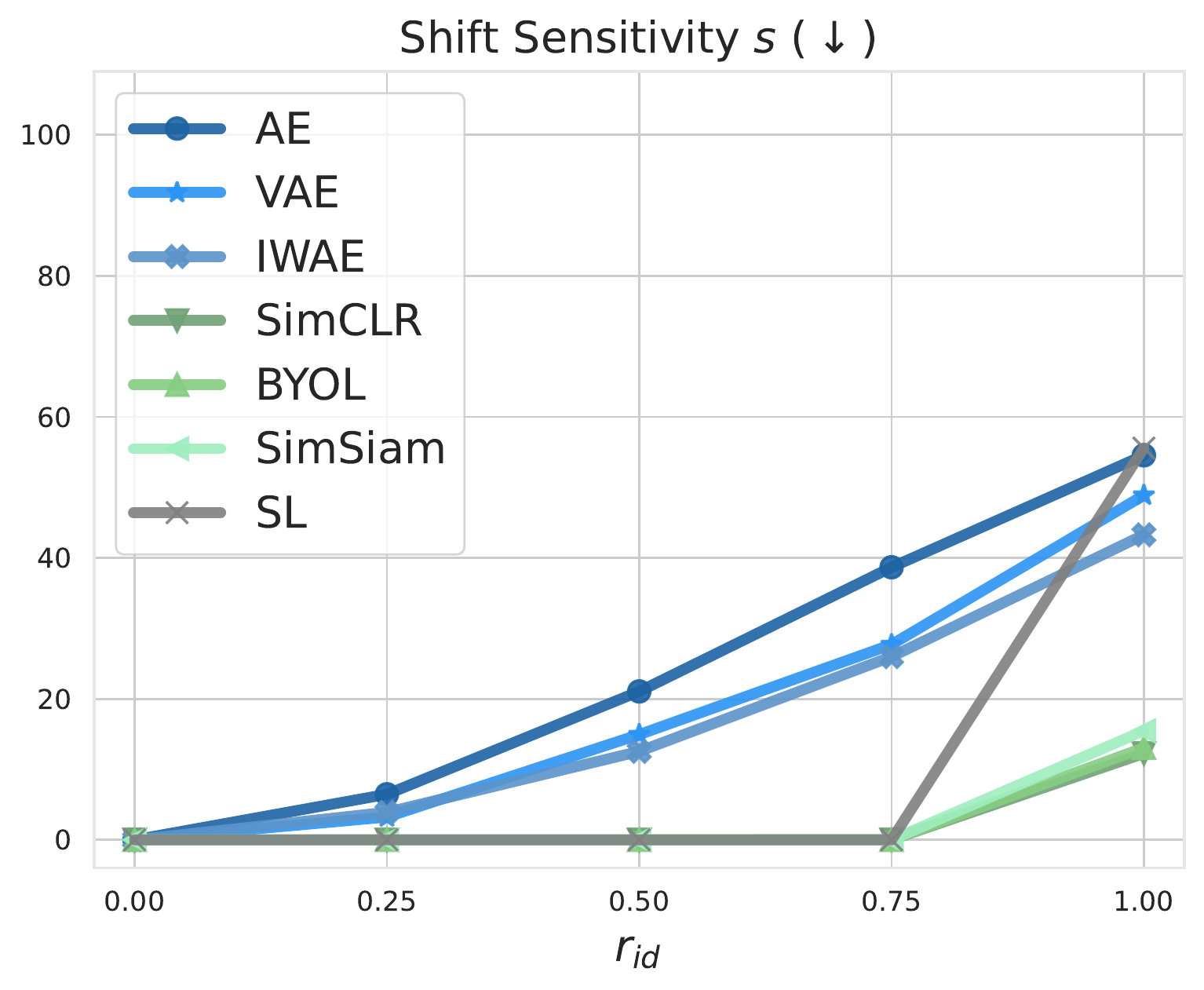}
    \caption{Shift sensitivity $s$  (\%).}\label{fig:cdsprites_rho_individual}
  \end{subfigure}
  \begin{subfigure}{0.32\linewidth}
    \centering
    \includegraphics[width=\linewidth]{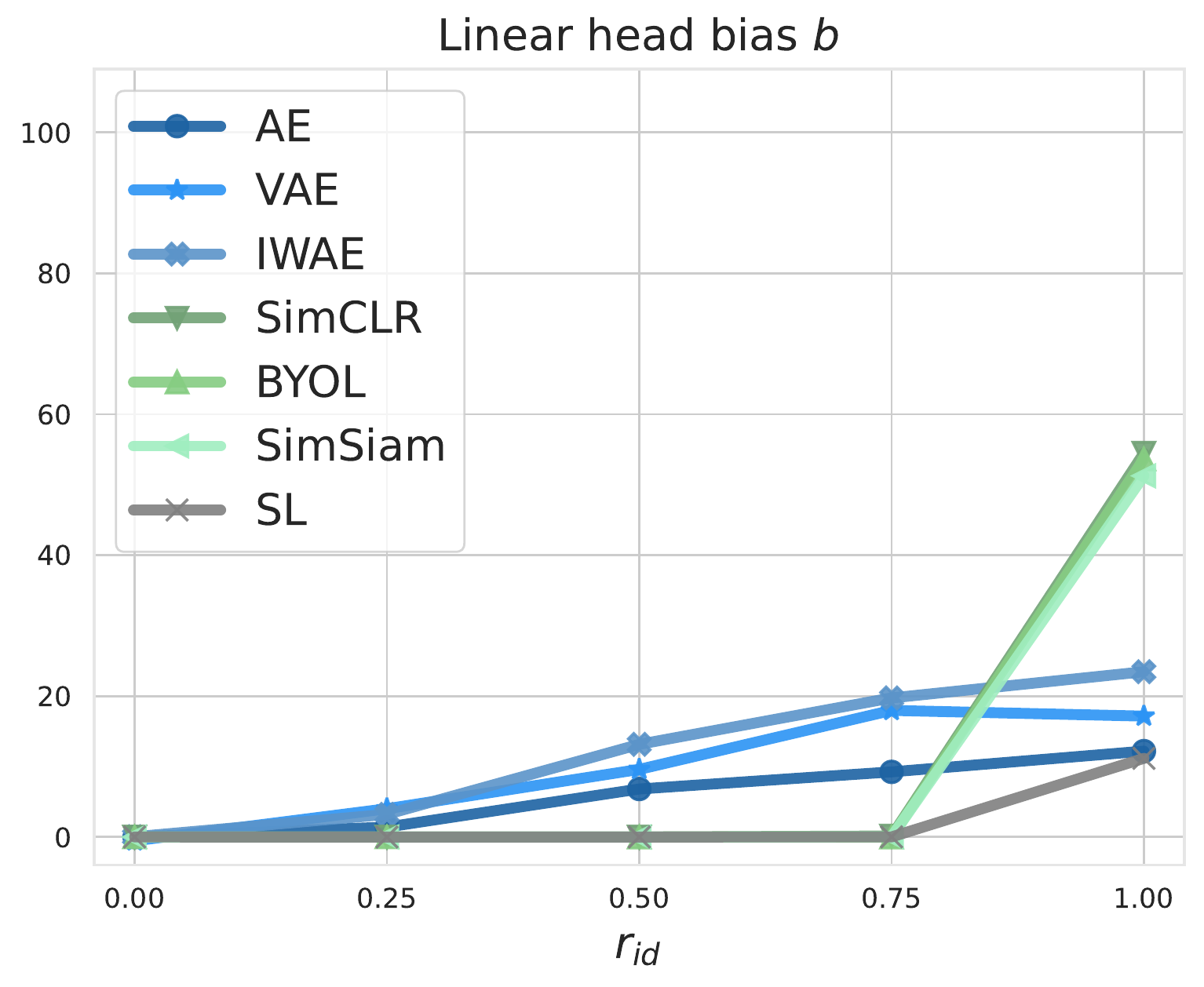}
    \caption{Linear head bias $b$ (\%).}\label{fig:cdsprites_b_individual}
  \end{subfigure}
  \vspace{-10pt}
  \caption{Evaluations on the CdSprites dataset with $r_{\text{id}}\in \{0.25, 0.5, 0.75, 1.0\}$. We report shape classification accuracy using OOD-trained linear head ($\text{acc}_o(f,c_o)$), shift sensitivity $s$, and linear head bias $b$. Results are shown for \emph{individual} models from the class of \gls{AE} (blue), \gls{SSL} (green), and \gls{SL} (grey) algorithms. The black horizontal line denotes the random baseline (33.3\% for three classes).}
  % Shape classification accuracy as a function of the correlation between shape and color features ($r_{\text{id}}$, x-axis) in the ID train split used to pre-train the respective backbone. Linear heads were trained on top of the frozen backbones using the OOD train split. 
%   \vspace{-12pt}
  \label{fig:cdsprites_individual}
\end{figure*}

\begin{figure}[H]
     \centering
     \begin{subfigure}[b]{0.32\linewidth}
         \centering
         \includegraphics[width=\linewidth]{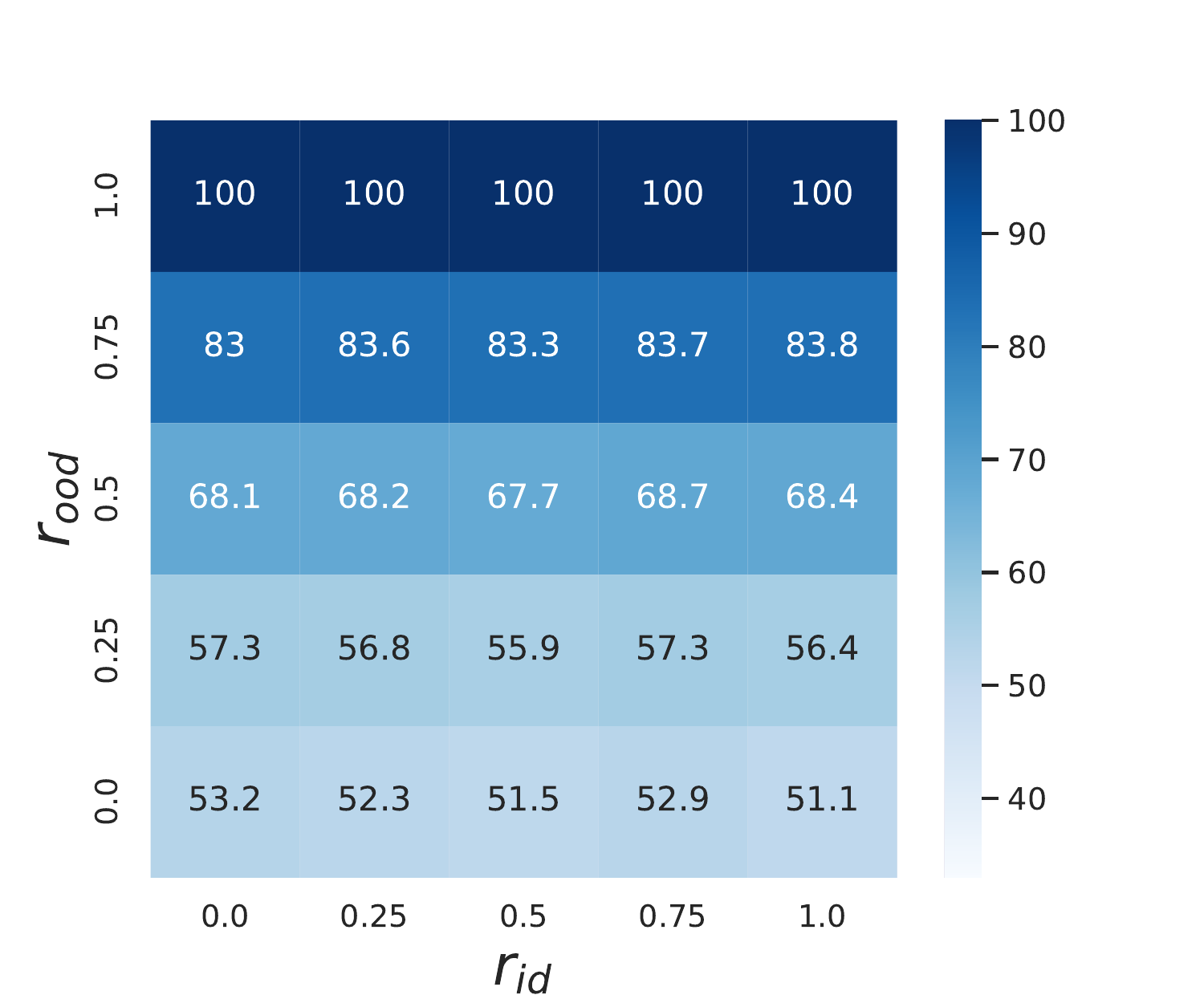}
         \caption{AE}
         \label{subfig:cdsprites_corrx_oody_ae}
     \end{subfigure}
     \begin{subfigure}[b]{0.32\linewidth}
         \centering
         \includegraphics[width=\linewidth]{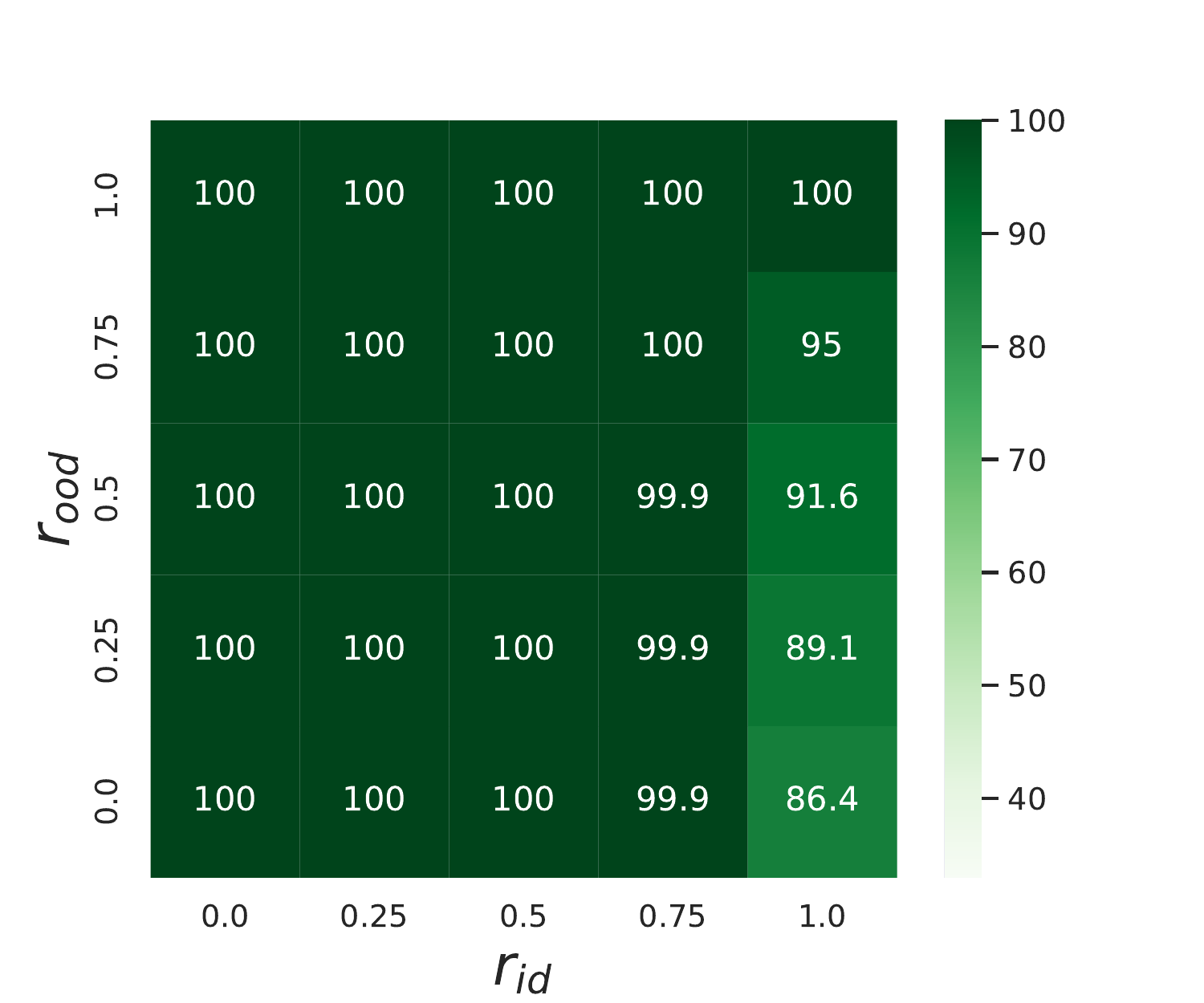}
         \caption{SSL}
         \label{subfig:cdsprites_corrx_oody_ssl}
     \end{subfigure}
     \begin{subfigure}[b]{0.32\linewidth}
         \centering
         \includegraphics[width=\linewidth]{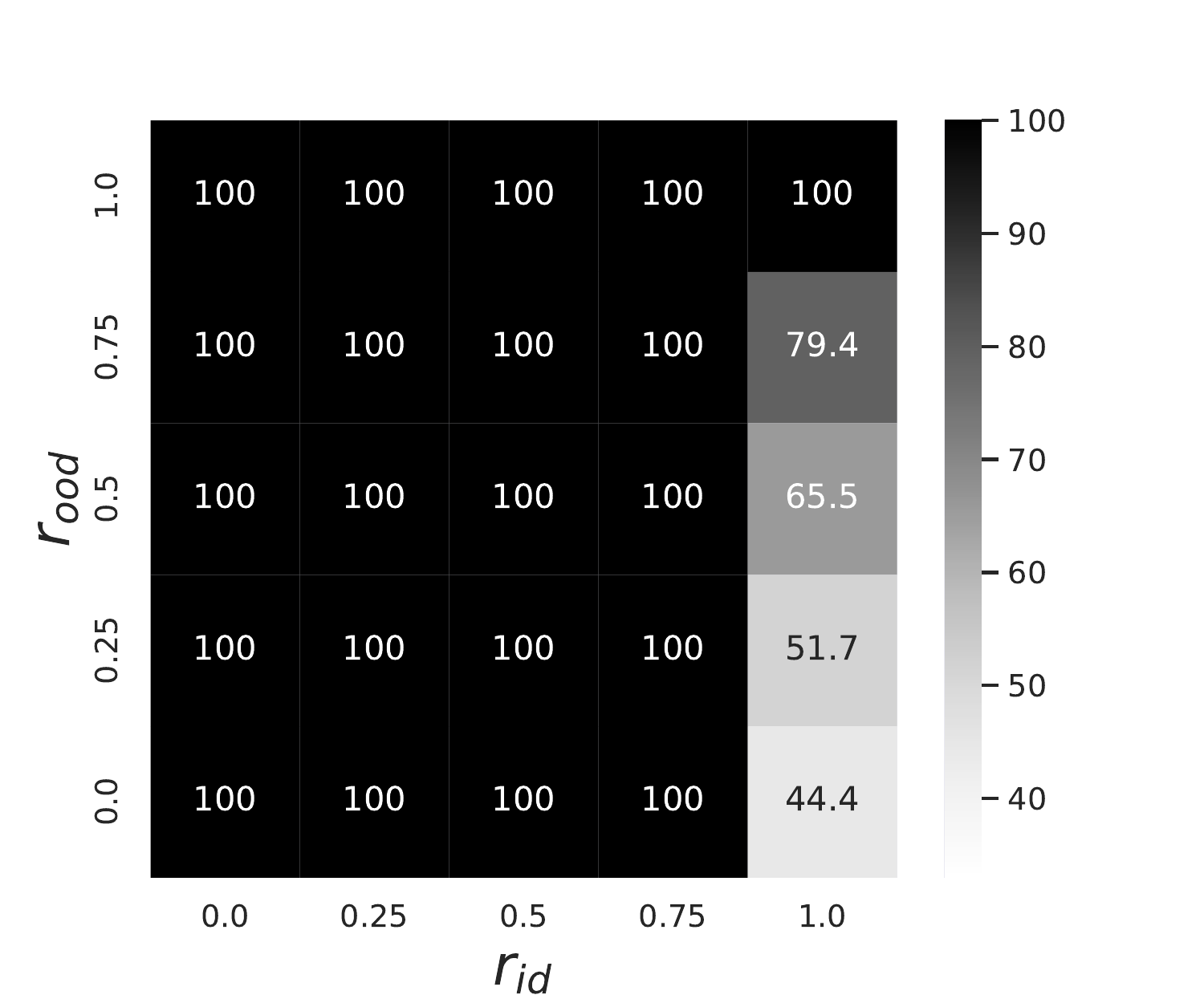}
         \caption{SL}
         \label{subfig:cdsprites_corrx_oody_sup}
     \end{subfigure}
     \caption{Correlation coefficient ablation for CdSprites. Shape classification accuracy for the CdSprites experiment with varying correlation of the ID training data ($r_{\text{id}}$, x-axis) and OOD training and test data ($r_{\text{ood}}$, y-axis). Backbones were trained on data with correlation $r_{\text{id}}$ and linear classifiers trained and evaluated on top of the frozen backbones with correlation $r_{\text{ood}}$.}
     \label{fig:cdsprites_corrx_oody}
\end{figure}

\subsection{Camelyon17-C and FMoW-C}\label{app:wilds_additional}

In this section we report the numerical results for Camelyon17-C and FMoW-C with $r_{\text{id}}=0.5$ (see \cref{tab:camelyon17corr50_results,tab:fmowcorr50_results}) and $r_{\text{id}}=1$ (see \cref{tab:camelyon17corr100_results,tab:fmowcorr100_results}).

\begin{table}
\parbox{.48\linewidth}{
\centering
\caption{Evaluations on test set of Camelyon17-C with $r_{\text{id}}=0.5$, all metrics computed using average accuracy.}
\vspace*{-0.8\baselineskip}
\scalebox{0.7}{
\begin{tabular}{llccc}    \toprule
    \multirow{2}{*}{Regime} &\multirow{2}{*}{Method} & \multicolumn{3}{c}{Metrics (\%)} \\
    \cmidrule{3-5} 
    & &  $\text{acc}_o(f,c_o)\uparrow$ & $s\downarrow$ & $b$ \\
    \midrule \rowcolor{lb}
    & AE          & 80.4	\std{3e+0} &	6.0	\std{2e+0} &	19.0	\std{4e+0} \\  \rowcolor{lb}
    & VAE         & 88.6	\std{2e+0} & -0.5	\std{1e+0} &	17.8	\std{6e+0} \\  \rowcolor{lb}
    & IWAE        & 87.8	\std{1e+0} & -0.2	\std{1e+0} &	26.4	\std{6e+0} \\  \rowcolor{lb}
    \multirow{-4}{*}{AE}
    & $\beta$-VAE & 88.5	\std{2e+0} &	0.1	\std{9e-1} &	19.7	\std{6e+0} \\  \rowcolor{db}
    \multicolumn{2}{l}{\emph{\textbf{AE average}}} & \emph{86.3	\std{2e+0}} &	\emph{1.4	\std{1e+0}} &	\emph{20.7	\std{5e+0}} \\
    \midrule \rowcolor{lg}
    & SimCLR      & 84.5	\std{2e+0} &	8.0	\std{1e+0} &	6.6 	\std{2e+0} \\ \rowcolor{lg}
    & SimSiam     & 86.1	\std{2e+0} &	5.7	\std{1e+0} &	8.3 	\std{4e+0}\\  \rowcolor{lg}
    \multirow{-3}{*}{SSL} 
    & BYOL        & 86.4	\std{2e+0} &	4.5	\std{2e+0} &	8.8 	\std{4e+0}\\  \rowcolor{dg}
    \multicolumn{2}{l}{\emph{\textbf{SSL average}}} & \emph{85.7	\std{2e+0}} &	\emph{6.1	\std{2e+0}} &	\emph{7.9 	\std{3e+0}} \\
    \midrule \rowcolor{g}
    SL
    & Supervised  & 81.5	\std{5e+0} & 13.2	\std{3e+0} &	3.4 	\std{4e+0}\\ 
    \bottomrule
\end{tabular}
\label{tab:camelyon17corr50_results}}
}
\hspace{10pt}
\parbox{.48\linewidth}{
\centering
\caption{Evaluations on test set of FMoW-C with $r_{\text{id}}=0.5$, all metrics computed using worst-group accuracy.}
\vspace*{-0.8\baselineskip}
\scalebox{0.7}{
\begin{tabular}{llccc}    \toprule
    \multirow{2}{*}{Regime} &\multirow{2}{*}{Method} & \multicolumn{3}{c}{Metrics (\%)} \\
    \cmidrule{3-5} 
    & &  $\text{acc}_o(f,c_o)\uparrow$ & $s\downarrow$ & $b$ \\
    \midrule \rowcolor{lb}
    & AE          &23.4 \std{1e+0} &	8.6  	\std{6e-1}	& 4.2	\std{8e-1} \\  \rowcolor{lb}
    & VAE         &18.7 \std{1e+0} &	7.7  	\std{6e-1}	& 1.5	\std{6e-1} \\  \rowcolor{lb}
    & IWAE        &18.5 \std{2e+0} &	7.3  	\std{2e+0}	& 2.2	\std{1e+0} \\  \rowcolor{lb}
    \multirow{-4}{*}{AE}
    & $\beta$-VAE &21.4 \std{3e-1} &	4.0  	\std{4e-1}	& 3.9	\std{7e-1} \\  \rowcolor{db}
    \multicolumn{2}{l}{\emph{\textbf{AE average}}} & \emph{20.5 \std{2e+0}} &	\emph{6.9  	\std{8e-1}}	& \emph{3.0	\std{9e-1}} \\
    \midrule \rowcolor{lg}
    & SimCLR      &29.5 \std{9e-1} &	9.2  	\std{6e-1}	& 6.9	\std{7e-1}\\ \rowcolor{lg}
    & SimSiam     &27.9 \std{2e+0} &	7.5  	\std{1e+0}	& 4.6	\std{1e+0}\\  \rowcolor{lg}
    \multirow{-3}{*}{SSL} 
    & BYOL        &32.6 \std{3e+0} &	7.9  	\std{2e+0}	& 8.5	\std{2e+0}\\  \rowcolor{dg}
    \multicolumn{2}{l}{\emph{\textbf{SSL average}}} & \emph{30.0 \std{2e+0}} &	\emph{8.2 \std{1e+0}} & \emph{6.7 \std{1e+0}}\\
    \midrule \rowcolor{g}
    SL
    & Supervised  & 32.3 \std{3e+0} &	25.1  	\std{3e+0}	& 6.0	\std{2e+0}\\ 
    \bottomrule
\end{tabular}
\label{tab:fmowcorr50_results}}
}
\end{table}

\begin{table}
\parbox{.48\linewidth}{
\centering
\caption{Evaluations on test set of Camelyon17-C with $r_{\text{id}}=1$, all metrics computed using average accuracy.}
\vspace*{-0.8\baselineskip}
\scalebox{0.7}{
\begin{tabular}{llccc}    \toprule
    \multirow{2}{*}{Regime} &\multirow{2}{*}{Method} & \multicolumn{3}{c}{Metrics (\%)} \\
    \cmidrule{3-5} 
    & &  $\text{acc}_o(f,c_o)\uparrow$ & $s\downarrow$ & $b$ \\
    \midrule \rowcolor{lb}
    & AE          &75.7	\std{5e+0}	& 7.3	 \std{2e+0} &	35.1	\std{4e+0} \\  \rowcolor{lb}
    & VAE         &86.0	\std{3e+0}	& 2.7	 \std{1e+0} &	12.4	\std{4e+0} \\  \rowcolor{lb}
    & IWAE        &86.1	\std{1e+0}	& 2.6	 \std{7e-1} &	9.1 	\std{3e+0} \\  \rowcolor{lb}
    \multirow{-4}{*}{AE}
    & $\beta$-VAE &84.7	\std{2e+0}	& 3.8	 \std{1e+0} &	15.5	\std{4e+0} \\  \rowcolor{db}
    \multicolumn{2}{l}{\emph{\textbf{AE average}}} & \emph{83.1	\std{3e+0}}	& \emph{4.1	 \std{1e+0}} &	\emph{18.0	\std{4e+0}}\\
    \midrule \rowcolor{lg}
    & SimCLR      &85.8	\std{8e+1}	& 2.8	 \std{4e-1} &	6.2 	\std{2e+0} \\ \rowcolor{lg}
    & SimSiam     &82.1	\std{1e+0}	& 6.0	 \std{7e-1} &	8.3 	\std{4e+0}\\  \rowcolor{lg}
    \multirow{-3}{*}{SSL} 
    & BYOL        &74.8	\std{5e+0}	& 10.3 \std{2e+0} &	-2.2	    \std{4e+0}\\  \rowcolor{dg}
    \multicolumn{2}{l}{\emph{\textbf{SSL average}}} &\emph{80.9	\std{2e+0}}	& \emph{6.3	 \std{1e+0}} &	\emph{4.1 	\std{3e+0}} \\
    \midrule \rowcolor{g}
    SL
    & Supervised  &73.4	\std{6e+0}	& 10.7 \std{3e+0} &	5.9 	    \std{8e+0}\\ 
    \bottomrule
\end{tabular}
\label{tab:camelyon17corr100_results}}
}
\hspace{10pt}
\parbox{.48\linewidth}{
\centering
\caption{Evaluations on test set of FMoW-C with $r_{\text{id}}=1$, all metrics computed using worst-group accuracy.}
\vspace*{-0.8\baselineskip}
\scalebox{0.7}{
\begin{tabular}{llccc}    \toprule
    \multirow{2}{*}{Regime} &\multirow{2}{*}{Method} & \multicolumn{3}{c}{Metrics (\%)} \\
    \cmidrule{3-5} 
    & &  $\text{acc}_o(f,c_o)\uparrow$ & $s\downarrow$ & $b$ \\
    \midrule \rowcolor{lb}
    & AE          &22.4	\std{1e+0} &   10.0	\std{6e-1} &	3.8	\std{7e-1} \\  \rowcolor{lb}
    & VAE         &16.6	\std{9e-1} &	8.6	\std{1e+0} &	2.8	\std{8e-1} \\  \rowcolor{lb}
    & IWAE        &17.2	\std{5e-1} &	8.7	\std{6e-1} &	3.8	\std{5e-1} \\  \rowcolor{lb}
    \multirow{-4}{*}{AE}
    & $\beta$-VAE &16.7	\std{3e-1} &	7.9	\std{4e-1} &	3.2	\std{4e-1} \\  \rowcolor{db}
    \multicolumn{2}{l}{\emph{\textbf{AE average}}} & \emph{18.3	\std{3e+0}} &	\emph{8.8	\std{7e-1}} &	\emph{3.4	\std{6e-1}}\\
    \midrule \rowcolor{lg}
    & SimCLR      &26.3	\std{1e+0} &   10.6	\std{8e-1} &	7.2	\std{1e+0}\\ \rowcolor{lg}
    & SimSiam     &27.1	\std{5e-1} &	6.3	\std{7e-1} &	7.8	\std{3e-1}\\  \rowcolor{lg}
    \multirow{-3}{*}{SSL} 
    & BYOL        &27.4	\std{2e+0} &   11.1	\std{1e+0} &	6.5	\std{2e+0}\\  \rowcolor{dg}
    \multicolumn{2}{l}{\emph{\textbf{SSL average}}} &\emph{26.9	\std{6e-1}} &	\emph{9.3	\std{1e+0}} &	\emph{7.2	\std{1e+0}}\\
    \midrule \rowcolor{g}
    SL
    & Supervised  &25.7	\std{5e-1} &   12.3	\std{2e+0} &	5.3	\std{1e+0}\\ 
    \bottomrule
\end{tabular}
\label{tab:fmowcorr100_results}}
}
\end{table}

}

\section{Architecture and Hyperparameters} \label{sec:app_hp}

In this appendix we list the architecture and hyperparameters used in our experiments.
Our code is developed on the amazing \texttt{solo-learn} code base \citep{solo-learn}, which is originally developed as a library for \gls{SSL} algorithms.
For all experiments we follow the standard set of augmentations established in \citet{he2020moco,chen2020improved}, including random resize crop, random color jittering, random grayscale, random Gaussian blur, random solorisation and random horizontal flip. An exception is the CdSprites experiment where we remove color jittering, as color classification is one of the tasks we are interested in and color jittering would add noise to the labels. For MNIST-CIFAR, we independently apply random augmentation to MNIST and CIFAR respectively (drawn from the same set of augmentations as detailed above) and then concatenate them to construct training examples. 

Please see implementation details for each dataset in the respective subsection.

\begin{table*}[t]
% \vspace*{-0.5\baselineskip}
\centering
\caption{Hyperparameter search range for MNIST-CIFAR, including base channel size of CNN ($C$), learning rate (\textit{lr.}), weight decay (\textit{wd.}), optimiser (\textit{optim.}), learning rate scheduler (\textit{lr. scheduler}).} \vspace{-5pt}
\scalebox{0.78}{
\begin{tabular}{lcccccccc}    
    \toprule
    &  $C$              &   lr.  &  wd.          &  Optim.     & lr. scheduler \\ \midrule \rowcolor{lb}
    AE  & \{16, 32, 64, 128\} &  \{1e-4, 5e-4, 1e-3, 5e-3, 1e-2\}  &  \{0, 1e-4\} &  \{Adam, SGD\}       & \{warmup cosine, step, none\} \\ \rowcolor{lg}
    SSL & \{16, 32, 64, 128\} &  uniformly sampled from [0.1, 1]  &  \{0, 1e-4\} &  \{Adam, SGD\}       & \{warmup cosine, step, none\} \\ \rowcolor{g} 
    SL  & \{16, 32, 64, 128\} &  \{1e-4, 5e-4, 1e-3, 5e-3, 1e-2, 1e-1, 5e-1\}  &  \{0, 1e-4\} &  \{Adam, SGD\}  & \{warmup cosine, step, none\} \\
    \bottomrule
\end{tabular}
\label{tab:hp_search_ms}}
\vspace*{0.5\baselineskip}
\end{table*}

\subsection{MNIST-CIFAR}
We use the same hyperparameter search range for models in each category of AE, SSL and SL, as outlined in \cref{tab:hp_search_ms}. The chosen hyperparameters for each model are specified in \cref{tab:hp_ms}.

In \citet{shah2020pitfalls} where MNIST-CIFAR was originally proposed, authors utilised more complex backbone architecture such as DenseNet and MobileNet. 
However in our experiments, we find that a lightweight 4-layer CNN can already achieve very high accuracy on both MNIST and CIFAR.
The architecture of the CNN we use can be found in \cref{tab:arch_ms}. Note that for SL and SSL we only use the encoder and for AE we use the decoder as well. The size of base channel $C$ and latent dimension $L$ are found through hyperparameter search.

\begin{table}[H]
\vspace*{0.5\baselineskip}
\centering
\caption{Chosen hyperparameters for MNIST-CIFAR including latent dimension ($L$), base feature size of CNN ($C$), batch size ($B$), learning rate (\textit{lr.}), weight decay (\textit{wd.}), optimiser (\textit{optim.}), learning rate scheduler (\textit{lr.scheduler}).} \vspace{-5pt}
\scalebox{0.78}{
\begin{tabular}{lcccccccc}    
    \toprule
                & $L$         &  $C$        & $B$         &   lr.  &  wd.          &  Optim.     & lr. scheduler \\ \midrule \rowcolor{lb}
    AE          & 128         & 16          & 128         &  1e-3  &  0            &  Adam       & warmup cosine \\ \rowcolor{lb}
    VAE         & 128         & 32          & 128         &  1e-4  &  0            &  Adam       & warmup cosine \\ \rowcolor{lb}
    IWAE        & 128         & 32          & 128         &  1e-4  &  0            &  Adam       & step          \\ \rowcolor{lb}
    $\beta$-VAE & 128         & 16          & 128         &  1e-4  &  0            &  Adam       & step          \\ \rowcolor{lg}\midrule
    SimCLR      & 128         & 32          & 128         &  6e-1  &  1e-4         &  SGD        & warmup cosine \\ \rowcolor{lg}
    BYOL        & 128         & 64          & 128         &  7e-1  &  0            &  SGD        & warmup cosine \\ \rowcolor{lg}
    SimSiam     & 128         & 128         & 128         &  6e-1  &  1e-5         &  SGD        & warmup cosine \\ \rowcolor{g}\midrule
    Supervised  & 128         & 16          & 128         &  1e-4  &  0            &  SGD        & warmup cosine \\
    \bottomrule
\end{tabular}
\label{tab:hp_ms}}
% \vspace*{-\baselineskip}
\end{table}

\begin{table}[H]
\centering
\scalebox{0.7}{%
\begin{tabular}{l}
    \toprule
    \textbf{Encoder}                                       \\
    \midrule
    Input $\in \mathbb{R}^{3\times64\times32}$                       \\
    4x4 conv.  $C$ stride  2x2 pad 1x1 \& ReLU\\
    4x4 conv.  2$C$ stride 2x2 pad 1x1 \& ReLU\\
    4x4 conv.  4$C$ stride 2x2 pad 1x1 \& ReLU\\
    4x1 conv.  4$C$ stride 2x1 pad 1x0 \& ReLU\\
    4x4 conv.  $L$ stride 1 pad 0, 4x4 conv. $L$ stride 1x1 pad 0x0                             \\
    \bottomrule
    \toprule
    \textbf{Decoder}                                      \\
    \midrule
    Input $\in \mathbb{R}^{L}$                            \\
    4x4 upconv. 4$C$ stride 1x1 pad 0x0 \& ReLU        \\
    4x1 upconv. 4$C$ stride 2x1 pad 1x0 \& ReLU        \\
    4x4 upconv. 2$C$ stride 2x2 pad 1x1 \& ReLU        \\
    4x4 upconv. $C$ stride 2x2 pad 1x1 \& ReLU        \\
    4x4 upconv. 3 stride 2x2 pad 1x1 \& Sigmoid    \\
    \bottomrule
\end{tabular}}
\caption{CNN architecture, MNIST-CIFAR dataset.}
\label{tab:arch_ms}
\end{table}

\subsection{CdSprites}

We found all models to be relatively robust to hyperparameters, as most configurations result in close to perfect shape and color classification accuracy on the ID validation set. The chosen hyperparameters for each model are specified in \cref{tab:hp_cdsprites}. We omit $\beta$-VAE from the comparison, as we empirically found that $\beta = 1$ leads to the best performance on the ID validation set and therefore the results for the $\beta$-VAE would be similar to the VAE.
We use the same augmentations (random crops and horizontal flips) for all models and use no color augmentations in order to keep the invariance of the learned representations with respect to color. The encoder and decoder architectures are described in \cref{tab:arch_cdsprites}.

\begin{table}[H]
\vspace*{1.0\baselineskip}
\centering
\caption{Chosen hyperparameters for CdSprites including latent dimension ($L$), base feature size of CNN ($C$), batch size ($B$), learning rate (\textit{lr.}), weight decay (\textit{wd.}), optimiser (\textit{optim.}), learning rate scheduler (\textit{lr.scheduler}).}
\vspace{-2pt}
\scalebox{0.78}{
\begin{tabular}{lccccccc}    
    \toprule
                & L   & C   & B    &   lr.  &  wd.         &  Optim.     & none          \\ \midrule\rowcolor{lb}
    AE          & 512 & 64 & 128  &  5e-5  &  1e-4         &  Adam       & none          \\ \rowcolor{lb}
    VAE         & 512 & 64 & 128  &  5e-5  &  1e-4         &  Adam       & none          \\ \rowcolor{lb}
    IWAE        & 512 & 64 & 128  &  5e-5  &  1e-4         &  Adam       & none          \\ \rowcolor{lg}\midrule
    %$\beta$-VAE & 512 & 64 & 128  &  5e-5  &  1e-4         &  Adam       & none          \\ \rowcolor{lg}\midrule
    SimCLR      & 64  & 32 & 64   &  5e-3  &  1e-5         &  SGD        & warmup cosine \\ \rowcolor{lg}
    BYOL        & 64  & 32 & 64   &  5e-1  &  1e-5         &  SGD        & warmup cosine \\ \rowcolor{lg}
    SimSiam     & 64  & 32 & 64   &  8e-2  &  1e-5         &  SGD        & warmup cosine \\ \rowcolor{g}\midrule
    Supervised  & 512 & 64 & 128  &  5e-5  &  1e-4         &  Adam       & none          \\
    \bottomrule
\end{tabular}
\label{tab:hp_cdsprites}}
% \vspace*{-\baselineskip}
\end{table}

\begin{table}[H]
\centering
\scalebox{0.7}{%
\begin{tabular}{l}
    \toprule
    \textbf{Encoder}                                       \\
    \midrule
    Input $\in \mathbb{R}^{3\times64\times64}$                       \\
    4x4 conv.  $C$ stride  2x2 pad 1x1 \& ReLU\\
    4x4 conv.  2$C$ stride 2x2 pad 1x1 \& ReLU\\
    4x4 conv.  4$C$ stride 2x2 pad 1x1 \& ReLU\\
    4x4 conv.  8$C$ stride 2x2 pad 1x1 \& ReLU\\
    4x4 conv.  $L$ stride 1 pad 0                        \\
    \bottomrule
    \toprule
    \textbf{Decoder}                                      \\
    \midrule
    Input $\in \mathbb{R}^{L}$                            \\
    4x4 upconv. 8$C$ stride 1x1 pad 0x0 \& ReLU        \\
    4x4 upconv. 4$C$ stride 2x2 pad 1x1 \& ReLU        \\
    4x4 upconv. 2$C$ stride 2x2 pad 1x1 \& ReLU        \\
    4x4 upconv. $C$ stride 2x2 pad 1x1 \& ReLU        \\
    4x4 upconv. 3 stride 2x2 pad 1x1 \\  % NOTE: no sigmoid
    \bottomrule
\end{tabular}}
\caption{CNN architecture, CdSprites dataset.}
\label{tab:arch_cdsprites}
\end{table}

\subsection{Camelyon17 and FMoW}
\begin{table*}[t]
\vspace*{1.0\baselineskip}
\centering
\caption{Hyperparameter search range for Camelyon17, including decoder type, latent dimension ($L$), learning rate (\textit{lr.}), weight decay (\textit{wd.}), optimiser (\textit{optim.}), learning rate scheduler (\textit{lr. scheduler}).} \vspace{-5pt}
\scalebox{0.6}{
\begin{tabular}{lcccccccc}    
    \toprule
    & Decoder type & $L$ &   lr.  &  wd.          &  Optim.     & lr. scheduler \\ \midrule \rowcolor{lb}
    AE  & [CNN, MLP, ResNet] & \{256, 512, 1024\}  & \{1e-4, 5e-4, 1e-3, 5e-3, 1e-2\}  &  \{0, 1e-4\} &  \{Adam, SGD\}       & \{warmup cosine, step, none\} \\ \rowcolor{lg}
    SSL & - & \{256, 512, 1024\} & \{1e-4, 5e-4, 1e-3, 5e-3, 1e-2, 1e-1, 5e-1, 1\}  &  \{0, 1e-3, 1e-4, 1e-5\} &  \{Adam, SGD\}       & \{warmup cosine, step, none\} \\ \rowcolor{g} 
    SL  & - & \{256, 512, 1024\} &  \{1e-4, 5e-4, 1e-3, 5e-3, 1e-2, 1e-1, 5e-1\}  &  \{0, 1e-4\} &  \{Adam, SGD\}  & \{warmup cosine, step, none\} \\
    \bottomrule
\end{tabular}
\label{tab:hp_search_camelyon}}
%\vspace*{-\baselineskip}
\end{table*}

For hyperparameters including batch size, max epoch and model selection criteria, we follow the same protocol as in WILDS \citep{koh2021wilds}: for Camelyon17 we use a batch size of 32, train all models for 10 epochs and select the model that results in the highest accuracy on the validation set, and for FMoW the batch size is 32, max epoch is 60 and model selection criteria is worst group accuracy on OOD validation set.
For the rest, we use the same hyperparameter search range for models in each category of AE, SSL and SL, as outlined in \cref{tab:hp_search_camelyon}. The chosen hyperparameters for Camelyon17 are specified in \cref{tab:hp_camelyon}, and for FMoW in \cref{tab:hp_fmow}.
For Camelyon17-C and FMoW-C we use these same hyperparameters.

\begin{table}[H]
% \vspace*{-0.5\baselineskip}
\centering
\caption{Chosen hyperparameters for Camelyon17 including latent dimension ($L$), learning rate (\textit{lr.}), weight decay (\textit{wd.}), optimiser (\textit{optim.}), learning rate scheduler (\textit{lr. scheduler}).} \vspace{-5pt}
\scalebox{0.78}{
\begin{tabular}{lcccccccc}    
    \toprule
                & Decoder     &   lr.  &  wd.          &  Optim.     & lr. scheduler \\ \midrule \rowcolor{lb}
    AE          & ResNet      &  5e-4  &  1e-5         &  SGD        & warmup cosine \\ \rowcolor{lb}
    VAE         & MLP         &  1e-4  &  0            &  Adam       & none          \\ \rowcolor{lb}
    IWAE        & MLP         &  1e-4  &  0            &  Adam       & none          \\ \rowcolor{lb}
    $\beta$-VAE & MLP         &  1e-4  &  0            &  Adam       & none          \\ \rowcolor{lg}\midrule
    SimCLR      & -           &  1e-1  &  0            &  SGD        & none          \\ \rowcolor{lg}
    BYOL        & -           &  1e-1  &  1e-5         &  SGD        & warmup cosine \\ \rowcolor{lg}
    SimSiam     & -           &  1e-1  &  1e-5         &  SGD        & warmup cosine \\ \rowcolor{g}\midrule
    Supervised  & -           &  1e-3  &  1e-3         &  SGD        & none          \\
    \bottomrule
\end{tabular}
\label{tab:hp_camelyon}}
\vspace*{-\baselineskip}
\end{table}

\begin{table}[H]
% \vspace*{-0.5\baselineskip}
\centering
\caption{Chosen hyperparameters for FMoW including latent dimension ($L$), learning rate (\textit{lr.}), weight decay (\textit{wd.}), optimiser (\textit{optim.}), learning rate scheduler (\textit{lr. scheduler}).} \vspace{-5pt}
\scalebox{0.78}{
\begin{tabular}{lcccccccc}    
    \toprule
                & Decoder     &   lr.  &  wd.          &  Optim.     & lr. scheduler \\ \midrule \rowcolor{lb}
    AE          & CNN         &  1e-1  &  1e-4         &  SGD        & none          \\ \rowcolor{lb}
    VAE         & MLP         &  1e-6  &  1e-4         &  Adam       & step          \\ \rowcolor{lb}
    IWAE        & MLP         &  1e-6  &  1e-4         &  Adam       & step          \\ \rowcolor{lb}
    $\beta$-VAE & MLP         &  1e-6  &  1e-4         &  Adam       & step          \\ \rowcolor{lg}\midrule
    SimCLR      & -           &  5e-4  &  1e-3         &  SGD        & step          \\ \rowcolor{lg}
    BYOL        & -           &  1e-2  &  1e-4         &  SGD        & step          \\ \rowcolor{lg}
    SimSiam     & -           &  5e-4  &  0            &  SGD        & step          \\ \rowcolor{g}\midrule
    Supervised  & -           &  1e-4  &  0            &  Adam       & step          \\
    \bottomrule
\end{tabular}
\label{tab:hp_fmow}}
\vspace*{-\baselineskip}
\end{table}

We follow \citet{koh2021wilds} and use DenseNet121 \citep{dense} as backbone architecture.
For the decoder of the \gls{AE} models, we perform hyperparameter search between three architectures: a CNN (see \cref{tab:arch_cnn_camelyon}), a simple 3-layer MLP  (see \cref{tab:arch_mlp_camelyon}) and a ResNet-like decoder with skip connections (see  \cref{tab:arch_resnet_camelyon}).

\begin{table}[H]
\centering
\scalebox{0.7}{%
\begin{tabular}{l}
    \toprule
    \textbf{CNN, Decoder} \\
    \midrule
    Input $\in \mathbb{R}^{L}$ \\
    4x4 upconv. 8$C$ stride 2x2 pad 1x1 \& ReLU        \\
    4x4 upconv. 8$C$ stride 2x2 pad 0x0 \& ReLU        \\
    4x4 upconv. 4$C$ stride 2x2 pad 1x1 \& ReLU        \\
    4x4 upconv. 2$C$ stride 2x2 pad 1x1 \& ReLU        \\
    4x4 upconv. $C$ stride 2x2 pad 1x1 \& ReLU        \\
    4x4 upconv. 3 stride 2x2 pad 1x1 \& Sigmoid    \\
    \bottomrule
\end{tabular}}
\caption{CNN architecture, Camelyon17 dataset.}
\label{tab:arch_cnn_camelyon}
\end{table}

\begin{table}[H]
\centering
\scalebox{0.7}{%
\begin{tabular}{l}
    \toprule 
    \textbf{MLP, Decoder} \\
    \midrule
    Input $\in \mathbb{R}^{L}$ \\
    fc. 2$L$ \& ReLU        \\
    fc. 4$L$ \& ReLU        \\
    fc. 3*96*96 \& ReLU        \\
    \bottomrule
\end{tabular}}
\caption{MLP architecture, Camelyon17 dataset.}
\label{tab:arch_mlp_camelyon}
\end{table}

\begin{table}[H]
\centering
\scalebox{0.7}{%
\begin{tabular}{l}
    \toprule
    \textbf{ResNet, Decoder} \\
    \midrule
    Input $\in \mathbb{R}^{L}$ \\
    fc. 2048 \& ReLU        \\
    3x3 conv. 16$C$ stride 1x1 pad 1x1  \\
    3x3 conv. 16$C$ stride 1x1 pad 1x1  \\
    x2 upsample       \\
    3x3 conv. 8$C$ stride 1x1 pad 1x1  \\
    3x3 conv. 8$C$ stride 1x1 pad 1x1  \\
    x2 upsample       \\
    3x3 conv. 8$C$ stride 1x1 pad 1x1  \\
    3x3 conv. 8$C$ stride 1x1 pad 1x1  \\
    3x3 conv. 3 stride 1x1 pad 1x1  \\
    \bottomrule
\end{tabular}}
\caption{ResNet decoder architecture, Camelyon17 dataset.}
\label{tab:arch_resnet_camelyon}
\end{table}
\section{Constructing Camelyon17-CS and FMoW-CS}\label{app:controllable_datasets}

We subsample Camelyon17 and FMoW dataset to create varying degree of spurious correlation between the domain and label information.
We refer to these datasets as Camelyon17-CS and FMoW-CS.
To construct such datasets, we first find some domain-label pairing in each dataset, such that if we sample the dataset according to this pairing, the population of each class with respect to the total number of examples in the dataset remains relatively stable.
%
% For instance, let us consider a dataset of size $N$, with domain $\{D 1,D 2\}$ as well as classes $\{C 1,C 2\}$.
% %
% Our goal is to find a domain-label pairing $(D i, C j)$ and $(\overline{D i}, \overline{C j})$ such that $|C j|/N \sim |D i \cap C j|/$
%
The $r_{\text{id}}=1$ versions of both Camelyon17-CS and FMoW-CS can be acquired by simply subsampling the dataset following the domain label pairing;
to ensure fairness in comparison, when constructing the $r_{\text{id}} \in \{0,0.5\}$ versions of these datasets, we first mix in anti-bias samples (i.e. samples that are not in the domain-label pairing) to change the spurious correlation, and then subsample the dataset such that the size of the dataset is the same as the $r_{\text{id}}=1$ version.

The domain-label pairing of Camelyon17-CS can be found in \cref{tab:camelyon17 c dl} and FMoW-CS in \cref{tab:fmow c dl}.

\begin{figure*}[t]
    \centering
    % \captionsetup[subfigure]{belowskip=1ex}
    % \begin{subfigure}{\linewidth}
      \centering
      \includegraphics[width=0.32\linewidth]{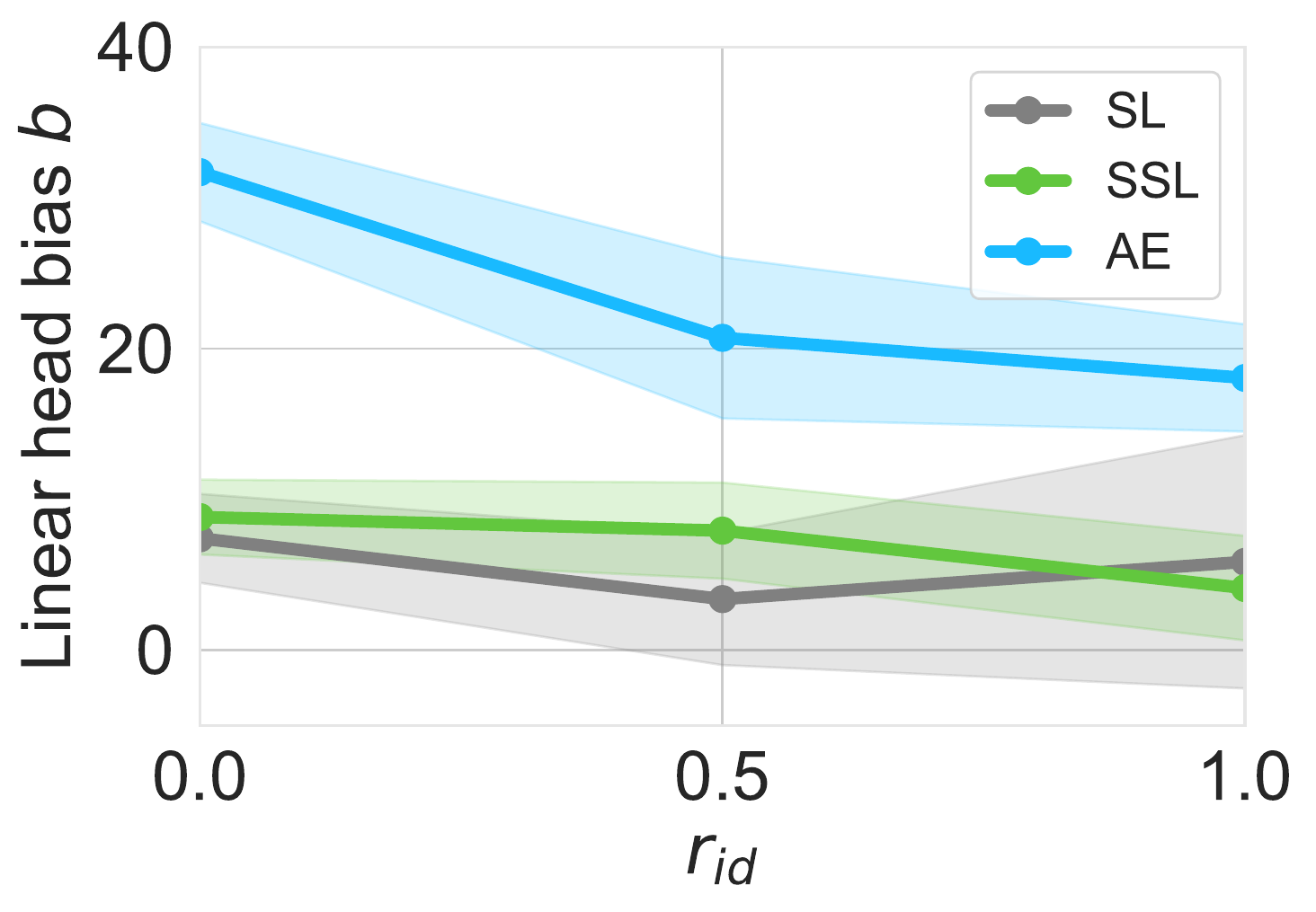}
      \includegraphics[width=0.32\linewidth]{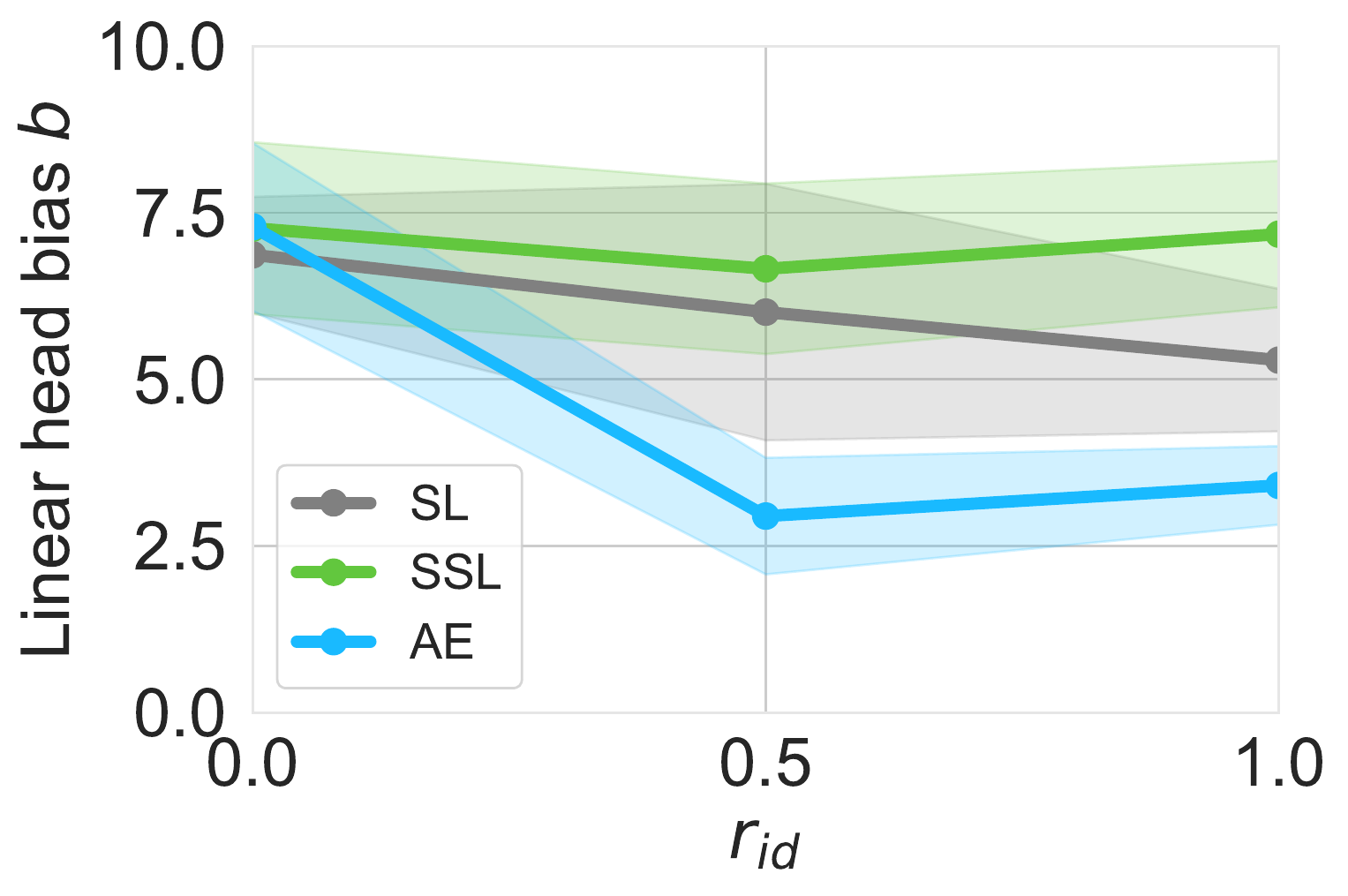}
      \includegraphics[width=0.32\linewidth]{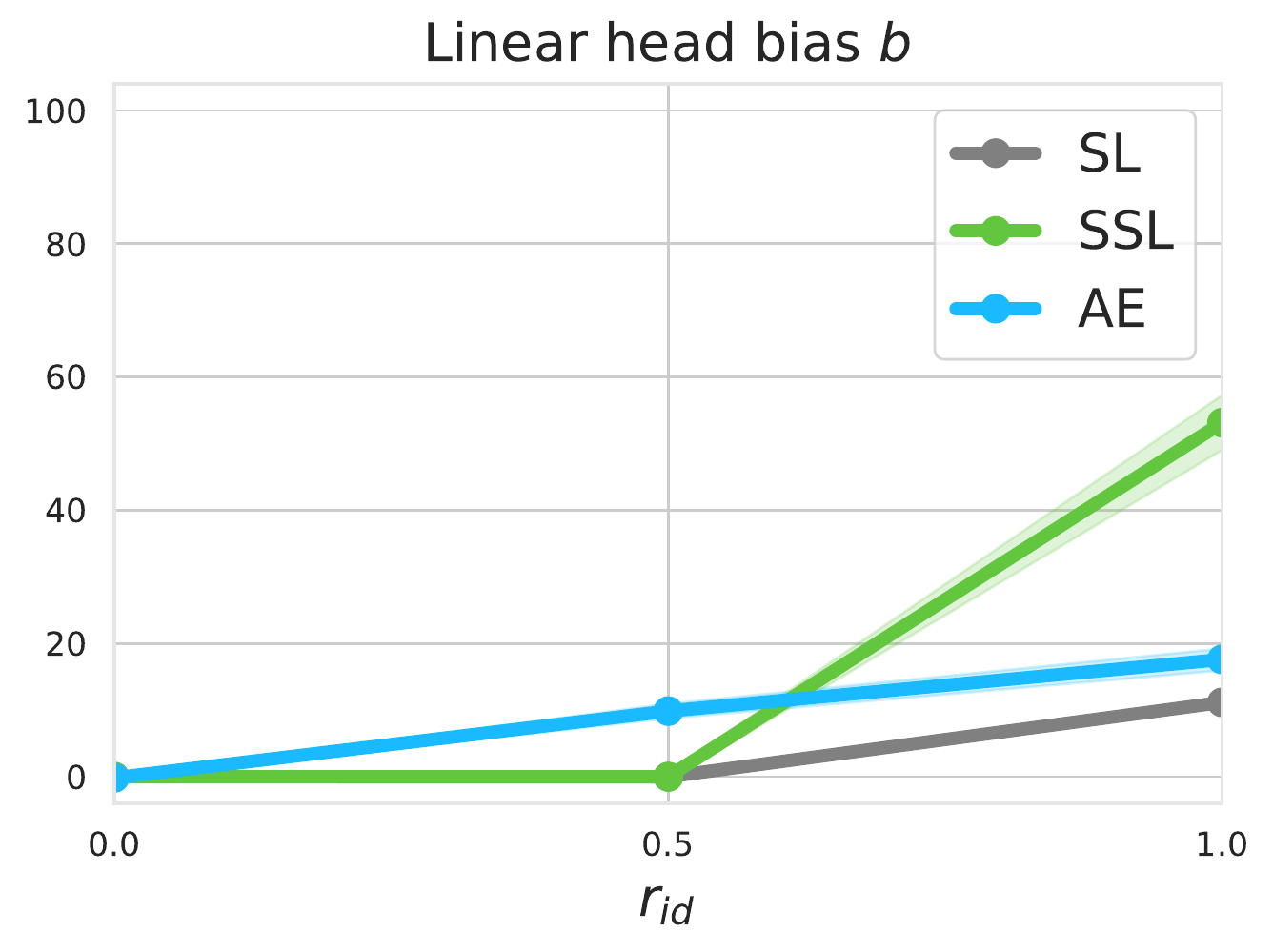}
      \caption{Linear head bias on controllable shift datasets}
      \label{fig:controllable_results_b}
    % \end{subfigure}
    % \begin{subfigure}
      % 
      % \includegraphics[width=0.32\linewidth]{images/camelyon_fmow_c/b_camelyon17.pdf}\vspace{-8pt}
      % \caption{Camelyon17-CS  (\%).}\label{fig:camelyon17_results}
    \end{figure*}

\paragraph{Linear head bias} We also plot the linear head bias for the
experiments conducted on Camelyon17-CS, FMoW-CS, and CdSprites
in~\Cref{fig:controllable_results_b}. The experimental protocol
follows that from~\Cref{fig:controllable_results}.

\begin{table}[H]
% \vspace*{-0.5\baselineskip}
\centering
\caption{Domain-label pairing for Camelyon17-CS.} \vspace{-5pt}
\scalebox{0.78}{
\begin{tabular}{lc}    
    \toprule
    Domain (hospital) & Label \\ \midrule
    Hospital 1, 2 & Benign \\
    Hospital 3    & Malignant \\
    \bottomrule
\end{tabular}
\label{tab:camelyon17 c dl}}
\vspace*{-\baselineskip}
\end{table}

% {Asia: [military facility, multi-unit residential, tunnel opening, wind farm, toll booth, road bridge, oil or gas facility, helipad, nuclear powerplant, police station, port], 

% Europe: [smokestack, barn, waste disposal, hospital, water treatment facility, amusement park, fire station, fountain, construction site, shipyard, solar farm, space facility], 

% Africa: [place of worship, crop field, dam, tower, runway, airport, electric substation, flooded road, border checkpoint, prison, archaeological site, factory or powerplant, impoverished settlement, lake or pond], 

% Americas: [recreational facility, swimming pool, educational institution, stadium, golf course, office building, interchange, car dealership, railway bridge, storage tank, surface mine, zoo],

% Oceania: [single-unit residential, parking lot or garage, race track, park, ground transportation station, shopping mall, airport terminal, airport hangar, lighthouse, gas station, aquaculture, burial site, debris or rubble], Other: [airport]}

\begin{table}[H]
% \vspace*{-0.5\baselineskip}
\centering
\caption{Domain-label pairing for FMoW-CS.} \vspace{-5pt}
\scalebox{0.78}{
\begin{tabular}{lc}    
    \toprule
    Domain (region) & Label \\ \midrule
    \multirow{4}{*}{Asia}  &  \multirow{4}{*}{\shortstack{Military facility, multi-unit residential, tunnel opening, \\ wind farm, toll booth, road bridge, oil or gas facility,\\ helipad, nuclear powerplant, police station, port}} \\
    & \\
    & \\
    & \\
    \multirow{4}{*}{Europe}  &  \multirow{4}{*}{\shortstack{Smokestack, barn, waste disposal, hospital, water  \\ treatment facility, amusement park, fire station, fountain, \\ construction site, shipyard, solar farm, space facility}} \\
    & \\
    & \\
    & \\
    \multirow{4}{*}{Africa}  &  \multirow{4}{*}{\shortstack{Place of worship, crop field, dam, tower, runway, airport, electric \\ substation, flooded road, border checkpoint, prison, archaeological site, \\factory or powerplant, impoverished settlement, lake or pond}} \\
    & \\
    & \\
    & \\
    \multirow{4}{*}{Americas}  &  \multirow{4}{*}{\shortstack{Recreational facility, swimming pool, educational institution, \\stadium, golf course, office building, interchange, \\ car dealership, railway bridge, storage tank, surface mine, zoo}} \\
    & \\
    & \\
    & \\
    \multirow{4}{*}{Oceania}  &  \multirow{4}{*}{\shortstack{Single-unit residential, parking lot or garage, race track, park, ground \\ transportation station, shopping mall, airport terminal, airport hangar,  \\ lighthouse, gas station, aquaculture, burial site, debris or rubble}} \\
    & \\
    & \\
    & \\
    \bottomrule
\end{tabular}
\label{tab:fmow c dl}}
\vspace*{-\baselineskip}
\end{table}

\section{Other Related Work} \label{app:related_work}

\paragraph{Explicit, extreme distribution shift} 
Refers to when the features that caused distribution shift is explicit, known, controllable, and in some cases, extreme (e.g. MNIST-CIFAR, CdSprites).
This type of settings are popular in works that investigate simplicity bias \citep{ shah2020pitfalls}, dataset bias \citep{torralba2011datasetbias1} and shortcut learning \citep{geirhos2020shortcut, lapuschkin2019shortcut2}, as it allows for users to easily adjust the level of distribution shift between train and test set.
Various specialised methods that either mitigate or address these problems under the supervised learning regime have been proposed, including \citet{teney2022evading} that proposes to find shortcut solutions by ensembles, \citet{luo2021rectifying} that avoids shortcut learning by extracting foreground objects for representation learning only, as well as \citet{torralba2011datasetbias1, kim2019datasetbias2, le2020datasetbias3} that re-sample the dataset to reduce the spurious correlation.

Importantly \citet{kirichenko2022last, kang2019decoupling} shows that this extreme simplicity bias can be mitigated in some cases by retraining the final linear layer.
This is a game changer, as it for the first time decouples the bias of the linear head from that of the main representation learning model.
Interestingly, \citet{robinson2021simplicity} also investigates the robustness of contrastive-SSL methods against simplicity bias. However, without training the linear head on OOD data, their finding is opposite to ours --- that SSL methods are not able to avoid shortcut solutions.

\paragraph{Implicit, subtle distribution shift}
This type of problem is commonly seen in realistic distribution shift datasets (such as WILDS) and are often studied in Domain generalisation (DG).
In this regime, the training data are sampled from multiple domains, while data from a new, unseen target domain is used as test set.
Note that here we omitted the discussion on domain adaptation (DA), as in DA models typically have access to unlabelled data from the target domain, which is different from the settings we consider in this work.

There are mainly two lines of work in DG, namely
1) \textit{Distributional Robustness approaches (DRO)}, which minimises the worst group accuracy to address covariate shift \citep{gretton2008covariate,gretton2009covariate} and subpopulation shift \citep{Sagawa2020dro, groupdro0hu};
2) \textit{Domain invariance}, which consists of methods that directly learn representations that are invariant across domains  \citep{ben2010theory,ganin2016dann,wang2019learning}, encourage the alignment of gradients from different domains \citep{koyama2021out,lopez-paz2017gradient,fish}, or optimise for representations that result in the same optimal classifier for different domains \citep{arjovsky2019irm}.
Apart from these supervised learning methods, the recent advancement of SSL has also inspired works in unsupervised domain generalisation \citep{zhang2022towards, harary2022unsupervised}.
While all these methods achieved impressive performance, we note that they are all specially designed for DG with the majority of the methods relying on domain information and label information.
In contrast, our work studies how existing standard representation learning methods such as SSL and AE performs on DG tasks, with none of the methods relying on human annotations.

\end{document}